\pdfoutput=1
\documentclass[10pt]{article} %
\usepackage[preprint]{tmlr}

\usepackage[utf8]{inputenc}
\usepackage{latexsym}
\usepackage{microtype}
\usepackage{graphicx}
\usepackage{booktabs}
\usepackage{array}
\usepackage{amsmath,amssymb,amsthm}
\usepackage{xcolor}
\definecolor{darkblue}{rgb}{0,0,0.5}
\usepackage{hyperref}
\hypersetup{colorlinks=true,citecolor=darkblue,linkcolor=darkblue,urlcolor=darkblue}
\usepackage{url}
\makeatletter\providecommand\insert@pcolumn{\insert@column}\makeatother
\usepackage{colortbl}
\usepackage{multirow}
\usepackage{xspace}
\usepackage[inline]{enumitem}
\usepackage{placeins}
\usepackage{needspace}
\usepackage[most]{tcolorbox}
\usepackage{algorithm}
\usepackage[noend]{algpseudocode}
\makeatletter
\providecommand{\theHALG@line}{}
\renewcommand{\theHALG@line}{\thealgorithm.\arabic{ALG@line}}
\makeatother
\usepackage{caption}
\AddToHook{env/table/begin}{\renewcommand{\arraystretch}{1.18}\setlength{\extrarowheight}{0.3pt}}
\usepackage[capitalize,nameinlink,noabbrev]{cleveref}
\usepackage{subcaption}
\usepackage{tikz}
\usetikzlibrary{shapes.geometric}

\newcommand{\coderepo}{\url{https://github.com/Starscream-11813/bangla-idiom-subspaces}}

\newcommand{\ssi}{\mathrm{SSI}}
\newcommand{\cond}[1]{\texttt{#1}}
\newcommand{\pp}{\,pp}
\newcommand{\E}{\mathbb{E}}
\newcommand{\R}{\mathbb{R}}
\newcommand{\tr}{\operatorname{tr}}
\newcommand{\spn}{\operatorname{span}}
\newcommand{\para}[1]{\vspace{1pt}\noindent\textbf{#1}\hspace{4pt}}
\newlength{\cellwd}
\newcommand{\fixbox}[3]{\settowidth{\cellwd}{#2}\makebox[\cellwd][#1]{#3}}
\newcommand{\mv}[1]{\fixbox{r}{00.0}{#1}}
\newcommand{\pmsd}[1]{\fixbox{l}{\,{\scriptsize$\pm$00.0}}{\,{\scriptsize$\pm$#1}}}
\theoremstyle{plain}
\newtheorem{proposition}{Proposition}
\newtheorem{lemma}{Lemma}
\theoremstyle{remark}
\newtheorem*{reading}{Reading}
\newtheorem{example}{Example}
\algrenewcommand\algorithmicrequire{\textbf{Input:}}
\algrenewcommand\algorithmiccomment[1]{\hfill{\color{muted}$\triangleright$~#1}}
\newcommand{\pref}[1]{Prop.~\labelcref{#1}}
\newcommand{\lref}[1]{Lem.~\labelcref{#1}}

\crefname{proposition}{Proposition}{Propositions}
\crefname{algorithm}{Algorithm}{Algorithms}
\crefname{lemma}{Lemma}{Lemmas}
\crefname{example}{Example}{Examples}
\crefname{figure}{Figure}{Figures}
\crefname{table}{Table}{Tables}
\crefname{equation}{Eq.}{Eqs.}
\Crefname{equation}{Equation}{Equations}

\definecolor{muted}{HTML}{5B6474}
\definecolor{hspurple}{HTML}{5445B5}
\definecolor{lavender}{HTML}{A7A8F0}
\definecolor{lavtint}{HTML}{ECECFC}
\definecolor{boxframe}{HTML}{A7A8F0}
\definecolor{boxtitle}{HTML}{8180DE}
\definecolor{boxback}{HTML}{F5F5FE}

\definecolor{keytint}{HTML}{E3E1FA}
\definecolor{negtint}{HTML}{FDE8DC}
\definecolor{bandtint}{HTML}{F2F2F7}
\definecolor{bntint}{HTML}{DDF2FC}
\definecolor{entint}{HTML}{FBE6FE}
\definecolor{bntag}{HTML}{C8ECFC}
\definecolor{entag}{HTML}{F6D2FE}
\makeatletter
\newcommand{\tint}[2]{\gdef\CT@cell@color{\CT@color{#1}\tint@edge#2\global\let\CT@cell@color\relax}}
\def\tint@edge#1{\if l#1\@tempdimb\z@\else\if r#1\@tempdimc\z@\fi\fi}
\makeatother
\newcommand{\hlkey}{\tint{keytint}{n}}\newcommand{\hlkeyl}{\tint{keytint}{l}}\newcommand{\hlkeyr}{\tint{keytint}{r}}
\newcommand{\hlneg}{\tint{negtint}{n}}\newcommand{\hlnegr}{\tint{negtint}{r}}
\newcommand{\band}[3]{\rowcolor{bandtint}[0pt][0pt]\multicolumn{#1}{@{}l@{}}{\hspace{3pt}%
  \textcolor{hspurple}{\textbf{#2}}\hspace{0.7em}\textcolor{muted}{#3}}}
\newcommand{\swatch}[1]{{\setlength{\fboxsep}{0pt}\setlength{\fboxrule}{0.35pt}%
  \fcolorbox{muted!45}{#1}{\rule{0pt}{0.68em}\rule{1.05em}{0pt}}}}
\newcommand{\hdtwo}[1]{\multirow{2}{*}[-0.55ex]{#1}}
\newtcbox{\langtag}[1]{on line, colback=#1, colframe=#1, boxrule=0pt, arc=1.3pt, boxsep=0pt, left=1.5pt, right=1.5pt,
  top=0.5pt, bottom=0.3pt, fontupper=\footnotesize}
\newcommand{\EN}{\langtag{entag}{EN}}
\newcommand{\BN}{\langtag{bntag}{BN}}

\newcommand{\hyp}[1]{\texorpdfstring{\ensuremath{H_{#1}}}{H#1}}
\newcommand{\tagsub}[2]{\texorpdfstring{#1\textsubscript{#2}}{#1#2}}
\newcommand{\expt}[1]{\tagsub{E}{#1}}
\newcommand{\fcl}[1]{\tagsub{F}{#1}}
\newcommand{\rqn}[1]{\tagsub{RQ}{#1}}

\newtcolorbox{casebox}[1]{enhanced, breakable, colback=boxback, colframe=lavender, boxrule=0.6pt, arc=2pt,
  left=5pt, right=5pt, top=3pt, bottom=3pt, fonttitle=\bfseries\small, coltitle=white, colbacktitle=hspurple,
  before title={\hypersetup{citecolor=white,linkcolor=white}}, title={#1}, fontupper=\small}
\newtcolorbox{caseboxfloat}[1]{enhanced, colback=boxback, colframe=lavender, boxrule=0.6pt, arc=2pt,
  left=5pt, right=5pt, top=3pt, bottom=3pt, fonttitle=\bfseries\small, coltitle=white, colbacktitle=hspurple,
  before title={\hypersetup{citecolor=white,linkcolor=white}}, title={#1}, fontupper=\small}
\newtcolorbox{tocbox}{enhanced, colback=boxback, colframe=boxframe, boxrule=0.5pt, arc=2pt, left=6pt, right=6pt,
  top=4pt, bottom=4pt, fontupper=\small}
\newcommand{\tocentry}[1]{\par\noindent\hangindent=#1\hangafter=1\rightskip=0pt plus 1fil\parfillskip=0pt
  \hyphenpenalty=10000\exhyphenpenalty=10000}
\newcommand{\tocdots}{\nobreak\leaders\hbox to .44em{\hss.\hss}\hskip 1.2em plus 1fill\kern0pt}
\newcommand{\tocline}[3]{\tocentry{1.4em}\makebox[1.4em][l]{\ref*{#1}}\hyperref[#1]{#3}\tocdots\pageref{#1}\par}
\newcommand{\tocsub}[3]{\tocentry{3.8em}\hspace*{1.4em}\makebox[2.4em][l]{\footnotesize\ref*{#1}}%
  \hyperref[#1]{\footnotesize #3}\tocdots{\footnotesize\pageref{#1}}\par}

\makeatletter
\newcommand\bn@set[4]{\expandafter\def\csname bn@#1\endcsname{{#2}{#3}{#4}}}
\newcommand\bn@text[3]{\expandafter\def\csname bnr@#1\endcsname{#2}\expandafter\def\csname bnl@#1\endcsname{#3}}
\bn@set{gache}{1}{9.940}{0.840}
\bn@text{gache}{gache kanthal gonphe tel}{jackfruit on the tree, oil on the mustache}
\bn@set{gacheSent}{2}{9.940}{1.880}
\bn@text{gacheSent}{}{}
\bn@set{gacheSentA}{3}{9.110}{1.880}
\bn@text{gacheSentA}{}{}
\bn@set{gacheSentB}{4}{9.940}{0.840}
\bn@text{gacheSentB}{}{}
\bn@set{gacheCtxB}{5}{9.100}{0.840}
\bn@text{gacheCtxB}{}{}
\bn@set{gacheFrame}{6}{9.940}{0.840}
\bn@text{gacheFrame}{}{}
\bn@set{gacheGloss}{7}{9.100}{0.790}
\bn@text{gacheGloss}{praptir agei bhoger ayojon}{arranging enjoyment before attainment}
\bn@set{gacheGlossFull}{8}{9.300}{2.650}
\bn@text{gacheGlossFull}{}{}
\bn@set{kata}{9}{9.110}{2.430}
\bn@text{kata}{kata ghaye nuner chhita}{a sprinkle of salt on a cut wound}
\bn@set{ulubon}{10}{9.080}{2.280}
\bn@text{ulubon}{ulubone mukta chhorano}{to scatter pearls in a field of thatch grass}
\bn@set{dudh}{11}{9.100}{2.280}
\bn@text{dudh}{dudh kola diye shap posha}{to rear a snake on milk and bananas}
\bn@set{ghora}{12}{9.100}{1.280}
\bn@text{ghora}{ghora dingiye ghash khawa}{to leap over the horse to eat the grass}
\bn@set{jotoGorje}{13}{9.300}{0.140}
\bn@text{jotoGorje}{joto gorje toto borshe na}{it does not rain as much as it thunders}
\bn@set{bamon}{14}{9.230}{0.790}
\bn@text{bamon}{bamon hoye chande hat}{a dwarf reaching for the moon}
\bn@set{otiChalak}{15}{9.100}{1.280}
\bn@text{otiChalak}{oti chalaker golay dori}{a rope around the neck of the over-clever}
\bn@set{chorer}{16}{7.510}{1.280}
\bn@text{chorer}{chorer mayer boro gola}{the thief's mother has the loudest voice}
\bn@set{nake}{17}{9.100}{2.080}
\bn@text{nake}{nake tel diye ghumano}{to sleep with oil in one's nose}
\bn@set{goroj}{18}{9.040}{1.280}
\bn@text{goroj}{goroj boro balai}{need is a great affliction}
\bn@set{konok}{19}{7.470}{0.140}
\bn@text{konok}{konokambhoj}{golden lotus}
\bn@set{bhikkha}{20}{9.100}{2.030}
\bn@text{bhikkha}{bhikkhaputro}{alms-son}
\bn@set{tori}{21}{9.100}{0.000}
\bn@text{tori}{tori-torkari}{greens and vegetables}
\bn@set{paka}{22}{7.490}{0.840}
\bn@text{paka}{paka chhele}{a ripe boy}
\newcommand\bn@scale{1.08}%
\DeclareRobustCommand\bn[1]{\@ifundefined{bn@#1}{\textbf{??}\@latex@warning{Bangla snippet #1 undefined}}%
  {\expandafter\expandafter\expandafter\bn@put\csname bn@#1\endcsname}}
\newcommand\bn@put[3]{\leavevmode\begingroup\dimen@=\f@size\p@ \dimen@=\bn@scale\dimen@ \dimen@=.1\dimen@
  \edef\bn@s{\strip@pt\dimen@}%
  \kern-\bn@s\dimexpr0.4pt\relax
  \raisebox{-\bn@s\dimexpr#3pt+1.6pt\relax}[\bn@s\dimexpr#2pt\relax][\bn@s\dimexpr#3pt\relax]%
    {\includegraphics[scale=\bn@s,page=#1]{figures/bangla_snippets.pdf}}%
  \kern-\bn@s\dimexpr0.4pt\relax\endgroup}
\DeclareRobustCommand\bnrom[1]{\textit{\csname bnr@#1\endcsname}}
\DeclareRobustCommand\bnlit[1]{`\csname bnl@#1\endcsname'}
\makeatother

\newcommand{\ThReadOK}{12}
\newcommand{\ThEraseOK}{11}
\newcommand{\ThBothOK}{11}
\newcommand{\ThCells}{12}
\newcommand{\ThOverlapF}{0.11--0.33}

\newcommand{\ThSqrtF}{3}
\newcommand{\ThSqrtS}{5}

\newcommand{\ThToyOmega}{0.56}
\newcommand{\ThToyRead}{0.74}
\newcommand{\ThToyErase}{0.35}

\newcommand{\FuHoneBothF}{3}
\newcommand{\FuHoneBothS}{2}

\newcommand{\FuHtwoNF}{6}
\newcommand{\FuHtwoNS}{5}

\newcommand{\FuHthreeNF}{6}
\newcommand{\FuHthreeNS}{0}
\newcommand{\FuHfourNF}{4}
\newcommand{\FuHfourNS}{3}

\newcommand{\FuOverlapFmin}{7.2--9.2}

\newcommand{\FuRescueFail}{12}
\newcommand{\FuCells}{12}

\newcommand{\LayerBestOnly}{BanglaBERT}
\newcommand{\LayerGap}{0.04--0.25}
\newcommand{\LayerLateTop}{0.41--0.55}
\newcommand{\LayerLateXLMR}{0.20--0.29}

\newcommand{\TxGlossSSIF}{0.55--1.06}
\newcommand{\TxCounterGlossSSIF}{0.28--0.59}

\newcommand{\TxCovMlrRatio}{1.9--4.3}
\newcommand{\TxCovMlrRatioN}{five}

\newcommand{\TxEraseBnF}{4.25--11.03}
\newcommand{\TxEraseBnFRound}{4--11}
\newcommand{\TxEraseEnMaxF}{0.74}
\newcommand{\TxEraseShareF}{$-$2\% to 12\%}

\definecolor{mkBanglaBERTcol}{HTML}{74D2FA}
\DeclareRobustCommand{\mkBanglaBERT}{\tikz[baseline=-0.6ex]\node[circle,fill=mkBanglaBERTcol,draw=mkBanglaBERTcol!75!black,line width=0.25pt,inner sep=0pt,minimum size=1.15ex]{};}

\DeclareRobustCommand{\mkBanglaBERTbar}{\tikz[baseline=-0.6ex]\draw[mkBanglaBERTcol,line width=2.4pt] (0,0) -- (0.95em,0);}
\definecolor{mkBanglishBERTcol}{HTML}{A7A8F0}
\DeclareRobustCommand{\mkBanglishBERT}{\tikz[baseline=-0.6ex]\node[rectangle,fill=mkBanglishBERTcol,draw=mkBanglishBERTcol!75!black,line width=0.25pt,inner sep=0pt,minimum size=1.0ex]{};}

\DeclareRobustCommand{\mkBanglishBERTbar}{\tikz[baseline=-0.6ex]\draw[mkBanglishBERTcol,line width=2.4pt] (0,0) -- (0.95em,0);}
\definecolor{mkmBERTcol}{HTML}{8180DE}
\DeclareRobustCommand{\mkmBERT}{\tikz[baseline=-0.6ex]\node[regular polygon,regular polygon sides=3,fill=mkmBERTcol,draw=mkmBERTcol!75!black,line width=0.25pt,inner sep=0pt,minimum size=1.4ex]{};}

\DeclareRobustCommand{\mkmBERTbar}{\tikz[baseline=-0.6ex]\draw[mkmBERTcol,line width=2.4pt] (0,0) -- (0.95em,0);}
\definecolor{mkXLMRcol}{HTML}{7A4DFF}
\DeclareRobustCommand{\mkXLMR}{\tikz[baseline=-0.6ex]\node[diamond,fill=mkXLMRcol,draw=mkXLMRcol!75!black,line width=0.25pt,inner sep=0pt,minimum size=1.3ex]{};}

\DeclareRobustCommand{\mkXLMRbar}{\tikz[baseline=-0.6ex]\draw[mkXLMRcol,line width=2.4pt] (0,0) -- (0.95em,0);}
\definecolor{mkMuRILcol}{HTML}{5445B5}
\DeclareRobustCommand{\mkMuRIL}{\tikz[baseline=-0.6ex]\node[regular polygon,regular polygon sides=3,shape border rotate=180,fill=mkMuRILcol,draw=mkMuRILcol!75!black,line width=0.25pt,inner sep=0pt,minimum size=1.4ex]{};}

\DeclareRobustCommand{\mkMuRILbar}{\tikz[baseline=-0.6ex]\draw[mkMuRILcol,line width=2.4pt] (0,0) -- (0.95em,0);}
\definecolor{mkLaBSEcol}{HTML}{211CBF}
\DeclareRobustCommand{\mkLaBSE}{\tikz[baseline=-0.6ex]\node[regular polygon,regular polygon sides=5,fill=mkLaBSEcol,draw=mkLaBSEcol!75!black,line width=0.25pt,inner sep=0pt,minimum size=1.35ex]{};}

\DeclareRobustCommand{\mkLaBSEbar}{\tikz[baseline=-0.6ex]\draw[mkLaBSEcol,line width=2.4pt] (0,0) -- (0.95em,0);}
\definecolor{mkQwenBcol}{HTML}{EC8DFF}

\DeclareRobustCommand{\encoderkey}{\mkBanglaBERT\,BanglaBERT, \mkBanglishBERT\,BanglishBERT, \mkmBERT\,mBERT, \mkXLMR\,XLM-R, \mkMuRIL\,MuRIL and \mkLaBSE\,LaBSE}
\DeclareRobustCommand{\encodercolors}{\mkBanglaBERTbar\,BanglaBERT, \mkBanglishBERTbar\,BanglishBERT, \mkmBERTbar\,mBERT, \mkXLMRbar\,XLM-R, \mkMuRILbar\,MuRIL and \mkLaBSEbar\,LaBSE}
\definecolor{mkgrey}{HTML}{5B6474}
\DeclareRobustCommand{\mkfilled}{\tikz[baseline=-0.6ex]\node[circle,fill=mkgrey,inner sep=0pt,minimum size=1.1ex]{};}
\DeclareRobustCommand{\mkopen}{\tikz[baseline=-0.6ex]\node[circle,fill=white,draw=mkgrey,line width=0.6pt,inner sep=0pt,minimum size=1.0ex]{};}

\newcommand{\NxPlanBothAcc}{4}
\newcommand{\NxPlanBothBal}{7}
\newcommand{\NxPlanBothFone}{6}
\newcommand{\NxCovBothAccZero}{9}
\newcommand{\NxCovBothBalZero}{12}

\newcommand{\NxCovBothAccOne}{12}
\newcommand{\NxCovBothBalOne}{12}
\newcommand{\NxReproAcc}{0.09}
\newcommand{\NxReproSSI}{0.008}
\newcommand{\NxReproSSIAll}{0.014}
\newcommand{\NxRhoBalPlan}{0.50}
\newcommand{\NxRhoBalCov}{0.87}
\newcommand{\NxRhoBalOne}{0.98}

\newcommand{\NxBalNew}{BanglishBERT F, BanglishBERT S, MuRIL S}
\newcommand{\NxZsBoth}{8}

\newcommand{\NxZsBothSrc}{6}

\newcommand{\NxZsAboveFloor}{8}
\newcommand{\NxZsBothAndFloor}{6}
\newcommand{\NxZsSupInside}{7}
\newcommand{\NxZsSupGap}{$-$0.6 to 3.6}

\newcommand{\NxZsRatioF}{55}
\newcommand{\NxZsRatioS}{86}

\newcommand{\NxZsMlrMedian}{$-$0.05}
\newcommand{\NxZsLeaceMedian}{$-$0.13}
\newcommand{\NxZsAmbMedian}{$-$0.12}
\newcommand{\NxZsMlrRand}{3}
\newcommand{\NxZsLeaceRand}{1}

\newcommand{\NxZsAmbBelowS}{6}
\newcommand{\NxZsAmbSAcc}{$-$9.4 to $-$0.7}
\newcommand{\NxZsAmbSrcSAcc}{$-$26.7 to $-$6.5}

\newcommand{\NxZsMlrBalF}{0.22--0.61}

\newcommand{\NxZsMlrBalRand}{10}

\newcommand{\NxZsGlossCovF}{0.70--1.01}
\newcommand{\NxZsTransCovF}{0.44--0.98}
\newcommand{\NxZsTransAmbF}{$-$0.33 to 0.62}
\newcommand{\NxFBNCOVBoth}{3}
\newcommand{\NxFBNMLRBoth}{2}
\newcommand{\NxDOMCOVBoth}{3}
\newcommand{\NxDOMMLRBoth}{2}

\newcommand{\NxThreeConcepts}{XLM-R, MuRIL, LaBSE}

\newcommand{\NxFBNAmiF}{0.19}
\newcommand{\NxDOMAmiF}{0.18}
\newcommand{\NxFBNAri}{0.42--0.53}
\newcommand{\NxDOMCoverage}{40}
\newcommand{\NxDOMMajority}{32}
\newcommand{\NxFBNRandMean}{0.33--0.63}
\newcommand{\NxFBNCovSSI}{0.42--0.91}

\newcommand{\NxGlossFBN}{11}
\newcommand{\NxGlossFBNOf}{12}
\newcommand{\NxGlossFBNpp}{5.1--16.4}
\newcommand{\NxGlossDOM}{5}
\newcommand{\NxGlossDOMPos}{12}

\newcommand{\NxLeaceDeficit}{$-$11.2 to 0.6}
\newcommand{\NxLeaceDeficitNeg}{23}
\newcommand{\NxLeaceDeficitCI}{21}
\newcommand{\NxLeaceDeficitOf}{24}
\newcommand{\NxLeaceLWShift}{0.8}
\newcommand{\NxLeaceLWShiftMax}{3.6}
\newcommand{\NxLeaceDistZero}{213.4}
\newcommand{\NxLeaceDistLW}{4.3}
\newcommand{\NxLeaceDistHundred}{0.10}
\newcommand{\NxLeaceDeficitHundred}{$-$1.4 to 1.2}
\newcommand{\NxSigmaQuarter}{$-$8.2}
\newcommand{\NxSigmaFull}{$-$6.8}
\newcommand{\NxSigmaExt}{$-$6.0}
\newcommand{\NxSigmaShrink}{17}
\newcommand{\NxProbeCSpread}{0.4}
\newcommand{\NxWoodbury}{$5.6\times10^{-8}$}
\newcommand{\NxIdempotence}{$10^{-11}$}
\newcommand{\NxProbeCSpreadMax}{1.8}

\newcommand{\NxEnglishEraseShareF}{0.44--0.82}
\newcommand{\NxCrossEraseShareF}{$-$0.05 to 0.07}
\newcommand{\NxLeaceCrossWorse}{23}
\newcommand{\NxLeaceCrossOf}{24}

\newcommand{\NxLeaceNativeBetter}{13}

\newcommand{\NxPoolN}{2,166}
\newcommand{\NxPoolTokDiff}{0.2}
\newcommand{\NxPoolCos}{0.9966}
\newcommand{\NxPoolCtxCovF}{$-$0.2 to 1.2}
\newcommand{\NxPoolCtxCovFSig}{0}
\newcommand{\NxPoolCtxAmbF}{0.4--4.9}
\newcommand{\NxPoolCtxAmbFSig}{4}
\newcommand{\NxPoolSpanCovF}{$-$3.6 to 1.9}

\newcommand{\NxPoolSpanAmbF}{$-$5.2 to $-$1.0}

\newcommand{\NxPoolSlotCovS}{2.0--9.6}
\newcommand{\NxPoolSlotCovSSig}{5}
\newcommand{\NxMlpGainF}{2.1--5.9}
\newcommand{\NxMlpSig}{3}
\newcommand{\NxMlpCells}{36}
\newcommand{\NxMlpCovDiff}{0.04}
\newcommand{\NxMlpLinRepro}{0.04}
\newcommand{\NxMdlPrior}{1.00--1.27}
\newcommand{\NxMdlPriorAbove}{47}
\newcommand{\NxMdlCells}{48}
\newcommand{\NxMdlUniform}{1.04--1.61}
\newcommand{\NxBitsCovF}{0.24--0.77}
\newcommand{\NxBitsCovS}{0.15--0.95}
\newcommand{\NxBitsLeace}{$-$0.32 to 0.19}
\newcommand{\NxBitsRho}{0.84}
\newcommand{\NxRetTransBefore}{0.13--0.32}
\newcommand{\NxRetTransAfter}{0.27--0.64}

\newcommand{\NxRetGlossBefore}{0.10--0.12}
\newcommand{\NxRetGlossAfter}{0.20--0.43}
\newcommand{\NxRetIdiomBefore}{0.00--0.14}
\newcommand{\NxRetIdiomAfter}{0.01--0.11}
\newcommand{\NxRetIdiomMax}{0.15}
\newcommand{\NxAlignDimsBefore}{1.2--3.6}
\newcommand{\NxAlignDimsAfter}{4.8--7.4}
\newcommand{\NxAlignSSIBefore}{0.31--0.82}
\newcommand{\NxAlignSSIAfter}{0.83--1.05}
\newcommand{\NxAlignEraseF}{0.17--0.63}
\newcommand{\NxAlignEraseS}{0.33--0.63}

\newcommand{\NxAlignOK}{10}
\newcommand{\NxAlignCells}{12}
\newcommand{\NxAlignFail}{BanglaBERT S, BanglishBERT S}
\newcommand{\NxAlignFailGap}{0.02--0.10}
\newcommand{\NxAlignReducedSSI}{0.70--0.86}
\newcommand{\NxOverlapZsRho}{0.67}
\newcommand{\NxOverlapSupRho}{0.71}
\newcommand{\NxOverlapZsRhoAll}{0.63}
\newcommand{\NxOverlapZsCells}{12}
\newcommand{\NxAlignZsCovF}{0.26--0.90}
\newcommand{\NxAlignZsCovS}{0.58--1.02}
\newcommand{\NxAlignZsCovUp}{11}
\newcommand{\NxAlignZsLeace}{$-$0.61 to 0.27}
\newcommand{\NxAlignZsLeaceMedian}{$-$0.11}

\newcommand{\NxAlignEraseMlrMax}{0.8}
\newcommand{\NxAlignFitSpace}{0.22--0.39}
\newcommand{\NxAlignFitTest}{0.04--0.15}
\newcommand{\NxAlignFitTestBefore}{0.04--0.16}
\newcommand{\NxLayerRand}{61}
\newcommand{\NxLayerBoth}{50}
\newcommand{\NxLayerCells}{64}
\newcommand{\NxLayerCovMlrPos}{68}
\newcommand{\NxLayerCovMlrSig}{49}
\newcommand{\NxLayerCovMlrNeg}{2}
\newcommand{\NxLayerCovMlrCells}{76}
\newcommand{\NxLayerBelowMedian}{5}
\newcommand{\NxLayerSelCells}{12}
\newcommand{\NxLayerXLMRSel}{0.31}
\newcommand{\NxDepthMlrRho}{0.83}
\newcommand{\NxDepthMlrTop}{0.48 and 0.50}
\newcommand{\NxDepthMlrRest}{0.10--0.44}
\newcommand{\NxDepthCovRho}{0.37}
\newcommand{\NxDepthCovHigher}{3 of the 4}
\newcommand{\NxDepthCovBoth}{5}
\newcommand{\NxDepthXLMRCov}{0.73}
\newcommand{\NxLayerXLMRMax}{0.80}
\newcommand{\NxPlanNativeGain}{4.6--15.2}
\newcommand{\NxPlanCrossGain}{1.1--7.3}
\newcommand{\NxPlanPassGainF}{3.0--7.3}
\newcommand{\NxPlanFloorF}{15.7}

\newcommand{\NyAlignSSIAfter}{0.64--0.89}
\newcommand{\NyAlignEraseF}{$-$0.04 to 0.28}

\newcommand{\NyAlignOK}{12}
\newcommand{\NyAlignCells}{12}
\newcommand{\NyAlignPairs}{3,092}
\newcommand{\NyAlignZsCovF}{0.25--0.82}
\newcommand{\NyAlignZsLeaceMedian}{$-$0.63}
\newcommand{\NyAlignSpaceDimsF}{5.6}
\newcommand{\NyAlignSpaceReadF}{0.93}
\newcommand{\NyAlignSpaceEraseF}{0.44}
\newcommand{\NyAlignPVDimsF}{3.2}
\newcommand{\NyAlignPVReadF}{0.68}
\newcommand{\NyAlignPVEraseF}{0.17}
\newcommand{\NyAlignValDimsF}{2.3}

\newcommand{\NyAlignShufDimsF}{0.7}
\newcommand{\NyAlignShufReadF}{0.18}
\newcommand{\NyAlignShufEraseF}{0.01}
\newcommand{\NyAlignNoneDimsF}{2.0}
\newcommand{\NyAlignKeepDims}{35}
\newcommand{\NyAlignKeepRead}{51}

\newcommand{\NyInlpGuarded}{20}
\newcommand{\NyInlpFits}{24}
\newcommand{\NyInlpRankF}{231}
\newcommand{\NyInlpRankRangeF}{165--264}
\newcommand{\NyInlpRankShareF}{26}
\newcommand{\NyInlpDropF}{1.6}
\newcommand{\NyInlpDropMaxF}{2.7}
\newcommand{\NyInlpNativeF}{8.6}
\newcommand{\NyInlpRandDropF}{0.3}

\newcommand{\NyInlpOmegaF}{0.25}
\newcommand{\NyInlpChanceF}{0.26}
\newcommand{\NyInlpOmegaEnF}{0.77}

\newcommand{\NyInlpDropS}{1.4}
\newcommand{\NyInlpNativeS}{7.3}

\newcommand{\NyInlpOmegaS}{0.28}
\newcommand{\NyInlpChanceS}{0.15}
\newcommand{\NyInlpPropOK}{10}
\newcommand{\NyInlpPropCells}{11}
\newcommand{\NyRlaceGuarded}{5}
\newcommand{\NyRlaceFits}{24}
\newcommand{\NyRlaceEnglishShareF}{0.01}
\newcommand{\NyRlaceOmegaEnF}{0.02}

\newcommand{\NyMlpShareF}{0.11--0.15}
\newcommand{\NyLinShareF}{0.02--0.14}
\newcommand{\NyMlpLeftF}{5.5--5.6}
\newcommand{\NyLinLeftF}{0.5--2.5}

\newcommand{\NyRotCovEraseF}{0.17}
\newcommand{\NyRotLeaceEraseF}{0.18}
\newcommand{\NyRotInlpEraseF}{0.10}

\newcommand{\NyTagCoverage}{89}
\newcommand{\NyTagAmiF}{0.19}
\newcommand{\NyTagAri}{0.60--0.61}
\newcommand{\NyReseedAri}{0.48--0.63}

\newcommand{\NyTaxCovAcc}{19}
\newcommand{\NyTaxCells}{24}
\newcommand{\NyTaxMlrAcc}{2, 4, 1}

\newcommand{\NyTaxMlrTag}{1}
\newcommand{\NyTaxMlrXLMR}{3}
\newcommand{\NyTaxMlrMuRIL}{2}
\newcommand{\NyTaxMlrLaBSE}{1}
\newcommand{\NyGlossCovSig}{24}
\newcommand{\NyGlossMlrSig}{23}
\newcommand{\NyGlossCells}{24}
\newcommand{\NyGlossCovReseed}{6.0--7.0}
\newcommand{\NyGlossCovTag}{5.1}
\newcommand{\NyGlossMlrReseed}{13.8--15.7}
\newcommand{\NyGlossMlrTag}{4.2}
\newcommand{\NyPathZsDefaultF}{0.02}
\newcommand{\NyPathZsBestF}{0.42}

\newcommand{\NyPathDimsDefaultF}{0.6}
\newcommand{\NyPathDimsBestF}{7.5}
\newcommand{\NyPathDimsLimitF}{10.3}

\newcommand{\NyPathRho}{0.57--0.79}
\newcommand{\NySelMlr}{22}
\newcommand{\NySelMlrOf}{24}

\newcommand{\NySelLeace}{23}
\newcommand{\NySelLeaceOf}{24}

\newcommand{\NySelMlrSmallC}{23}
\newcommand{\NySelLeaceRidge}{22}
\newcommand{\NyPathEnValGainF}{3.0}
\newcommand{\NyIsoCovModels}{MuRIL and LaBSE}

\newcommand{\NyIsoCovBalF}{4}

\newcommand{\NyIsoCovBalS}{4}

\newcommand{\NyIsoMlrModelsF}{MuRIL}
\newcommand{\NyIsoMlrBalF}{3}
\newcommand{\NyIsoCovTopSSI}{1.05--1.13}
\newcommand{\NyIsoZsTop}{0.93--1.15}
\newcommand{\NyIsoZsBoth}{5}

\newcommand{\NyCtxNativeLossF}{7.8}
\newcommand{\NyCtxCrossLossF}{2.6}

\newcommand{\NyIsoGainIdiomF}{8.4}
\newcommand{\NyIsoGainCtxF}{6.7}
\newcommand{\NyIsoNativeIdiomF}{9.1}
\newcommand{\NyIsoNativeCtxF}{13.1}

\newcommand{\NzProPass}{10}
\newcommand{\NzProEncoders}{five}
\newcommand{\NzProFail}{mBERT}
\newcommand{\NzProSsiMedian}{1.03}

\newcommand{\NzProGain}{\mbox{5.8--18.0}}
\newcommand{\NzProAccPass}{9}

\newcommand{\NzProZsPass}{11}
\newcommand{\NzProZsPassAcc}{10}
\newcommand{\NzProZsMedianF}{0.86}
\newcommand{\NzPlanHash}{e423d9f21f2f}
\newcommand{\NzCovRandcovFigAcc}{12}
\newcommand{\NzCovRandcovFigBal}{12}
\newcommand{\NzMlrRandcovFigAcc}{3}
\newcommand{\NzMlrRandcovFigBal}{4}
\newcommand{\NzCovPcaFigBal}{12}
\newcommand{\NzCovPcaFigAcc}{10}
\newcommand{\NzMlrPcaFigAcc}{2}
\newcommand{\NzMlrPcaFigBal}{4}
\newcommand{\NzCovRandcovIdiomAcc}{7}
\newcommand{\NzCovRandcovIdiomBal}{8}

\newcommand{\NzMlrRandcovIdiomBal}{1}
\newcommand{\NzCovPcaIdiomBal}{9}

\newcommand{\NzMlrPcaIdiomBal}{1}
\newcommand{\NzCovPcaRange}{\mbox{1.5--10.9}}
\newcommand{\NzPlannedSurvive}{three}
\newcommand{\NzPlannedRejections}{four}
\newcommand{\NzPlannedLost}{LaBSE's F test}
\newcommand{\NzMlrPcaBanglaBERT}{\ensuremath{-}4.2}
\newcommand{\NzOpaqueBelowFig}{12}
\newcommand{\NzTranspAboveFig}{11}

\newcommand{\NzOpaqueBelowAllFig}{9}
\newcommand{\NzOpaqueDropFig}{0.17}
\newcommand{\NzOpaqueBelowIdiom}{9}
\newcommand{\NzTranspAboveIdiom}{11}
\newcommand{\NzCellsIdiom}{11}
\newcommand{\NzOpaqueBelowAllIdiom}{8}

\newcommand{\NzSubsetSize}{1,367}
\newcommand{\NzSubsetDraws}{20}
\newcommand{\NzRefitWidth}{1.8}
\newcommand{\NzRefitCovAcc}{7}
\newcommand{\NzCondCovAcc}{10}
\newcommand{\NzRefitCovBal}{12}
\newcommand{\NzCondCovBal}{12}
\newcommand{\NzRefitMlrAcc}{4}
\newcommand{\NzCondMlrAcc}{9}
\newcommand{\NzRefitMlrBal}{11}
\newcommand{\NzCondMlrBal}{12}
\newcommand{\NzRefitDraws}{200}
\newcommand{\NzRefitAgree}{98.8}
\newcommand{\NzEvalIdiom}{\ensuremath{-}3.4}
\newcommand{\NzEvalIdiomCI}{\mbox{[\ensuremath{-}7.4, +0.7]}}
\newcommand{\NzEvalIdiomP}{.12}

\newcommand{\NzIdiomCpIdiomCI}{\mbox{[\ensuremath{-}10.6, \ensuremath{-}3.7]}}
\newcommand{\NzIdiomCpIdiomHolm}{.001}
\newcommand{\NzIdiomCpIdiomAbs}{7.3}

\newcommand{\NzIdiomCpNativeIdiom}{\ensuremath{-}6.7}
\newcommand{\NzIdiomCpNativeIdiomHolm}{.11}
\newcommand{\NzEvalFig}{+0.5}
\newcommand{\NzEvalFigCI}{\mbox{[\ensuremath{-}4.6, +5.9]}}
\newcommand{\NzEvalFigP}{.87}

\newcommand{\NzIdiomCpFigCI}{\mbox{[\ensuremath{-}11.1, \ensuremath{-}2.7]}}
\newcommand{\NzIdiomCpFigHolm}{.011}
\newcommand{\NzIdiomCpFigAbs}{6.9}
\newcommand{\NzAptNativeFig}{+15.6}
\newcommand{\NzAptNativeFigCI}{\mbox{[+8.8, +22.3]}}
\newcommand{\NzAptCrossFig}{+1.9}
\newcommand{\NzIdiomCpNativeFig}{\ensuremath{-}7.9}
\newcommand{\NzIdiomCpNativeFigHolm}{.08}
\newcommand{\NzIdiomCpZs}{\ensuremath{-}2.7}
\newcommand{\NzIdiomCpMlr}{\ensuremath{-}3.3}

\newcommand{\NzTranspAuc}{0.75}
\newcommand{\NzOpaqueIdiomShare}{83}

\newcommand{\HvKappaClass}{0.51}

\newcommand{\HvAgreeClass}{68}

\newcommand{\HvKappaFunction}{0.63}

\newcommand{\HvAgreeFunction}{81}

\newcommand{\HvKappaType}{0.74}

\newcommand{\HvAgreeType}{88}

\newcommand{\HvKappaEquiv}{0.40}

\newcommand{\HvAgreeEquiv}{75}
\newcommand{\HvAcOneEquiv}{0.71}
\newcommand{\HvKappaRange}{\mbox{0.40--0.74}}

\newcommand{\HvPoorBoth}{22}
\newcommand{\HvAptBoth}{63}

\newcommand{\HvReferentialBoth}{52}
\newcommand{\HvEvaluativeBoth}{30}
\newcommand{\HvIdiomBoth}{65}
\newcommand{\HvParaphraseBoth}{20}
\newcommand{\HvProverbBoth}{3}
\newcommand{\HvEquivBoth}{63}
\newcommand{\HvNotEquivBoth}{1}
\newcommand{\HvAtLeastPartly}{95--97}

\newcommand{\HvEvalInsult}{82}
\newcommand{\HvEvalFolly}{96}
\newcommand{\HvEvalPeople}{72}
\newcommand{\HvEvalPeopleTransfer}{0.91}
\newcommand{\HvRhoEvalTransfer}{\ensuremath{-}0.54}
\newcommand{\HvPEvalTransfer}{.07}
\newcommand{\HvRhoEvalTransferCore}{\ensuremath{-}0.40}

\newcommand{\HvRhoGoodTransfer}{0.33}
\newcommand{\HvPGoodTransfer}{.30}

\newcommand{\HvRhoRatioNative}{0.87}
\newcommand{\HvRhoEvalNative}{\ensuremath{-}0.61}
\newcommand{\HvPartialEval}{\ensuremath{-}0.02}
\newcommand{\HvTravelTop}{\mbox{56--91}}
\newcommand{\HvTravelBottom}{31}

\newcommand{\authormark}[1]{\textsuperscript{\normalfont #1}}
\newcommand{\authornote}[2]{\begingroup\renewcommand{\thefootnote}{#1}\footnotetext[0]{#2}\endgroup}

\title{What Travels Across Languages?\\Reading, Transfer and Erasure of Cross-Lingual\\Concept Subspaces in Bangla Idioms}

\author{\name Syed Rifat Raiyan\authormark{\textdagger\textdaggerdbl} \email rifatraiyan@iut-dhaka.edu \\
      \addr Systems and Software Lab (SSL)\\
      Department of Computer Science and Engineering\\
      Islamic University of Technology, Gazipur, Bangladesh
      \AND
      \name Yousuf Ibne Kamal Niloy\authormark{\textdagger} \email yousufniloy@iut-dhaka.edu \\
      \addr Systems and Software Lab (SSL)\\
      Department of Computer Science and Engineering\\
      Islamic University of Technology, Gazipur, Bangladesh}

\hypersetup{pdfauthor={Syed Rifat Raiyan and Yousuf Ibne Kamal Niloy},pdftitle={What Travels Across Languages? Reading, Transfer and Erasure of Cross-Lingual Concept Subspaces in Bangla Idioms}}

\begin{document}

\maketitle
\authornote{\textdagger}{Equal contribution.}
\authornote{\textdaggerdbl}{Corresponding author.}

\begin{abstract}
Cross-lingual concept analysis requires distinguishing the information that a subspace retains from the classifiers and erasers that transfer through it. We study these operations in six frozen encoders using 10,359 Bangla (Bengali) idiom senses, eight input conditions, and figurative-class and sentiment labels. Class-mean directions provide the most consistent evidence of shared coarse meaning: in \textit{post hoc} analyses, they outperform random-subspace, permuted-label and covariance-matched nulls in all 12 model--label tests under balanced accuracy. A prospective internal re-split supports idiom-alone sharing in 10 of 12 tests, spanning five encoders. The original confirmatory logistic-regression family rejects both planned nulls in 4 of 12 tests, where English directions gain only a few accuracy points above the majority-class floor (3.0--7.3). Classifier transfer depends on the input map and regularization: English-trained class-mean readouts beat both zero-shot nulls in 8 of 12 discovery tests, and stronger source-selected penalties improve transfer through logistic and LEACE directions. These readouts use Bangla-validation-selected layers and unlabeled target preprocessing. Input controls show greater coarse-class sharing for explicit glosses than for idiom counterparts. Reading, however, does not imply effective erasure: English-fitted LEACE removes less than one percentage point of Bangla figurative-class accuracy, compared with 4--11 points for Bangla-fitted LEACE; INLP shows a similar asymmetry. Pattern--filter analysis, covariance-weighted input maps and a binary Gaussian overlap bound explain how these operations can diverge. The analogous multiclass ordering is an empirical diagnostic, observed in 11 of 12 cells. Together, the results support an estimator-sensitive account of cross-lingual concept sharing and motivate evaluating reading, classifier transfer, erasure and overlap separately. Code, frozen analysis plans and aggregate results are available at \coderepo.
\end{abstract}

\section{Introduction}
\label{sec:intro}

Concept subspaces underpin probes and interventions that read or remove information from model representations \citep{ravfogel-etal-2020-null,belrose-etal-2023-leace,naowarat-etal-2026-framework}. Multilingual encoders align translations \citep{pires-etal-2019-multilingual,wu-dredze-2019-beto,conneau-etal-2020-emerging} and contain both language-sensitive and shared components \citep{libovicky-etal-2020-language,chang-etal-2022-geometry,xie-etal-2022-discovering}. This motivates transferring English-fitted tools to other languages: difference-of-means probes \citep{marks-tegmark-2024-geometry}, steering vectors \citep{rimsky-etal-2024-steering}, and erasers such as iterative nullspace projection (INLP), linear adversarial concept erasure (R-LACE) and least-squares concept erasure (LEACE) \citep{ravfogel-etal-2020-null,ravfogel-etal-2022-linear,belrose-etal-2023-leace}. Their operations, however, need not transfer together.

\begin{figure}[!t]
\centering
\includegraphics[width=0.72\linewidth]{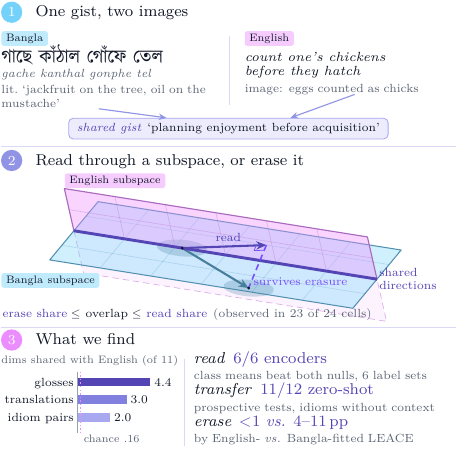}
\caption{\textbf{Reading, transfer and erasure need not travel together.} (1)~The Bangla \bn{gache} (\bnrom{gache}, lit.\ \bnlit{gache}) and English \textit{count one's chickens before they hatch} share a figurative gist but no image. (2)~English concept directions (pink) overlap partly with Bangla class-mean directions (cyan): projection can retain a class difference while erasure leaves another component. The displayed ordering is proved for binary or pairwise-averaged shares (\cref{prop:read-erase}); its multiclass counterpart is an empirical diagnostic (\ThBothOK{} of \ThCells{} cells before a held-out rotation, \NyAlignOK{} after). (3)~The main findings: class-mean reading across encoders, prospective classifier transfer on idioms without context, and limited English-fitted erasure (points of F accuracy). Shared dimensions are medians across encoders; the six label sets are F, S, three F reseeds and TAG. The prospective test re-splits the same corpus.}
\label{fig:teaser}
\end{figure}

Two precedents frame our question. \citet{gonen-etal-2022-analyzing} established cross-lingual reading/removal dissociation and related it to partially shared gender directions. \citet{naowarat-etal-2026-framework} established that retention and leakage depend on the subspace estimator. We connect these two lines of work by separating the subspace image, the map that reads inputs into it, and the classifier trained on those coordinates. Across idiom, gloss, translation and context conditions, we test how estimator choice and regularization affect reading, classifier transfer and erasure.

Idioms provide a test of meaning shared across different forms. Bangla (Bengali) speakers mock the man who oils his mustache while the jackfruit still hangs on the tree (\bn{gache}, \bnrom{gache}); English speakers warn him not to count his chickens before they hatch. The phrases share a figurative gist despite different images. Because idiom meanings are conventional rather than fully compositional \citep{nunberg-etal-1994-idioms}, comparing idioms with glosses, translations and context helps distinguish coarse semantic sharing from properties of their inputs.

We estimate English concept subspaces and test three operations on Bangla (\cref{fig:teaser}): \emph{reading} with a Bangla-trained probe, \emph{transfer} with an English-trained classifier, and \emph{erasure} by removing the English-fitted component. Our shared-subspace index (SSI) expresses each readout as a share of native above-floor accuracy. The study covers 10,359 Bagdhara idiom senses \citep{sakhawat-etal-2026-words}, eight input conditions, figurative and sentiment labels with alternative taxonomies, six encoders and seven estimators, with null controls, two prospective internal replications and a bilingual human check (\cref{sec:method,sec:setup,app:method}). \par\Needspace{10\baselineskip}Four questions guide us:

\begin{enumerate}[label=\textbf{RQ\textsubscript{\arabic*}},leftmargin=*,itemsep=1pt,topsep=3pt]
\item Do English idiom directions retain Bangla figurative distinctions beyond random and permuted-label controls? (\cref{sec:rq1})
\item Which inputs carry the shared signal: glosses, translations, idioms or context? (\cref{sec:rq2})
\item In what form is meaning shared, and does an English classifier transfer? (\cref{sec:rq3})
\item Does erasing an English subspace remove the Bangla signal it reads? (\cref{sec:rq4})
\end{enumerate}

The study reports three findings:

\begin{itemize}[leftmargin=*,itemsep=1pt,topsep=3pt]
\item \textbf{Robust class-mean sharing across input controls.} Class-mean directions beat three nulls in all six encoders under balanced accuracy, survive taxonomy reseeds, and retain idiom-alone signal in five encoders in a prospective internal test. Glosses share more coarse-class information than idiom counterparts (\cref{sec:rq1,sec:rq2}).
\item \textbf{Transfer depends on the input map and penalty.} COV and LEACE share a class-mean image but expose different coordinates to a probe. Class-mean readouts transfer more consistently than default filters, while source-selected regularization substantially improves filter transfer. Strong-penalty and covariance-weighting analyses explain mechanisms that can produce these differences (\cref{sec:rq3}).
\item \textbf{Reading and erasure require different geometric conditions.} English-fitted removal preserves most Bangla class information despite successful reading. A binary and pairwise-averaged overlap bound formalizes this distinction, and separate multiclass measurements characterize it empirically (\cref{sec:rq4}).
\end{itemize}

\section{Related Work}
\label{sec:related}

\paragraph{Concept subspaces, probing and erasure.}
Linear decodability depends on the probe and its controls \citep{belinkov-2022-probing,hewitt-liang-2019-designing,voita-titov-2020-information}. Concept activation vectors and the linear representation hypothesis connect directions to model sensitivity and steering \citep{kim-etal-2018-tcav,park-etal-2024-linear}; erasure tests the effect of removing information \citep{elazar-etal-2021-amnesic}. \citet{naowarat-etal-2026-framework} compare five supervised estimators through retention, leakage and disentanglement measures. We adopt this estimator-comparison perspective for cross-lingual reading, transfer and erasure. The filter--pattern distinction is also established: decoder weights need not follow class differences \citep{haufe-etal-2014-interpretation}, and mean-based probes can generalize better \citep{marks-tegmark-2024-geometry}. Our contribution is to connect cross-language estimator differences to input maps and penalty choice, using Aitken's theorem \citep{aitken-1936-least} to characterize covariance-dependent optimality.

\paragraph{Shared multilingual representations.}
Language-agnostic components also appear in decoder LLMs \citep{wendler-etal-2024-llamas,dumas-etal-2025-separating}. Alignment uses orthogonal maps \citep{schonemann-1966-generalized}, hubness-corrected retrieval \citep{conneau-etal-2018-word,artetxe-etal-2018-robust} and contextual anchors \citep{schuster-etal-2019-cross,aldarmaki-diab-2019-context,cao-etal-2020-multilingual}; SSI instead measures shared class information. Cross-lingual debiasing \citep{reusens-etal-2023-investigating} directly motivates transfer tests. In particular, \citet{gonen-etal-2022-analyzing} show that mBERT gender classification transfers while INLP removal does not, and analyze shared versus language-specific directions. We extend that question to idioms, comparing estimators, matched input conditions and overlap with both target-trained reading and English-trained classifier transfer.

\paragraph{Idioms in language models.}
Human idiom comprehension depends on configuration and context \citep{cacciari-tabossi-1988-comprehension,gibbs-etal-1989-kick}. Models show partial idiomaticity encoding \citep{tayyar-madabushi-etal-2021-astitchinlanguagemodels,garcia-etal-2021-probing,he-etal-2025-investigating}, overly literal translation \citep{dankers-etal-2022-transformer,liu-etal-2023-crossing}, memorization \citep{haviv-etal-2023-understanding} and competition between figurative and literal readings \citep{oh-etal-2026-tug}. For Bangla idioms, no single LLM leads across paraphrasing, span detection and meaning identification \citep{akter-etal-2026-extent}. Models that score candidate readings are drawn to an idiom's literal gloss when it is offered, more strongly in Bangla than in English on the same items \citep{sakhawat-raiyan-2026-literal}. Complementing behavioral multilingual studies \citep{tedeschi-etal-2022-id10m,khoshtab-etal-2025-comparative}, we use idiom, gloss and context controls to ask which coarse meaning distinctions are shared in representations, and which operations preserve them across languages.

\section{Method}
\label{sec:method}

We distinguish the image estimated from labeled inputs, the map used to read those inputs, and the operations evaluated across languages. \Cref{app:method} provides the algorithms and \cref{tab:notation} collects the notation.

\subsection{Concept subspaces: patterns and filters}
\label{sec:subspaces}

\para{Inputs and splits.} Each input is encoded on its own, mean-pooled over its target tokens into $x\in\R^D$ and centered on its language's mean over SPACE, the estimation split. Records are linked into connected components, and whole components are assigned to four disjoint splits (\cref{sec:data}): SPACE estimates subspaces, PROBE trains readouts, VALIDATION selects layers and, where stated, penalties, and TEST scores every readout.

\para{Estimators.} An estimator maps the centered SPACE inputs of one input condition and their labels ($K$ classes) to a rank-$M$ \emph{image} (the subspace) with orthonormal basis $Q\in\R^{D\times M}$ and an \emph{input map} $G\in\R^{M\times D}$ with $GQ=I_M$, which reads inputs into it. Probes read the compact coordinates $z=Gx$, and erasure removes $P=QG$. Unless stated otherwise $M=K-1$, the largest rank the class means support: 11 for twelve figurative classes and 3 for four sentiment labels (\cref{fig:app-rank} varies the rank).

\begin{figure}[t]
\centering
\begin{subfigure}[t]{0.5274\textwidth}
\centering
\includegraphics[width=\linewidth,viewport=18.03 0.00 231.93 187.82,clip]{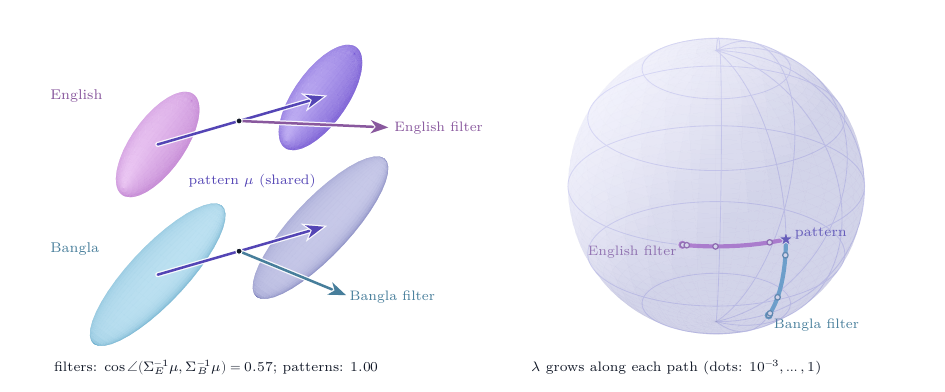}
\caption{shared means, language-specific covariance}\label{fig:app-theory-patterns-a}
\end{subfigure}\hfill
\begin{subfigure}[t]{0.4526\textwidth}
\centering
\includegraphics[width=\linewidth,viewport=249.52 0.00 433.12 187.82,clip]{figures/app_theory_patterns.pdf}
\caption{the ridge path $(\Sigma+\lambda I)^{-1}\mu$ on the unit sphere}\label{fig:app-theory-patterns-b}
\end{subfigure}
\caption{\textbf{Patterns are shared where filters are not.} (a)~Two languages with the same class-mean difference $\mu$ (purple arrows; ellipsoids: English classes pink and violet, Bangla cyan and lavender) but differently oriented within-class covariance. Their patterns coincide; their filters $\Sigma_E^{-1}\mu$ and $\Sigma_B^{-1}\mu$ (mauve and steel-blue arrows, drawn from the midpoint of the class means) point in different directions. (b)~Each language's ridge path $(\Sigma+\lambda I)^{-1}\mu$ on the unit sphere: weak penalties give its own filter, strong ones the shared pattern (star), the ridge counterpart of \cref{prop:penalty}; dots mark $\lambda=10^{-3},10^{-2},10^{-1},1$. Schematic, in three dimensions; measured counterpart: \cref{fig:estimators}e.}
\label{fig:app-theory-patterns}
\end{figure}

\para{Patterns and filters.} Under a shared-covariance Gaussian model, classes differ along \emph{patterns}, the class-mean offsets $\mu_k$ from the grand mean, whereas discriminant weights follow \emph{filters} $\Sigma^{-1}\mu_k$. A filter discounts directions that are noisy in its fitting language, so patterns can survive a change of language where filters need not (\cref{fig:app-theory-patterns}). The feature--label cross-covariance estimator (COV) uses the left singular vectors of $\Sigma_{xz}=[\pi_k\mu_k]_k$, whose range is $\spn\{\mu_k\}$. Class-centroid principal component analysis (CPCA) estimates the same pattern image at full rank; both use $G=Q^{\top}$. Multinomial logistic regression (MLR) uses the row space of logistic weights at the framework's default inverse penalty $C=1$, while linear discriminant analysis (LDA) supplies a second filter estimator. LEACE pairs COV's image at full rank $M=K-1$ with the covariance-weighted map $G_\Sigma=(Q^{\top}\Sigma^{-1}Q)^{-1}Q^{\top}\Sigma^{-1}$. Random subspaces (RAND) and principal component analysis (PCA) are label-free references (\cref{tab:estimators-glance}).

\begin{table}[!htb]
\caption{\textbf{The seven estimators at a glance.} Each estimator maps centered SPACE inputs and labels to a rank-$M$ subspace, its \emph{image}; a probe then reads the inputs through an \emph{input map} onto that image. Patterns and filters differ in the image. COV and LEACE share the image at $M=K{-}1$ but not the input map (\cref{eq:operator}), so their probes see different metrics (\cref{lem:metric}) and transfer differently (\cref{prop:portable}).}
\label{tab:estimators-glance}
\centering
\small
\setlength{\tabcolsep}{3.2pt}
\begin{tabular}{@{}ll>{\raggedright\arraybackslash}m{4.15cm}l>{\raggedright\arraybackslash}m{5.25cm}@{}}
\toprule
\textbf{Estimator} & \textbf{Kind} & \textbf{Image} & \textbf{Input map} & \textbf{Role here} \\
\midrule
RAND & null & random orthonormal directions & orthogonal & RAND null; chance overlap (\cref{prop:chance}) \\
PCA & unsupervised & leading principal components & orthogonal & label-free baseline \\
MLR & filter & logistic-regression weight rows & orthogonal & planned estimator; tends to COV as $C\to0$ (\cref{prop:penalty}) \\
LDA & filter & discriminant directions $S_w^{-1/2}U$ & orthogonal & a second filter \\
CPCA & pattern & principal components of class centroids & orthogonal & check on COV; same image at $M=K{-}1$ \\
COV & pattern & class-mean directions, from $\Sigma_{xz}$ & orthogonal & pattern estimator; \hyp{1}--\hyp{3}; portable (\cref{prop:portable}) \\
LEACE & eraser & class-mean directions (COV's at $M=K{-}1$) & oblique, through $\Sigma$ & eraser; \hyp{4}, \expt{3}; generalized least squares \\
\bottomrule
\end{tabular}
\end{table}

\para{The map matters as well as the image.} COV and LEACE span the same subspace at $M=K{-}1$, but COV reads it by orthogonal projection and LEACE through an oblique map that depends on the covariance it was fitted with. An L2 probe on $z=Gx$ searches $\operatorname{range}(G^{\top})$ under the penalty $w^{\top}(G^{\top}G)^{\dagger}w$ (\cref{lem:metric}). Two estimators with one image can therefore differ in what a probe reads, what transfers and what is erased (\cref{app:method-estimators}).

\subsection{Reading, transfer and erasure}
\label{sec:operations}

Let $e$ be the source language (English) and $b$ the target (Bangla), and let $(Q_e,G_e)$ and $(Q_b,G_b)$ be one estimator fitted on the inputs of each.

\para{Read.} To \emph{read}, an L2 logistic probe is trained on $G_ex$ for labeled target-language PROBE items and scored on target TEST items, giving $R_{e\to b}$. With $R_0$ the TEST accuracy of the PROBE majority class, the \emph{shared-subspace index}
\begin{equation}
\ssi_{e\to b} = \frac{R_{e\to b}-R_0}{R_{b\to b}-R_0}
\label{eq:ssi}
\end{equation}
is the share of native above-floor accuracy that source directions recover (undefined if the denominator is at most 2\pp; it can be negative or exceed one). Its native reference $R_{b\to b}$ reads Bangla's own subspace from the same estimator, not the full space, which it can exceed (\cref{tab:app-estimators}); estimators, whose SSIs thus differ in denominator, are compared by above-floor accuracy. When we set reading beside erasure, we call the SSI of a COV image its \emph{read share}.

\para{Transfer.} A Bangla-trained probe can recalibrate whatever directions it is given; an English classifier cannot. To \emph{transfer}, the probe is trained on English PROBE items, with its penalty tuned on English VALIDATION, and applied unchanged to Bangla TEST items. No Bangla label enters readout training or penalty selection, but the layer was selected using Bangla validation labels (\cref{sec:models}). We call this a zero-shot readout conditional on that layer, rather than a target-label-free pipeline. The zero-shot SSI keeps the denominator of \cref{eq:ssi}, so the two indices share a scale.

\para{Erase.} To \emph{erase}, we remove $P_e=Q_eG_e$ or Bangla's own $P_b$, retrain the Bangla probe and compare the accuracy each removal costs. With $A$ the probe's Bangla accuracy on the full representation and $a_e$ and $a_b$ its accuracies after each removal, the \emph{erase share}
\begin{equation}
\rho_{\mathrm{erase}} = \frac{A-a_e}{A-a_b}
\label{eq:erase}
\end{equation}
is the share of the native eraser's effect that the English eraser achieves. Retraining after removal estimates residual recoverability under the specified linear-probe procedure.

\para{Overlap and retrieval.} The \emph{overlap} of two rank-$M$ images is $\omega=\frac1M\sum_i\cos^2\theta_i=\|Q_e^{\top}Q_b\|_F^2/M$ over their principal angles $\theta_i$, and $M\omega$ counts their shared dimensions. Random pairs have $\E\,\omega=M/D$ (\cref{prop:chance}), so two random rank-11 subspaces of our encoders share 0.12--0.16 dimensions. To ask whether items rather than classes correspond, we also retrieve each English item's Bangla pair by cosine similarity after per-condition centering. \Cref{alg:protocol} lists the steps; \cref{app:method} justifies each.

\begin{algorithm}[t]
\caption{Read, transfer and erase.}
\label{alg:protocol}
\footnotesize
\begin{algorithmic}[1]
\Require $e$ and $b$ inputs, labels $y$, disjoint splits, $\mathcal{E}$, rank $M$
\State center each language on its own SPACE mean \Comment{\pref{prop:portable}}
\State $(Q_\ell,G_\ell)\gets\mathcal{E}(X_\ell^{\text{SPACE}},y)$, $\ell=e,b$ \Comment{\pref{prop:penalty}}
\State \textbf{read}: probe $G_ex$ on $b$-PROBE, test on $b$ \Comment{\lref{lem:metric}}
\State $R_{b\to b}$ with $G_b$; floor $R_0$; SSI by \cref{eq:ssi} \Comment{\pref{prop:read-erase}}
\State \textbf{transfer}: probe $G_ex$ on $e$-PROBE, test on $b$ \Comment{\pref{prop:portable}}
\State \textbf{erase}: remove $Q_eG_e$ or $Q_bG_b$, retrain on $b$ \Comment{\lref{lem:containment}}
\State \textbf{overlap}: $\omega\gets\|Q_e^{\top}Q_b\|_F^2/M$ \Comment{\pref{prop:chance}}
\State \textbf{nulls}: steps 3--6 with random $Q$ (RAND) and $\mathcal{E}$ on permuted labels (PERM); Monte Carlo $p$, Holm \Comment{\cref{app:method-inference}}
\end{algorithmic}
\end{algorithm}

\subsection{Inference}
\label{sec:measures}

\para{Nulls.} Two nulls calibrate every SSI, with 1,999 draws each in the planned test (999 or 199 in some exploratory families). RAND reads the target through rank-matched random subspaces. RAND supplies a reference distribution for comparison; interpreting its Monte Carlo tail probability as a calibrated null-test $p$-value requires the observed statistic to be exchangeable with, or stochastically dominated by, that reference under the null. Because a probe on a random projection still exploits target labels, RAND calibrates what target supervision extracts from \emph{any} projection. PERM refits the estimator on labels permuted across SPACE components of equal size, which keeps the label dependence among records that share an idiom; it calibrates what the estimator finds without true labels. Robustness tests add 1,999 random subspaces drawn from the source covariance (RANDCOV), which meet its high-variance directions as often as the data do, and label-free PCA directions of the same rank. Monte Carlo $p$-values \citep{phipson-smyth-2010-permutation} are Holm-corrected within each family of tests \citep{holm-1979-simple}, with intervals from a 1,000-draw component bootstrap. Native accuracy and floor are fixed within a test, so a null test on SSI is a test on raw cross-lingual accuracy (gains: \cref{tab:app-absolute}).

\para{Planned test.} Before the campaign we fixed one confirmatory family: MLR, English counterparts (\cond{en-idiom}) $\to$ Bangla idioms in context (\cond{bn-fig}), F at $M{=}11$, S at $M{=}3$, six encoders (12 tests), each against RAND and PERM. The smallest attainable Holm-adjusted $p$ is $12/2000=.006$. All other analyses are exploratory, including the \textit{post hoc} families of \cref{tab:map}, each Holm-corrected within itself.

\para{Internal replication.} A 48-test plan was frozen before any outcome on a new split (seed 1) was computed: COV against both nulls in three nominated encoders (\hyp{1}), and paired comparisons of estimators (\hyp{2}), fixed-target sources (\hyp{3}) and bounded erasure (\hyp{4}) in all six encoders for F and S, under one Holm family (\cref{app:followup}). A second plan, frozen before seed 1 was examined for the idiom alone (\cond{bn-idiom}), tests it with COV against all three nulls (\cref{app:idiom-prospective}). Both reuse the corpus (932 of seed 1's 2,590 TEST records were discovery TEST records), so they are internal, not independent external validation. Our repository holds the code, both frozen plans and the aggregates of every analysis (\cref{app:repro}).

\para{Inferential status.} Every claim carries one of three labels: \emph{confirmatory} (the planned family), \emph{prospective} (frozen before the seed-1 outcomes it tests were computed) or \textit{post hoc} (planned after the discovery results). \Cref{tab:claims} summarizes the three central findings; \cref{tab:claims-detail} records the full evidence and inferential status.

\section{Experimental Setup}
\label{sec:setup}

\subsection{Data and inputs}
\label{sec:data}

Bagdhara pairs each Bangla idiom sense with a Bangla example sentence and its English translation, Bangla and English figurative glosses, English counterparts (98.8\% of records) and a sentiment label that LLMs drafted and at least two experts validated or replaced (\S3.2 of the corpus paper). Our export has 10,359 sense records (8,877 idiom strings). We build the eight inputs of \cref{tab:conditions}; the two conditions that locate the idiom in its sentence (\cond{bn-fig}, \cond{bn-context}) exclude the 15\% of records without an exact span match. Span conditions mean-pool the subword tokens that overlap the idiom's character span, context conditions all tokens but \texttt{[MASK]}, and sentences and glosses all tokens (\cref{app:data}). Connected components of records that share an idiom form, a variant or any identical input are assigned to SPACE (40\%; estimation), PROBE (25\%; readouts), VALIDATION (10\%; layer selection) and TEST (25\%), so no input crosses splits. TEST holds 2,590 records in each split. \cond{bn-fig} and \cond{bn-context} are scored on those with an exact span, 2,166 in 1,053 components on the discovery split (the planned population) and 2,213--2,214 on seed 1; every other condition is scored on all TEST records that have the input (English counterparts: 98.8\%).

\begin{table}[!htb]
\caption{\textbf{Eight inputs for one record.} The Bangla idiom \bn{gache} (\bnrom{gache}, lit.\ \bnlit{gache}) and its English counterpart in each condition. Bangla inputs (\swatch{bntint}; English \swatch{entint}) appear in English translation with the idiom in Bangla script; the models see Bangla script throughout (\cref{tab:bangla-inputs}). Pool: tokens whose hidden states are averaged.}
\label{tab:conditions}
\centering
\small
\setlength{\tabcolsep}{3pt}
\begin{tabular}{@{}l>{\raggedright\arraybackslash}m{12.2cm}l@{}}
\toprule
\textbf{Condition} & \textbf{Input} & \textbf{Pool} \\
\midrule
\tint{bntint}{l}\cond{bn-fig} & `He is already dreaming of a new car\ldots{} \bn{gache}.' & idiom \\
\tint{bntint}{l}\cond{bn-context} & same sentence, idiom $\rightarrow$ \texttt{[MASK]} & context \\
\tint{bntint}{l}\cond{bn-sentence} & same sentence & sentence \\
\tint{bntint}{l}\cond{bn-idiom} & `Here the expression ``\bn{gache}'' is used.' & idiom \\
\tint{bntint}{l}\cond{bn-para} & `Arranging enjoyment before attainment\ldots' & gloss \\
\tint{entint}{l}\cond{en-idiom} & `Here the expression ``Counting one's chickens before they hatch'' is used.' & span \\
\tint{entint}{l}\cond{en-trans} & `He's already dreaming of buying a new car assuming he'll get the job -- that's like counting his chickens before they hatch.' & sentence \\
\tint{entint}{l}\cond{en-para} & `Planning enjoyment before acquisition\ldots' & gloss \\
\bottomrule
\end{tabular}
\end{table}

\subsection{Concepts and human checks}
\label{sec:concepts}

\para{Concepts.} F (figurative class) clusters the English figurative glosses of SPACE records into 12 classes by $k$-means over embeddings from \texttt{all-MiniLM-L6-v2}, a sentence encoder excluded from the probed models \citep{reimers-gurevych-2019-sentence,wang-etal-2020-minilm}. F is thus a machine taxonomy of coarse meaning, which changes substantially across clustering seeds (adjusted Rand index, ARI, 0.40--0.63 with F) and whose classes we name \textit{post hoc} (\cref{app:taxonomy}). As F favors English glosses by construction, the planned test uses no gloss, and three label sets not built from English glosses complement F: S, the export's expert-validated sentiment labels (negative 47.8\%, neutral 34.1\%, positive 16.6\%, mixed 1.5\%; \cref{app:taxonomy}); FBN, 12 clusters of the \emph{Bangla} glosses in the never-probed NLLB-200 encoder \citep{nllb-2022-no}; and DOM, the 12 most frequent usage-domain tags (\cref{app:concepts}). Robustness analyses rebuild F with three further $k$-means seeds and build TAG by applying F's recipe to the corpus's topical tags, so that TAG's twelve classes use neither gloss field (\cref{app:taxonomies}).

\para{Human checks.} Two undergraduate volunteers, native Bangla speakers fluent in English and not authors, independently judged 300 TEST items (25 per F class; \cref{app:human}). For each item they judged whether its F class describes the idiom's main meaning, whether the idiom mainly describes or judges, what kind of expression its English counterpart is, and whether the counterpart preserves the meaning. Both found the assigned class apt for \HvAptBoth\% of items and misleading for \HvPoorBoth\% (\cref{tab:travel}), and both judged \HvIdiomBoth\% of English counterparts conventional idioms and \HvEquivBoth\% equivalent in meaning (agreement $\kappa$ \HvKappaRange).

\subsection{Models and layers}
\label{sec:models}

The six encoders differ in cross-lingual training signal: BanglaBERT (Bangla only) and BanglishBERT (Bangla and English) \citep{bhattacharjee-etal-2022-banglabert}, mBERT \citep{devlin-etal-2019-bert} and XLM-R-large \citep{conneau-etal-2020-unsupervised} (pretrained on monolingual corpora), and MuRIL-large \citep{khanuja-etal-2021-muril} and LaBSE \citep{feng-etal-2022-language} (translation pairs). For each model, one layer, chosen from every third layer by within-Bangla F validation accuracy and never by transfer, serves all estimators (all layers and fixed relative depths: \cref{app:layers}). \Cref{tab:models} lists the frozen configurations; an exploratory decoder is reported in \cref{app:decoder}.

\begin{table}[t]
\caption{\textbf{The encoders.} Cross-lingual training signal, frozen configuration, backbone parameter count and the layer selected on within-Bangla validation accuracy in the discovery study (MLR) and the follow-up (COV). The decoder Qwen3-8B is discovery-only and exploratory (\cref{app:decoder}). Complete revisions and model-source and license records are in the repository.}
\label{tab:models}
\centering
\small
\setlength{\tabcolsep}{3.6pt}
\begin{tabular}{@{}llrrrlcc@{}}
\toprule
 & & & & & & \multicolumn{2}{c}{\textbf{Selected layer}} \\
\cmidrule(l){7-8}
\textbf{Backbone} & \textbf{Cross-lingual signal} & \textbf{Params (M)} & \textbf{Width} & \textbf{Blocks} & \textbf{Pinned revision} & \textbf{Discovery} & \textbf{Seed 1} \\
\midrule
BanglaBERT & Bangla only & 110.0 & 768 & 12 & \texttt{9ce791f33057} & 9 & 6 \\
BanglishBERT & Bangla and English & 110.0 & 768 & 12 & \texttt{88f2777acd65} & 12 & 9 \\
mBERT & monolingual corpora & 177.9 & 768 & 12 & \texttt{3f076fdb1ab6} & 12 & 9 \\
XLM-R & monolingual corpora & 559.9 & 1024 & 24 & \texttt{c23d21b0620b} & 24 & 15 \\
MuRIL & translation pairs & 505.9 & 1024 & 24 & \texttt{ace319f0d175} & 12 & 15 \\
LaBSE & translation pairs & 470.9 & 768 & 12 & \texttt{5afa72968dfe} & 6 & 12 \\
\addlinespace[2.5pt]
Qwen3-8B & decoder (\cref{app:decoder}) & 7568.4 & 4096 & 36 & \texttt{b968826d9c46} & 16 & -- \\
\bottomrule
\end{tabular}
\end{table}

\FloatBarrier
\section{Results}
\label{sec:results}

The four research questions yield three findings: class-mean sharing, estimator-sensitive transfer, and the separation of reading from erasure. \Cref{tab:claims} summarizes them; \cref{tab:claims-detail} records their inferential status.

\begin{table}[t]
\caption{\textbf{Where English directions recover Bangla figurative classes.} SSI (\cref{eq:ssi}) of English-counterpart
directions on Bangla idioms in context; each SSI divides by Bangla's own subspace from the same estimator, not the full
space. \emph{Planned test}: MLR, accuracy, 95\% bootstrap intervals, \NxPlanCrossGain\,pp
above the floor (\cref{tab:app-absolute}); \emph{balanced acc.}: the same test by macro recall, nulls rescored. Then COV
directions on the discovery split, on seed 1, \emph{zero-shot} (an English-trained probe; \cref{sec:rq3}) and on Bangla idioms
without context (\emph{idiom alone}: seed 1, balanced accuracy; \cref{app:idiom-prospective}). TEST records: the 2,166 with an
exact idiom span (seed 1: 2,213--2,214) and, for the idiom alone, 2,588--2,589. \swatch{keytint}\,Bold:
beats both nulls after Holm correction within its column pair (1,999 draws; zero-shot 999), and for the idiom alone a
covariance-matched null too. The planned MLR family is confirmatory. The seed-1 column shows an exploratory 12-cell
reanalysis; prospective decisions for the three nominated encoders use the separate Holm-48 family
(\cref{tab:followup-h1}). The idiom-alone column reports the later prospective internal test. Both prospective designs
reuse the corpus (932 of seed 1's 2,590 TEST records also appear in discovery TEST).}
\label{tab:map}
\centering
\small
\setlength{\tabcolsep}{2.9pt}
\begin{tabular}{@{}l cc cc cc cc cc cc@{}}
\toprule
 & \multicolumn{2}{c}{\textbf{Planned test}} & \multicolumn{2}{c}{\textbf{Balanced acc.}} & \multicolumn{2}{c}{\textbf{Patterns}}
 & \multicolumn{2}{c}{\textbf{Seed 1}} & \multicolumn{2}{c}{\textbf{Zero-shot}} & \multicolumn{2}{c}{\textbf{Idiom alone}} \\
\cmidrule(lr){2-3}\cmidrule(lr){4-5}\cmidrule(lr){6-7}\cmidrule(lr){8-9}\cmidrule(lr){10-11}\cmidrule(l){12-13}
\textbf{Model} & \textbf{F [95\% CI]} & \textbf{S [95\% CI]} & \textbf{F} & \textbf{S} & \textbf{F} & \textbf{S} & \textbf{F} & \textbf{S} & \textbf{F} & \textbf{S} & \textbf{F} & \textbf{S} \\
\midrule
BanglaBERT & .16\,{\scriptsize[.03, .28]} & .15\,{\scriptsize[.00, .29]} & .33 & .15 & \hlkey \textbf{.48} & \hlkey \textbf{.62} & \hlkey \textbf{.66} & \hlkey \textbf{.56} & \hlkey \textbf{.32} & \hlkey \textbf{.53} & \hlkey \textbf{.76} & \hlkeyr \textbf{.54} \\
BanglishBERT & .20\,{\scriptsize[$-$.06, .43]} & .38\,{\scriptsize[.15, .58]} & \hlkey \textbf{.47} & \hlkey \textbf{.37} & .33 & \hlkey \textbf{.88} & \hlkey \textbf{.73} & \hlkey \textbf{.84} & .04 & .68 & \hlkey \textbf{1.00} & \hlkeyr \textbf{1.09} \\
mBERT & .20\,{\scriptsize[$-$.14, .47]} & .40\,{\scriptsize[.06, .77]} & .50 & .36 & \hlkey \textbf{.32} & .28 & \hlkey \textbf{.61} & \hlkey \textbf{.58} & \hlkey \textbf{.09} & .24 & .85 & .59 \\
XLM-R & \hlkey \textbf{.29}\,{\scriptsize[.10, .46]} & \hlkey \textbf{.47}\,{\scriptsize[.29, .65]} & \hlkey \textbf{.47} & \hlkey \textbf{.43} & \hlkey \textbf{.31} & .38 & \hlkey \textbf{.95} & \hlkey \textbf{1.22} & \hlkey \textbf{.13} & .31 & \hlkey \textbf{1.16} & \hlkeyr \textbf{1.75} \\
MuRIL & \hlkey \textbf{.48}\,{\scriptsize[.34, .62]} & .32\,{\scriptsize[.11, .52]} & \hlkey \textbf{.59} & \hlkey \textbf{.42} & \hlkey \textbf{.82} & \hlkey \textbf{.92} & \hlkey \textbf{.89} & \hlkey \textbf{.91} & \hlkey \textbf{.76} & \hlkey \textbf{.93} & \hlkey \textbf{.96} & \hlkeyr \textbf{1.12} \\
LaBSE & \hlkey \textbf{.50}\,{\scriptsize[.33, .66]} & .30\,{\scriptsize[.13, .43]} & \hlkey \textbf{.60} & .30 & \hlkey \textbf{.65} & \hlkey \textbf{.87} & \hlkey \textbf{.67} & \hlkey \textbf{1.00} & \hlkey \textbf{.69} & \hlkey \textbf{.94} & \hlkey \textbf{1.06} & \hlkeyr \textbf{1.31} \\
\bottomrule
\end{tabular}
\end{table}

\subsection{\rqn{1}: Class-mean directions support robust cross-lingual reading}
\label{sec:rq1}

\para{The planned test.} English counterpart directions recover F beyond both nulls in XLM-R, MuRIL and LaBSE (SSI 0.29, 0.48 and 0.50; Holm $p=.006$ against each null) and S in XLM-R (0.47; \cref{tab:map}). The F gains are \NxPlanPassGainF{} accuracy points above a \NxPlanFloorF\% floor; native gains of \NxPlanNativeGain{} points keep the SSI denominators well above the 2-point cutoff (\cref{tab:app-absolute}).

\para{Default MLR depends on the metric and null.} MLR's verdicts vary with the metric and taxonomy (\cref{tab:app-balanced,tab:app-taxonomies}), and only \NzPlannedSurvive{} of its \NzPlannedRejections{} planned rejections survive RANDCOV. Its limited advantage over covariance-matched random directions and label-free PCA weakens the evidence for class-specific sharing at the default penalty (\cref{tab:app-matched-nulls}).

\para{Class-mean reading survives stronger controls.} In the \textit{post hoc} analysis, COV directions beat both nulls and RANDCOV in all 12 tests under balanced accuracy, and PCA by \NzCovPcaRange{} points (\cref{tab:app-matched-nulls}). Alternative taxonomies, nonlinear readouts and fixed relative depth support this result (\cref{app:taxonomies,app:concepts,app:nonlinear,app:layers}). Refitting subspaces and probes widens bootstrap intervals by a median \NzRefitWidth{} times, but all COV intervals remain above zero under balanced accuracy (\cref{tab:app-refit}). The prospective internal replication supports the F result in all three nominated encoders (\cref{tab:followup}).

\begin{table}[t]
\caption{\textbf{A covariance-matched null and a label-free reference.} SSI under balanced accuracy on the discovery split,
against each estimator's own native reference, Bangla's subspace from the same estimator (TEST: 2,166 exact-span records in
context, 2,590 for the idiom alone). RANDCOV: mean SSI of 1,999 random subspaces whose directions
follow the covariance of English SPACE inputs; PCA: English SPACE's top principal directions of the same rank.
\swatch{keytint}\,Bold: beats RANDCOV after Holm correction over the 12 tests of its target and estimator.
COV$-$PCA: paired balanced-accuracy difference on the same TEST items (points), \swatch{keytint}\,where its
component-bootstrap interval lies above zero. \textit{Post hoc}.}
\label{tab:app-matched-nulls}
\centering
\footnotesize
\setlength{\tabcolsep}{3.6pt}
\begin{tabular}{@{}l ccccc ccccc@{}}
\toprule
 & \multicolumn{5}{c}{\textbf{F ($M{=}11$)}} & \multicolumn{5}{c}{\textbf{S ($M{=}3$)}} \\
\cmidrule(lr){2-6}\cmidrule(l){7-11}
\textbf{Model} & \textbf{MLR} & \textbf{COV} & \textbf{RANDCOV} & \textbf{PCA} & \textbf{COV$-$PCA} & \textbf{MLR} & \textbf{COV} & \textbf{RANDCOV} & \textbf{PCA} & \textbf{COV$-$PCA} \\
\midrule
\band{11}{Idioms in context (\cond{bn-fig})}{} \\
BanglaBERT & .33 & \hlkey \textbf{.58} & .36 & .41 & \hlkey +4.1 & .15 & \hlkey \textbf{.54} & .17 & .19 & \hlkeyr +6.3 \\
BanglishBERT & .47 & \hlkey \textbf{.50} & .30 & .28 & \hlkey +3.1 & \hlkey \textbf{.37} & \hlkey \textbf{.72} & .29 & .35 & \hlkeyr +2.6 \\
mBERT & .50 & \hlkey \textbf{.53} & .20 & .35 & \hlkey +2.7 & .36 & \hlkey \textbf{.38} & .08 & .18 & \hlkeyr +1.5 \\
XLM-R & \hlkey \textbf{.47} & \hlkey \textbf{.53} & .13 & .13 & \hlkey +4.6 & \hlkey \textbf{.43} & \hlkey \textbf{.49} & .09 & .01 & \hlkeyr +2.7 \\
MuRIL & \hlkey \textbf{.59} & \hlkey \textbf{.87} & .32 & .46 & \hlkey +10.4 & .42 & \hlkey \textbf{.94} & .17 & .24 & \hlkeyr +10.9 \\
LaBSE & .60 & \hlkey \textbf{.73} & .38 & .55 & \hlkey +3.4 & .30 & \hlkey \textbf{.94} & .25 & .50 & \hlkeyr +4.0 \\
\addlinespace[2.5pt]
\band{11}{The idiom alone (\cond{bn-idiom})}{} \\
BanglaBERT & .35 & \hlkey \textbf{.70} & .49 & .60 & \hlkey +1.8 & .23 & \hlkey \textbf{.61} & .24 & .26 & \hlkeyr +3.6 \\
BanglishBERT & .61 & \hlkey \textbf{.75} & .51 & .45 & \hlkey +2.6 & \hlkey \textbf{.93} & \hlkey \textbf{1.03} & .44 & .37 & \hlkeyr +3.0 \\
mBERT & .50 & .43 & .34 & .50 & \ensuremath{-}0.5 & .11 & .53 & .18 & .41 & +0.5 \\
XLM-R & .35 & .54 & .36 & .51 & +0.3 & .23 & -- & -- & -- & \hlkeyr +0.9 \\
MuRIL & .65 & \hlkey \textbf{.96} & .39 & .53 & \hlkey +7.5 & .49 & \hlkey \textbf{1.11} & .20 & .29 & \hlkeyr +9.0 \\
LaBSE & .83 & \hlkey \textbf{1.03} & .50 & .70 & \hlkey +3.6 & .57 & \hlkey \textbf{1.09} & .31 & .65 & \hlkeyr +2.9 \\
\bottomrule
\end{tabular}
\end{table}

\subsection{\rqn{2}: Explicit meaning travels further than idioms}
\label{sec:rq2}

\begin{figure}[t]
\centering
\begin{subfigure}[t]{0.3239\textwidth}
\centering
\includegraphics[width=\linewidth,viewport=0.84 0.84 154.80 105.72,clip]{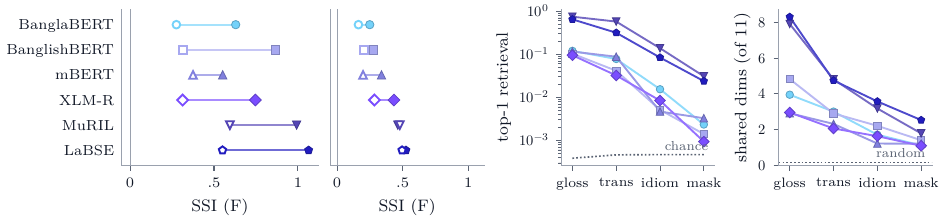}
\caption{Bangla glosses}\label{fig:gap-a}
\end{subfigure}\hfill
\begin{subfigure}[t]{0.1639\textwidth}
\centering
\includegraphics[width=\linewidth,viewport=156.49 0.84 234.41 105.72,clip]{figures/fig_idiom_gap.pdf}
\caption{Bangla idioms}\label{fig:gap-b}
\end{subfigure}\hfill
\begin{subfigure}[t]{0.2363\textwidth}
\centering
\includegraphics[width=\linewidth,viewport=236.09 0.84 348.45 105.72,clip]{figures/fig_idiom_gap.pdf}
\caption{item retrieval}\label{fig:gap-c}
\end{subfigure}\hfill
\begin{subfigure}[t]{0.2159\textwidth}
\centering
\includegraphics[width=\linewidth,viewport=350.13 0.84 452.75 105.72,clip]{figures/fig_idiom_gap.pdf}
\caption{class-mean overlap}\label{fig:gap-d}
\end{subfigure}
\caption{\textbf{Explicit meaning travels further than idioms.} (a--b)~SSI for F on a fixed Bangla target, from English directions estimated on glosses or translations (\mkfilled) or on idiom counterparts (\mkopen): explicit sources read more in all six encoders in (a) and in five in (b). (c)~Top-1 retrieval of the paired item (log scale; dotted: chance) and (d)~dimensions shared by independent English and Bangla COV subspaces (of 11; dotted: random) follow the same ladder. Encoders: \encoderkey. Discovery split (TEST: 2,589--2,590 Bangla glosses in a, 2,166 exact-span idioms in b); exploratory.}
\label{fig:gap}
\end{figure}

\para{A fixed target.} Holding the Bangla target fixed keeps the SSI native reference constant (\cref{fig:gap}a--b). On Bangla glosses, English-gloss directions outperform counterpart directions in all six encoders (SSI \TxGlossSSIF{} \textit{vs.}\ \TxCounterGlossSSIF); seed 1 confirms the advantage for F in \FuHthreeNF{} of 6 encoders, not for S. F's English-gloss construction partly favors this contrast, but it persists with alternative taxonomies, including labels built without English glosses (\cref{tab:app-taxonomies,app:concepts}).

\para{The idiom alone shares its class, prospectively.} Masked context alone carries substantial class information: median native above-floor F accuracy is 11.5 points, compared with 10.8 for the idiom span (\cref{tab:app-readability}). After a \textit{post hoc} discovery analysis, we froze the design and decision rule for idioms alone in a neutral frame, then tested them on the internal seed-1 re-split (\cref{app:idiom-prospective}). COV beats RAND, PERM and RANDCOV after Holm correction in \NzProPass{} of 12 balanced-accuracy tests; \NzProFail{} alone fails PERM. Median SSI is \NzProSsiMedian. English classifiers also transfer in \NzProZsPass{} of 12 tests under the conditional zero-shot design of \cref{sec:rq3}, whereas MLR passes only one idiom-alone test (\cref{tab:app-idiom-prospective}).

\begin{table}[t]
\caption{\textbf{The idiom without its context, prospectively.} The frozen plan of \cref{app:idiom-prospective} on the
follow-up split: SSI under balanced accuracy of English-counterpart directions on Bangla idioms alone in a neutral frame
(\cond{bn-idiom}; 2,588--2,589 TEST records), with component-bootstrap intervals; COV and MLR SSIs each divide by Bangla's
own subspace from the same estimator. \emph{Gain}: COV's balanced accuracy above the floor (points);
\emph{Zero-shot}: an English-trained probe through COV directions at a Bangla-validation-selected layer. \swatch{keytint}\,Bold: beats the
random-subspace, permuted-label and covariance-matched nulls (1,999 draws each; zero-shot: the first two, 999 draws)
after Holm correction over the 12 tests. Primary test: COV; the other columns are the plan's secondary tests.}
\label{tab:app-idiom-prospective}
\centering
\small
\setlength{\tabcolsep}{3.2pt}
\begin{tabular}{@{}l cccc cccc@{}}
\toprule
 & \multicolumn{4}{c}{\textbf{F ($M{=}11$)}} & \multicolumn{4}{c}{\textbf{S ($M{=}3$)}} \\
\cmidrule(lr){2-5}\cmidrule(l){6-9}
\textbf{Model} & \textbf{COV SSI [95\% CI]} & \textbf{Gain} & \textbf{Zero-shot} & \textbf{MLR} & \textbf{COV SSI [95\% CI]} & \textbf{Gain} & \textbf{Zero-shot} & \textbf{MLR} \\
\midrule
BanglaBERT & \hlkey \textbf{.76}\,{\scriptsize[.68, .83]} & 16.0 & \hlkey \textbf{.52} & .47 & \hlkey \textbf{.54}\,{\scriptsize[.44, .64]} & 5.8 & \hlkey \textbf{.51} & .01 \\
BanglishBERT & \hlkey \textbf{1.00}\,{\scriptsize[.88, 1.12]} & 13.5 & \hlkey \textbf{.82} & .77 & \hlkey \textbf{1.09}\,{\scriptsize[.94, 1.27]} & 8.2 & \hlkey \textbf{1.11} & .35 \\
mBERT & .85\,{\scriptsize[.70, 1.04]} & 7.2 & \hlkey \textbf{.68} & .33 & .59\,{\scriptsize[.38, .80]} & 2.1 & .47 & .46 \\
XLM-R & \hlkey \textbf{1.16}\,{\scriptsize[1.00, 1.33]} & 11.6 & \hlkey \textbf{1.10} & 1.02 & \hlkey \textbf{1.75}\,{\scriptsize[1.39, 2.33]} & 7.2 & \hlkey \textbf{2.08} & .71 \\
MuRIL & \hlkey \textbf{.96}\,{\scriptsize[.87, 1.04]} & 18.0 & \hlkey \textbf{.90} & .88 & \hlkey \textbf{1.12}\,{\scriptsize[.99, 1.30]} & 11.2 & \hlkey \textbf{1.21} & 1.05 \\
LaBSE & \hlkey \textbf{1.06}\,{\scriptsize[.93, 1.23]} & 13.0 & \hlkey \textbf{1.02} & \hlkey \textbf{.81} & \hlkey \textbf{1.31}\,{\scriptsize[1.09, 1.63]} & 6.9 & \hlkey \textbf{1.53} & .77 \\
\bottomrule
\end{tabular}
\end{table}

\para{Idiom-like counterparts share less.} Alignment raises gloss-pair retrieval from \NxRetGlossBefore{} to \NxRetGlossAfter{} in the four weaker encoders, while idiom-pair retrieval stays at or below \NxRetIdiomMax{} (\cref{tab:app-alignment}). The least gloss-like third of counterparts yields weaker class readout than random thirds on average in all \NzOpaqueBelowFig{} cells (\cref{app:counterpart-type}). Items whose counterparts both volunteers judged conventional idioms are read correctly less often, by \NzIdiomCpIdiomAbs{} points alone and \NzIdiomCpFigAbs{} in context, within class and net of native readout (Holm $p\le\NzIdiomCpFigHolm$; \cref{tab:app-human-items}). These associations concern coarse class sharing; they do not isolate idiomaticity.

\subsection{\rqn{3}: Input maps and regularization shape classifier transfer}
\label{sec:rq3}

\begin{figure}[t]
\centering
\begin{subfigure}[t]{0.2927\textwidth}
\centering
\includegraphics[width=\linewidth,viewport=0.84 0.84 143.45 115.80,clip]{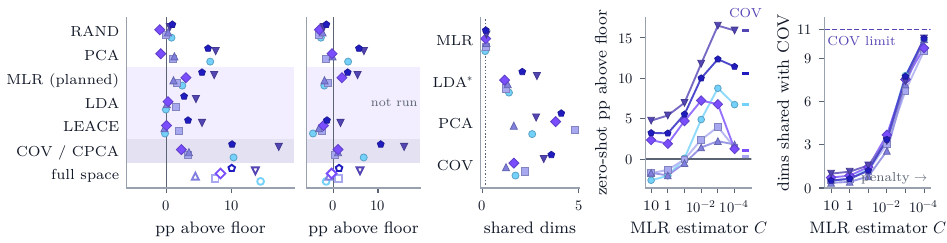}
\caption{supervised}\label{fig:estimators-a}
\end{subfigure}\hfill
\begin{subfigure}[t]{0.1206\textwidth}
\centering
\includegraphics[width=\linewidth,viewport=145.13 0.84 203.87 115.80,clip]{figures/fig_estimators.pdf}
\caption{zero-shot}\label{fig:estimators-b}
\end{subfigure}\hfill
\begin{subfigure}[t]{0.1563\textwidth}
\centering
\includegraphics[width=\linewidth,viewport=205.55 0.84 281.72 115.80,clip]{figures/fig_estimators.pdf}
\caption{image overlap}\label{fig:estimators-c}
\end{subfigure}\hfill
\begin{subfigure}[t]{0.1775\textwidth}
\centering
\includegraphics[width=\linewidth,viewport=283.40 0.84 369.86 115.80,clip]{figures/fig_estimators.pdf}
\caption{zero-shot along $C$}\label{fig:estimators-d}
\end{subfigure}\hfill
\begin{subfigure}[t]{0.1729\textwidth}
\centering
\includegraphics[width=\linewidth,viewport=369.86 0.84 454.09 115.80,clip]{figures/fig_estimators.pdf}
\caption{Penalty limit}\label{fig:estimators-e}
\end{subfigure}
\caption{\textbf{Patterns, filters and the penalty between them.} (a)~Bangla F accuracy above the 15.7\% floor through English rank-11 directions, read by a Bangla probe (shaded: filters violet, patterns lavender; \mkopen: full space), and (b)~zero-shot by an English probe. (c)~Dimensions shared by separately fitted English and Bangla images (dotted: random; $^*$one coincident, uninformative axis). (d)~Zero-shot accuracy through English MLR directions rises as the penalty grows ($C$ falls), toward COV's (ticks), as (e)~the MLR subspace approaches the COV image (\cref{prop:penalty}). Encoders: \encoderkey. Discovery split, 2,166 exact-span TEST records; exploratory.}
\label{fig:estimators}
\end{figure}

\para{The estimator changes the answer.} At MLR's default $C{=}1$, COV and CPCA directions recover \TxCovMlrRatio{} times MLR's above-floor accuracy in \TxCovMlrRatioN{} of six encoders (\cref{fig:estimators}a; \cref{tab:app-estimators}). Separate English and Bangla MLR subspaces share 0.14--0.20 of 11 dimensions, near chance; COV subspaces share 1.2--3.6 (\cref{fig:estimators}c). The penalty accounts for most of the difference.

\begin{proposition}[Filters approach patterns]\label{prop:penalty}
Let inputs be centered and let $W_C$ minimize $\frac12\|W\|_F^2+C\sum_i\ell(Wx_i,y_i)$ for the multinomial logistic loss without intercept. As $C\to0$, $W_C=Cn\,[\pi_k\mu_k^{\top}]_{k=1}^{K}+O(C^2)$, where $\pi_k$ and $\mu_k$ are the class priors and class-mean offsets; at full supported rank $M=K-1$ the MLR subspace therefore converges to the COV image.
\end{proposition}

This known strong-penalty limit has a closed-form ridge analog (\cref{app:theory-patterns}): in the eigenbasis of $\Sigma$, weights scale by $1/(s_j+\lambda)$, amplifying low-variance directions at weak penalties and approaching equal weighting at strong penalties. Contextual anisotropy \citep{ethayarajh-2019-contextual,timkey-van-schijndel-2021-bark} can thus make a filter nearly orthogonal to its pattern (\cref{eq:antieigen}). As $C$ falls to $10^{-4}$, MLR's overlap with COV rises from \NyPathDimsDefaultF{} to \NyPathDimsLimitF{} of 11 dimensions (\cref{fig:estimators}e).

\para{Regularization improves classifier transfer.} Zero-shot fitting and penalty selection use English labels, conditional on the Bangla-validation-selected layer. COV classifiers beat both nulls in \NxZsBoth{} of 12 tests, retaining median \NxZsRatioF\% (F) and \NxZsRatioS\% (S) of supervised SSI (\cref{tab:app-zero-shot}). Default MLR, LEACE and full-space readouts have negative median accuracy-based SSI, although balanced accuracy retains signal. Increasing the penalty raises MLR's median zero-shot SSI from \NyPathZsDefaultF{} at $C{=}1$ to \NyPathZsBestF{} at $C{=}10^{-3}$; regularizing LEACE's whitening also improves transfer. English-selected penalties yield classifiers that beat the random-subspace null in \NySelMlr{} of \NySelMlrOf{} MLR and \NySelLeace{} of \NySelLeaceOf{} LEACE tests (\cref{tab:app-penalty-path}).

\begin{table}[t]
\caption{\textbf{Zero-shot readout.} SSI of an English-trained probe applied unchanged to Bangla idioms in context, through
each estimator's English subspace or the full English representation, beside COV's supervised SSI (Bangla-trained
probe; \cref{tab:app-estimators}). Zero-shot inputs are centered per language on unlabeled SPACE means, except
\emph{COV, source}, which uses the English mean for Bangla inputs. Both settings retain the Bangla-validation-selected layer and the pooled SPACE scale (\cref{app:method-inputs}). Supervised and zero-shot SSIs
divide by the same native gain, that of Bangla's own subspace from the same estimator (for the full space, the full
Bangla representation), so they share a scale. TEST: the 2,166 exact-span records. \swatch{keytint}\,Bold: beats both zero-shot nulls (999 random subspaces
and 999 permuted-label estimators, each read by an English-trained probe) after Holm correction over 12 tests;
\swatch{negtint}\,below the majority floor.}
\label{tab:app-zero-shot}
\centering
\footnotesize
\setlength{\tabcolsep}{3pt}
\begin{tabular}{@{}l c ccccc cccc@{}}
\toprule
 & \textbf{Supervised} & \multicolumn{5}{c}{\textbf{Zero-shot SSI (accuracy)}} & \multicolumn{4}{c}{\textbf{Zero-shot SSI (balanced accuracy)}} \\
\cmidrule(lr){2-2}\cmidrule(lr){3-7}\cmidrule(l){8-11}
\textbf{Model} & \textbf{COV} & \textbf{COV} & \textbf{MLR} & \textbf{LEACE} & \textbf{Full space} & \textbf{COV, source} & \textbf{COV} & \textbf{MLR} & \textbf{LEACE} & \textbf{Full space} \\
\midrule
\band{11}{Figurative classes F ($M{=}11$)}{} \\
BanglaBERT & .48 & \hlkey \textbf{.32} & \hlneg $-$.14 & \hlneg $-$.25 & \hlneg $-$.12 & \hlkey \textbf{.27} & .47 & .22 & .15 & .28 \\
BanglishBERT & .33 & .04 & \hlneg $-$.17 & \hlneg $-$.22 & .15 & \hlneg $-$.09 & .36 & .40 & .30 & .57 \\
mBERT & .32 & \hlkey \textbf{.09} & \hlneg $-$.36 & \hlneg $-$.47 & \hlneg $-$.23 & \hlkey \textbf{.10} & .30 & .34 & .27 & .42 \\
XLM-R & .31 & \hlkey \textbf{.13} & .18 & \hlneg $-$.21 & \hlneg $-$.05 & .08 & .44 & .51 & .30 & .41 \\
MuRIL & .82 & \hlkey \textbf{.76} & .35 & .09 & .44 & \hlkey \textbf{.64} & .83 & .61 & .43 & .59 \\
LaBSE & .65 & \hlkey \textbf{.69} & .29 & \hlneg $-$.10 & .13 & \hlkey \textbf{.64} & .80 & .60 & .28 & .46 \\
\band{11}{Sentiment S ($M{=}3$)}{} \\
BanglaBERT & .62 & \hlkey \textbf{.53} & \hlneg $-$.18 & \hlneg $-$.16 & \hlneg $-$.93 & .43 & .40 & .22 & .26 & .29 \\
BanglishBERT & .88 & .68 & .21 & .18 & \hlneg $-$1.25 & .42 & .63 & .54 & .47 & .55 \\
mBERT & .28 & .24 & \hlneg $-$.06 & \hlneg $-$.20 & -- & \hlkey \textbf{.33} & .20 & .39 & .25 & .18 \\
XLM-R & .38 & .31 & .12 & .04 & \hlneg $-$.55 & .37 & .51 & .52 & .48 & .43 \\
MuRIL & .92 & \hlkey \textbf{.93} & \hlneg $-$.07 & \hlneg $-$.06 & \hlneg $-$.08 & \hlkey \textbf{.74} & .85 & .53 & .38 & .56 \\
LaBSE & .87 & \hlkey \textbf{.94} & \hlneg $-$.04 & \hlneg $-$.10 & \hlneg $-$.96 & .77 & .93 & .44 & .38 & .37 \\
\bottomrule
\end{tabular}
\end{table}

\begin{proposition}[Patterns are portable]\label{prop:portable}
Let class $k$ in language $\ell\in\{e,b\}$ be $\mathcal{N}(m_k,\Sigma_\ell)$ after centering, with $\Sigma_\ell\succ0$, offsets $m_k$ shared by the two languages and spanning $\operatorname{range}(Q)$, $Q^{\top}Q=I_M$.
(a)~Every input map $G$ with $GQ=I_M$ gives both languages the same class means, and noise $G\Sigma_\ell G^{\top}\succeq(Q^{\top}\Sigma_\ell^{-1}Q)^{-1}$, with equality iff $G=G_\ell=(Q^{\top}\Sigma_\ell^{-1}Q)^{-1}Q^{\top}\Sigma_\ell^{-1}$, LEACE's map when fitted in $\ell$. The pattern map $Q^{\top}$ does not depend on $\Sigma_e$; $G_e$ is optimal in $e$ and can be arbitrarily poor in $b$.
(b)~For two equiprobable classes in $D\ge2$, $m_\pm=\pm\delta/2$ with $\delta\ne0$, the pattern rule $\operatorname{sign}(\delta^{\top}x)$ has accuracy $\Phi\!\left(\frac{\|\delta\|^2}{2\sqrt{\delta^{\top}\Sigma_b\delta}}\right)$ in $b$ whatever $\Sigma_e$, while the zero-shot accuracy of $e$'s Bayes filter $\operatorname{sign}(\delta^{\top}\Sigma_e^{-1}x)$ takes every value in $\big(\frac12,\Phi(\frac12\sqrt{\delta^{\top}\Sigma_b^{-1}\delta})\big]$ over positive definite $\Sigma_e$.
\end{proposition}

Part (a) is Aitken's Gauss--Markov theorem: COV uses ordinary least-squares coordinates, whereas LEACE uses generalized coordinates optimal under the fitted covariance (\cref{app:theory-portable}). An English-optimal coordinate can be arbitrarily poor in Bangla (\cref{ex:portable}); probes recovering English's filters inherit this vulnerability (\cref{lem:filter}). Covariance shift is therefore a possible explanation for weak raw-accuracy transfer, not an established sole cause. With partly shared means, zero-shot SSI correlates with image overlap ($\rho=\NxOverlapZsRho$, \NxOverlapZsCells{} cells; \cref{lem:partial}).

\para{Why LEACE reads less.} LEACE shares COV's image yet reads less through English directions in \NxLeaceCrossWorse{} of \NxLeaceCrossOf{} cells. Part (a) allows this difference but does not explain why covariance matching fails; a whitening ridge approaching COV closes the gap (\cref{tab:e3,fig:app-leace}).

\subsection{\rqn{4}: Reading does not license erasure}
\label{sec:rq4}

Bangla-fitted LEACE reduces Bangla F accuracy by \TxEraseBnF\pp{}, whereas English-fitted LEACE removes at most \TxEraseEnMaxF\pp{}, or \TxEraseShareF{} of the native effect (\cref{tab:app-erasure}). Yet it removes \NxEnglishEraseShareF{} of held-out English's above-floor F signal. INLP shows the same asymmetry: English-fitted removal costs Bangla a median \NyInlpDropF\pp{}, versus \NyInlpNativeF\pp{} for native INLP, despite removing a median \NyInlpRankF{} directions (\cref{tab:app-erasers}). Its F overlap is near the rank-matched random reference ($\omega=\NyInlpOmegaF$; reference \NyInlpChanceF). MLPs agree; R-LACE insufficiently erases its own language to establish a transfer comparison (\cref{app:erasers}).

\begin{table}[t]
\caption{\textbf{Erasers fitted in English, applied to Bangla idioms in context.} Medians over six encoders; the reference eraser is fitted in Bangla (BN-fitted). \emph{Rank}: dimensions removed (INLP: $K-1$ per step; R-LACE: $M$). \emph{Guards}: source-validation accuracy no more than 1\,pp above the majority rate for INLP and 3\,pp for other methods; these diagnostic criteria differ from population linear guardedness. INLP uses the first successful step or the final available step. \emph{English}: share of held-out English above-floor accuracy removed (above 1 if accuracy falls below the floor). \emph{Bangla drop}: accuracy lost after removal and probe retraining; \emph{erase share}: English-fitted drop divided by the corresponding Bangla-fitted drop. $\omega$: overlap with Bangla's COV image (chance: $R/D$). \emph{Ordering}: erase share $\le\omega\le$ read share, a multiclass diagnostic motivated by \cref{prop:read-erase}. \emph{MLP}: erase share with MLP probes. \emph{Rotated}: erase share after the held-out rotation (\cref{app:heldout}). Ratios are undefined (--) when the native eraser removes at most 2\,pp. Discovery split, 2,166 exact-span TEST records; \textit{post hoc}.}
\label{tab:app-erasers}
\centering
\footnotesize
\setlength{\tabcolsep}{3.2pt}
\begin{tabular}{@{}l c c c cc c c c c c@{}}
\toprule
\textbf{\hdtwo{Eraser}} & \textbf{\hdtwo{Rank}} & \textbf{\hdtwo{Guards}} & \textbf{\hdtwo{English}} & \multicolumn{2}{c}{\textbf{Bangla drop (pp)}} &
\textbf{\hdtwo{Erase share}} & \textbf{\hdtwo{$\omega$ (chance)}} & \textbf{\hdtwo{Ordering}} & \textbf{\hdtwo{MLP}} & \textbf{\hdtwo{Rotated}} \\
\cmidrule(lr){5-6}
 & & & & \textbf{EN-fitted} & \textbf{BN-fitted} & & & & & \\
\midrule
\band{11}{Figurative class F}{} \\
COV image & 11 & 6/6 & .59 & 0.1 & 7.3 & .02 & .18 (.01) & 6/6 & .11 & .17 \\
LEACE & 11 & 6/6 & .56 & 0.5 & 6.7 & .06 & .18 (.01) & 1/6 & .13 & .18 \\
MLR filter & 11 & 0/6 & .02 & $-$0.1 & 0.3 & -- & .01 (.01) & -- & -- & -- \\
\hlkeyl INLP & \hlkey 231 & \hlkey 3/6 & \hlkey .46 & \hlkey 1.6 & \hlkey 8.6 & \hlkey .14 & \hlkey .25 (.26) & \hlkey 6/6 & \hlkey .15 & \hlkeyr .10 \\
R-LACE & 11 & 2/6 & .01 & $-$0.1 & 1.5 & .00 & .02 (.01) & 3/3 & $-$.07 & .11 \\
\addlinespace[2.5pt]
\band{11}{Sentiment S}{} \\
COV image & 3 & 6/6 & 1.34 & 0.8 & 7.2 & .09 & .16 (.00) & 5/6 & .10 & .18 \\
LEACE & 3 & 6/6 & 1.38 & 1.1 & 7.2 & .12 & .16 (.00) & 2/6 & .18 & .18 \\
MLR filter & 3 & 1/6 & .00 & 0.1 & 0.5 & -- & .00 (.00) & -- & -- & -- \\
\hlkeyl INLP & \hlkey 130 & \hlkey 6/6 & \hlkey 1.65 & \hlkey 1.4 & \hlkey 7.3 & \hlkey .17 & \hlkey .28 (.15) & \hlkey 4/5 & \hlkey .22 & \hlkeyr .24 \\
R-LACE & 3 & 1/6 & $-$.01 & 0.1 & $-$0.2 & -- & .00 (.00) & -- & -- & -- \\
\bottomrule
\end{tabular}
\end{table}

\para{Reading needs an angle; erasure needs containment.} A linear readout of $P_Sx$ can improve on the optimal constant predictor under a convex loss only if $S$ is not orthogonal to the target class-mean span $U$. Removing $S$ equalizes the means exactly when $U\subseteq S$ (\cref{lem:containment}). This distribution-free statement concerns convex loss, not accuracy; equal-dimensional containment requires equality. \Cref{prop:read-erase} quantifies the gap under Gaussian assumptions.

\Needspace{11\baselineskip}
\begin{samepage}
\begin{proposition}[Reading needs an angle, erasure needs containment]\label{prop:read-erase}
Let two equiprobable classes be $\mathcal{N}(\pm\delta/2,\sigma^2 I)$ and let $S$ be a subspace with $c=\|P_S\delta\|/\|\delta\|$. With Bayes-optimal linear readouts, the share $\rho_{\mathrm{read}}$ of native above-chance accuracy recovered by reading $P_Sx$ and the share $\rho_{\mathrm{erase}}$ destroyed by removing $S$ satisfy
\[\rho_{\mathrm{erase}}\le1-\sqrt{1-c^2}\le c^2\le c\le\rho_{\mathrm{read}}.\]
If the normalized pairwise class differences are isotropic in their span $U$ ($\dim U=M$), so that $\mathbb E[\hat\delta\hat\delta^{\top}]=P_U/M$, then $\mathbb E[c^2]=\omega$, and the corresponding weighted average of equal-prior binary pairwise shares satisfies erase share $\le\omega\le$ read share.
\end{proposition}
\end{samepage}

\begin{figure}[t]
\centering
\begin{subfigure}[t]{0.3256\textwidth}
\centering
\includegraphics[width=\linewidth,viewport=0.00 7.20 159.40 158.40,clip]{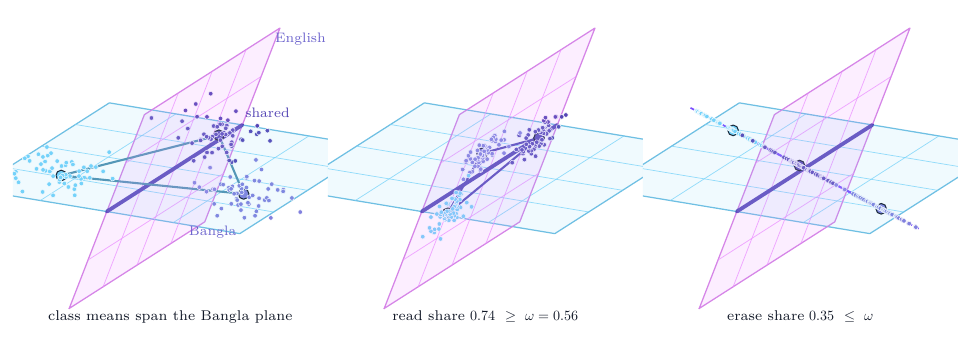}
\caption{Bangla classes, English subspace}\label{fig:app-theory-read-erase-a}
\end{subfigure}\hfill
\begin{subfigure}[t]{0.3088\textwidth}
\centering
\includegraphics[width=\linewidth,viewport=159.40 7.20 310.60 158.40,clip]{figures/app_theory_read_erase.pdf}
\caption{read: project onto it}\label{fig:app-theory-read-erase-b}
\end{subfigure}\hfill
\begin{subfigure}[t]{0.3256\textwidth}
\centering
\includegraphics[width=\linewidth,viewport=310.60 7.20 470.01 158.40,clip]{figures/app_theory_read_erase.pdf}
\caption{erase: remove it}\label{fig:app-theory-read-erase-c}
\end{subfigure}
\caption{\textbf{A three-dimensional schematic of reading and erasure.} (a)~Three Bangla classes whose means span the cyan plane, and an English subspace (pink) that meets it along one shared direction (purple) and crosses it at $70^\circ$. (b)~Projecting onto the English subspace keeps the classes apart, mostly along the shared direction. (c)~Removing the English subspace leaves its normal line (dashed), along which the classes stay separated. For isotropic Gaussian classes in this geometry, Monte Carlo Bayes accuracies give a read share of \ThToyRead{} and an erase share of \ThToyErase{} at overlap $\omega=\ThToyOmega$. In three dimensions two planes must share a line; two rank-11 subspaces of a 768- or 1,024-dimensional space need share nothing, and the measured F overlaps are \ThOverlapF.}
\label{fig:app-theory-read-erase}
\end{figure}

Partial projection can preserve binary isotropic Gaussian accuracy while leaving informative complementary separation (\cref{fig:app-theory-read-erase,app:theory-read-erase}). The pairwise extension does not bound multiclass SSI; the corresponding ordering below is an empirical diagnostic. Native gains of \NxPlanNativeGain{} points invite a qualitative comparison with the weak-signal regime, where the shares approach $c$ and $1-\sqrt{1-c^2}$, the latter approximately $c^2/2$ for small $c$. COV read shares exceed overlap in all \ThReadOK{} cells and erase shares fall below it in \ThEraseOK{}; INLP satisfies the ordering in \NyInlpPropOK{} of \NyInlpPropCells{} cells. On the internal seed-1 replication, the prospective erasure criterion---below a quarter of the native effect---holds after Holm correction in \FuHfourNF{} of 6 encoders for F and \FuHfourNS{} for S.

\begin{figure}[t]
\centering
\begin{minipage}{0.82\linewidth}
\begin{subfigure}[t]{0.5466\linewidth}
\centering
\includegraphics[width=\linewidth,viewport=0.84 0.84 120.66 119.40,clip]{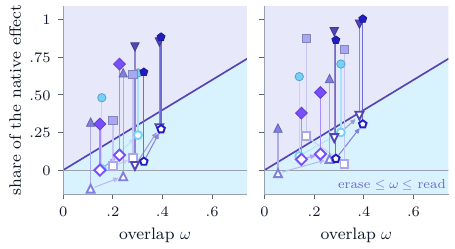}
\caption{figurative class F}\label{fig:read-erase-a}
\end{subfigure}\hfill
\begin{subfigure}[t]{0.4334\linewidth}
\centering
\includegraphics[width=\linewidth,viewport=122.34 0.84 217.32 119.40,clip]{figures/fig_read_erase.pdf}
\caption{sentiment S}\label{fig:read-erase-b}
\end{subfigure}
\end{minipage}
\caption{\textbf{Overlap separates reading from erasure.} Read share (\mkfilled; COV SSI) and erase share (\mkopen; Bangla accuracy lost by removing the English COV image, relative to Bangla's own) against the overlap $\omega$ of the two images, before and after (arrows) a held-out rotation into Bangla coordinates. Motivated by the binary and pairwise-averaged bound in \cref{prop:read-erase}, we check the analogous multiclass ordering: \ThBothOK{} of \ThCells{} cells satisfy this ordering before rotation, \NyAlignOK{} after. Encoders: \encoderkey. Discovery split, 2,166 exact-span TEST records; \textit{post hoc}.}
\label{fig:read-erase}
\end{figure}

\para{Rotation raises erasure.} A Procrustes rotation fitted on held-out pairs, none defining a class mean, raises English--Bangla COV overlap from \NyAlignNoneDimsF{} to \NyAlignPVDimsF{} of 11 dimensions (shuffled: \NyAlignShufDimsF), read shares to \NyAlignSSIAfter{} and erase shares to \NyAlignEraseF{} for F. All \NyAlignCells{} cells satisfy the multiclass diagnostic (\cref{fig:read-erase,tab:app-heldout}). Rotated COV also transfers zero-shot (F SSI \NyAlignZsCovF{}); rotated LEACE has median SSI \NyAlignZsLeaceMedian. Aligning images therefore does not resolve every transfer limitation, consistent with the possible covariance-shift mechanism of \cref{prop:portable}.

\para{How informative is the ordering?} It is a weak diagnostic for F before rotation: erase shares lie at least 0.14 below overlap, and read shares well above it. S is tighter and fails once, in BanglishBERT (erase share 0.23, overlap 0.17). Rotation brings erase shares within 0.02--0.09 of overlap in four of 12 cells; none crosses it. The reading side remains loose, with margins of at least 0.12 before and 0.29 after rotation. These multiclass observations do not test the binary theorem itself.

\subsection{Summary of the evidence}
\label{sec:summary}

\Cref{tab:claims} summarizes the findings and their scope; \cref{tab:claims-detail} records the evidence and inferential status of each claim. Only the planned family is confirmatory. The 48 seed-1 follow-up tests and the later idiom-alone plan are prospective but internal to the corpus. All other analyses, including empirical checks motivated by the propositions, are \textit{post hoc}.

\section{Discussion}
\label{sec:discussion}

\begin{table}[t]
\caption{\textbf{Three findings and their scope.} The original MLR family is confirmatory; stronger pattern results are
\textit{post hoc} or prospective internal replications on the same corpus. The complete claim-by-claim evidence and
inferential status are retained in \cref{tab:claims-detail}.}
\label{tab:claims}
\centering
\small
\setlength{\tabcolsep}{4pt}
\begin{tabular}{@{}>{\raggedright\arraybackslash}m{4.6cm}>{\raggedright\arraybackslash}m{7.2cm}>{\raggedright\arraybackslash}m{4.0cm}@{}}
\toprule
\textbf{Finding} & \textbf{Decisive evidence} & \textbf{Scope} \\
\midrule
\textbf{Coarse-class information is shared most robustly through class means.} &
The planned MLR family rejects both nulls in 4/12 tests. COV beats three nulls in all 12 contextual tests under balanced accuracy, and in \NzProPass/12 prospective idiom-alone tests (\cref{tab:map,tab:app-matched-nulls,tab:app-idiom-prospective}). &
One language pair and corpus; glosses share more than counterparts. Seed 1 is an internal re-split. \\
\addlinespace[4pt]
\textbf{Transfer depends on the input map and penalty.} &
COV supports English-trained readouts; default MLR and LEACE have much lower median accuracy SSI. Stronger source-selected penalties improve transfer, and COV and LEACE differ despite sharing an image (\cref{fig:estimators}; \cref{tab:app-zero-shot}). &
Readout training is source-only, but layer selection uses Bangla labels. Covariance shift is a possible mechanism. \\
\addlinespace[4pt]
\textbf{Reading does not establish effective cross-language erasure.} &
English-fitted LEACE removes under 1\,pp of F accuracy, versus \TxEraseBnFRound{}\,pp for Bangla-fitted LEACE. INLP shows the same gap; held-out rotation raises overlap and erasure (\cref{tab:app-erasers}; \cref{fig:read-erase}). &
The bound covers binary or pairwise-averaged shares; the multiclass ordering is descriptive. \\
\bottomrule
\end{tabular}
\end{table}

\paragraph{What this study adds.}
The study connects cross-lingual concept sharing to the representation exported across languages: its image, input map and fitted readout. Extending the reading/removal dissociation of \citet{gonen-etal-2022-analyzing} and the estimator comparisons of \citet{naowarat-etal-2026-framework}, the idiom controls show that class-mean sharing persists without illustrative context. The estimator analyses further show that a shared image does not determine transfer: covariance weighting changes the coordinates available to a probe, and regularization moves logistic filters toward class-mean patterns. These distinctions explain why reading, classifier transfer and erasure must be assessed separately.

\paragraph{What is shared.}
The evidence concerns coarse semantic classes. Glosses share more than idiom counterparts, and what survives without illustrative context remains a class-level signal. Idiom-like counterparts support weaker English-direction reading in exploratory comparisons, without isolating idiomaticity from other lexical differences. The class-level association with evaluative meaning disappears after accounting for native recall and within-class variation (\cref{tab:travel,tab:app-human-items}). Encoder rankings vary with split and layer, so they do not establish a causal benefit of translation-pair training (\cref{tab:app-refit,tab:app-fixed-depth}).

\paragraph{What should be measured.}
Cross-lingual concept studies should compare patterns with source-tuned filters, supervised with source-trained
readouts, and balanced with raw accuracy. They should report geometric overlap alongside residual recoverability
after erasure. The idealized bounds explain how reading and erasure can differ; they do not certify a learned
multiclass intervention. An eraser should therefore be evaluated in each language where it is applied, even when
its directions already support a useful transferred readout.

\section{Limitations}
\label{sec:limitations}

\para{Generality and inference.} The study covers one language pair, one corpus export and linear subspaces. The prospective studies re-partition the same corpus under locally frozen plans; they are internal replications. Other robustness analyses are \textit{post hoc}, with multiplicity correction within the stated families where applicable. Refitting the subspaces and probes widens intervals by a median factor of \NzRefitWidth{}, so encoder rankings remain descriptive (\cref{app:refit}).

\para{Labels and human validation.} F is a machine taxonomy of English glosses with \textit{post hoc} class names; its re-seed ARIs are 0.40--0.63. F and FBN favor gloss inputs by construction, whereas DOM and TAG avoid gloss fields but cover only \NxDOMCoverage\% and \NyTagCoverage\% of records.
The 300-item, class-balanced human check uses two independent volunteers, with class-aptness and equivalence agreement of $\kappa=\HvKappaClass$ and $\HvKappaEquiv$; disagreements were not adjudicated. Corpus-label provenance and the export's four sentiment values are documented in \cref{app:taxonomy,app:human}.

\para{Inputs and supervision.} Corpus examples illustrate their idioms and can themselves reveal the class. The idiom-alone analysis removes this illustrative context but uses neutral constructed frames. English counterparts include idioms, proverbs and paraphrases, so counterpart-type associations do not isolate idiomaticity. For classifier transfer, readouts are fitted using English labels at Bangla-validation-selected layers, with target centering and pooled SPACE scaling; the pipeline is therefore not target-label-free. The pooling and alignment analyses measure the effects of their specified inputs and paired data (\cref{app:pooling,app:heldout}).

\para{Readouts and theory.} MLP and codelength analyses support the reading findings, while nonlinear information can remain after native erasure. R-LACE's limited source-language erasure prevents a conclusive transfer comparison. The theoretical results characterize strong-penalty limits, shared-mean covariance models, and binary or pairwise-averaged shares under isotropic Gaussian noise; they do not bound the reported multiclass SSI. They identify mechanisms that can produce the observed distinctions, while covariance-matching ablations leave the cause of LEACE's deficit unresolved. The decoder results are descriptive (\cref{app:decoder}).

\section{Conclusion}
\label{sec:conclusion}

Cross-lingual concept sharing depends on how a subspace is estimated, read and removed. Across six encoders, class-mean directions retain Bangla coarse-class information and beat null controls more consistently than default logistic directions do, with prospective internal support for idioms presented without context. Classifier transfer depends on the input map and regularization, while English-fitted erasure removes substantially less Bangla information than native erasure. The input controls and geometric analysis connect these findings: shared class-mean structure can support reading without supporting the same classifier or effective removal. Evaluating reading, classifier transfer, erasure and overlap separately therefore provides a more informative account of concept reuse across languages.

\subsubsection*{Broader Impact Statement}

We reuse the published Bagdhara resource under its Kaggle CC BY-SA 4.0 license for representation analysis. The export has 10,359 records against the 10,361 that the publication describes, and its sentiment field has four values (152 records are \emph{mixed}) rather than the publication's three; we use the export unchanged. Our only new data are the judgments of two computer-science undergraduates (\cref{app:human}), native Bangla speakers fluent in English and not authors, who were told the study's purpose, agreed to take part and were not paid. Brief written instructions covered their task; no personal information about them was recorded, and no institutional review was sought. The corpus contains culturally situated examples and evaluative or potentially offensive expressions; its metadata about culture and religion are properties of annotated expressions, not inferences about individuals. We have not independently certified the absence of personal information in every example. Our repository omits the raw corpus; its annotation sheets for the 300 judged items and six illustrative records quote corpus text, under the corpus's CC BY-SA 4.0 license. The FBN labels use the encoder of a translation model, under its noncommercial research terms, only to cluster glosses.

The experiments are research diagnostics, not a validated fairness or safety intervention. Linear erasure is also used to remove protected attributes. Our results show that an eraser validated only in English can leave its concept readable in another language, so erasure should be checked in each language where it is meant to hold. Language-specific outcomes should not be read as a ranking of languages or communities. English-derived labels and illustrative contexts may privilege particular semantic distinctions and usage styles. Larger bilingual validation and broader contexts remain necessary before stronger cultural or linguistic claims can be made.

Compute infrastructure, model licenses and budgets are documented in \cref{app:repro}.

\subsubsection*{Code and Data Availability}

Our repository, \coderepo, holds the code of every experiment and analysis, its configurations, the frozen analysis plans with their digests, and the aggregate results behind the paper's tables and result figures. It also holds the volunteers' answer sheets and the provenance records of \cref{app:repro}: the corpus export's checksum, split statistics, model revisions, software versions and layer-selection scores. The corpus (Bagdhara, export of 2026-03-01; \citealp{sakhawat-etal-2026-words}) and the model weights come from their providers; item-level predictions and checkpoints are not included. All runs used one NVIDIA RTX 3090 with 24\,GiB of memory, and the three phases of the discovery campaign took about 9.5 hours (\cref{app:repro}).

\begingroup
\raggedbottom
\bibliography{references2}
\bibliographystyle{tmlr}
\clearpage
\endgroup

\appendix
\section*{Appendix}
\label{app:contents}

\begin{tocbox}
\tocline{app:method}{A}{Methodology in Detail}
\tocsub{app:method-inputs}{A.1}{Inputs, splits and preprocessing}
\tocsub{app:method-estimators}{A.2}{Estimating a subspace (\cref{alg:estimate})}
\tocsub{app:method-readouts}{A.3}{Readouts (\cref{alg:readout})}
\tocsub{app:method-inference}{A.4}{Inference (\cref{alg:inference})}
\tocline{app:theory}{B}{Theory}
\tocsub{app:theory-patterns}{B.1}{Patterns, filters and penalties (\cref{prop:penalty})}
\tocsub{app:theory-portable}{B.2}{Portable and covariance-weighted coordinates (\cref{prop:portable})}
\tocsub{app:theory-read-erase}{B.3}{Reading \textit{versus} erasure (\cref{prop:read-erase})}
\tocsub{app:theory-chance}{B.4}{Overlap by chance}
\tocsub{app:theory-metric}{B.5}{Metrics, oblique erasers and floors}
\tocline{app:qualitative}{C}{Qualitative Analysis}
\tocline{app:data}{D}{Data, Split and Inputs}
\tocline{app:taxonomy}{E}{The Figurative Taxonomy}
\tocline{app:human}{F}{Human Validation}
\tocline{app:results}{G}{Additional Results}
\tocsub{app:measurement}{G.1}{Further measurement diagnostics}
\tocsub{app:claims}{G.2}{Full evidence ledger}
\tocline{app:newexp}{H}{Robustness and Mechanism Analyses}
\tocsub{app:balanced}{H.1}{Metrics with rescored nulls}
\tocsub{app:zero-shot}{H.2}{Zero-shot readout}
\tocsub{app:concepts}{H.3}{Concepts not built from English}
\tocsub{app:leace}{H.4}{The LEACE ablation}
\tocsub{app:pooling}{H.5}{The pooling-matched design}
\tocsub{app:nonlinear}{H.6}{Nonlinear readouts and description length}
\tocsub{app:alignment}{H.7}{Hubness correction and alignment}
\tocsub{app:layers}{H.8}{Layers}
\tocsub{app:heldout}{H.9}{Rotations fitted on held-out pairs}
\tocsub{app:erasers}{H.10}{Other erasers and nonlinear readouts}
\tocsub{app:taxonomies}{H.11}{Re-seeded and tag-built taxonomies}
\tocsub{app:penalty-path}{H.12}{Transfer along the penalty path}
\tocsub{app:idiom-alone}{H.13}{The idiom without its context}
\tocsub{app:idiom-prospective}{H.14}{The idiom alone, prospectively}
\tocsub{app:matched-nulls}{H.15}{A covariance-matched null and PCA}
\tocsub{app:counterpart-type}{H.16}{Which counterparts carry the signal}
\tocsub{app:refit}{H.17}{Intervals that refit the subspaces}
\tocline{app:followup}{I}{Internal Replication and Mechanism Checks}
\tocline{app:decoder}{J}{Decoder}
\tocline{app:robustness}{K}{Robustness and Negative Results}
\tocline{app:identity-audit}{L}{Identity Retrieval Audit}
\tocline{app:repro}{M}{Reproducibility and Resources}
\end{tocbox}

\FloatBarrier
\section{Methodology in Detail}
\label{app:method}

\Cref{alg:estimate,alg:readout,alg:inference} specify estimation, readout and inference. The accompanying explanations connect each step to the theoretical results in \cref{app:theory}.

\begin{table}[t]
\caption{\textbf{Notation} used in the main text, \cref{app:method} and \cref{app:theory}.}
\label{tab:notation}
\centering
\small
\setlength{\tabcolsep}{4pt}
\begin{tabular}{@{}l>{\raggedright\arraybackslash}m{11.6cm}@{}}
\toprule
\textbf{Symbol} & \textbf{Meaning} \\
\midrule
$X\in\R^{n\times D}$ & centered SPACE inputs of one condition; $D$ is the hidden size \\
$e$, $b$ & source (English) and target (Bangla) language or condition \\
$K$, $\pi_k$, $\mu_k$ & number of classes, class priors, class-mean offsets from the grand mean \\
$\Sigma$, $\Sigma_{xz}$ & feature covariance $X^{\top}X/n$ and feature--label cross-covariance $X^{\top}\bar Z/n=[\pi_k\mu_k]_k$ \\
$M$ & subspace rank: 11 for F, 3 for S ($=K-1$) \\
$Q$, $P_S$ & orthonormal basis of a subspace $S$ (an estimator's image) and its orthogonal projector $QQ^{\top}$ \\
$G$, $P$ & input map with $GQ=I_M$, giving compact coordinates $z=Gx$; removed component $P=QG$ \\
$R_{e\to b}$, $R_0$ & target TEST accuracy through a source subspace; majority floor \\
$\theta_i$, $\omega$ & principal angles between two rank-$M$ subspaces; overlap $\omega=\frac1M\sum_i\cos^2\theta_i$ \\
$C$, $\lambda$ & inverse penalty of logistic regression; ridge penalty \\
$\delta$, $c$, $d$ & class difference, $\|P_S\delta\|/\|\delta\|$, and half-separation $\|\delta\|/(2\sigma)$ \\
$U$ & span of the class-mean differences (the COV image at $M=K-1$) \\
\bottomrule
\end{tabular}
\end{table}

\subsection{Inputs, splits and preprocessing}
\label{app:method-inputs}
\para{Steps.} (i)~From each record we build the eight inputs of \cref{tab:conditions}, encode each on its own and mean-pool the hidden states of its target tokens at the selected layer (\cref{app:data}). (ii)~Records that share a normalized idiom form, a listed variant or any identical model input are linked, and whole connected components are assigned to SPACE, PROBE, VALIDATION and TEST. (iii)~A global mean and a global root-mean-square (RMS) scale, fitted on unlabeled SPACE rows of all conditions, standardize every input; each estimator then centers its condition on that condition's SPACE mean.

\para{Rationale.} (ii)~Components are the sampling units: records sharing an input would otherwise carry the same representation into training and evaluation. Assigning whole components to splits prevents this leakage. Bootstrap and permutation inference treats components as the sampling units under the exchangeability assumptions stated in \cref{app:method-inference}.

(iii)~\Cref{prop:penalty,prop:portable} concern centered inputs. Without centering, unequal class priors can introduce a grand-mean component into the logistic limit (\cref{app:theory-patterns}), and a language offset would shift every class in a zero-shot readout. Centering uses no labels. A single scalar scale preserves relative geometry while making a fixed probe penalty comparable across layers and models.

\subsection{Estimating a subspace}
\label{app:method-estimators}

\begin{algorithm}[t]
\caption{Fit an estimator on one condition's SPACE inputs. Returns the image $Q$ (orthonormal, $D\times M$) and the input map $G$ ($GQ=I_M$).}
\label{alg:estimate}
\small
\begin{algorithmic}[1]
\Require centered SPACE inputs $X\in\R^{n\times D}$, labels $y$ with $K$ classes, rank $M$, estimator $\mathcal{E}$
\State $\Sigma\gets X^{\top}X/n$; $\Sigma_{xz}\gets X^{\top}\bar Z/n=[\pi_k\mu_k]_k$ \Comment{$\bar Z$: centered one-hot labels}
\If{$\mathcal{E}=$ COV} $Q\gets$ top-$M$ left singular vectors of $\Sigma_{xz}$; $G\gets Q^{\top}$
\ElsIf{$\mathcal{E}=$ CPCA} $Q\gets$ top-$M$ principal components of the class centroids; $G\gets Q^{\top}$
\ElsIf{$\mathcal{E}=$ MLR} $W\gets\arg\min_W\frac12\|W\|_F^2+C\sum_i\ell(Wx_i,y_i)$, rows normalized; $Q\gets$ top-$M$ right singular vectors of $W$; $G\gets Q^{\top}$
\ElsIf{$\mathcal{E}=$ LDA} $Q\gets$ orthonormalized $S_w^{-1/2}U$, $U$ the top-$M$ left singular vectors of $S_w^{-1/2}[\sqrt{n_k/n}\,\mu_k]_k$; $G\gets Q^{\top}$
\ElsIf{$\mathcal{E}=$ LEACE} $W\gets\Sigma^{-1/2}$ (Moore--Penrose); $U\gets$ top-$M$ left singular vectors of $W\Sigma_{xz}$; $QR\gets W^{+}U$ (reduced QR); $G\gets RU^{\top}W$
\ElsIf{$\mathcal{E}=$ RAND} $Q\gets$ orthonormalized standard Gaussian $D\times M$ matrix; $G\gets Q^{\top}$
\Else\ ($\mathcal{E}=$ PCA) $Q\gets$ top-$M$ eigenvectors of $\Sigma$; $G\gets Q^{\top}$
\EndIf
\State \Return $(Q,G)$; compact coordinates $z=Gx$ for centered $x$; removed component $P=QG$
\end{algorithmic}
\end{algorithm}

\para{Estimator rationale.} \emph{COV}: $\Sigma_{xz}=[\pi_k\mu_k]_k$, so its range is $\spn\{\mu_k\}$. At $M=K-1$, its image is exactly the class-mean span $U$, the smallest subspace whose orthogonal removal equalizes the class means (\cref{lem:containment}); at lower ranks it keeps the directions of largest prior-weighted mean separation (Eckart--Young). \emph{CPCA} reaches the same span at full rank without prior weighting and serves as a check on COV.

\emph{MLR} follows the reference framework's no-intercept, row-normalized logistic estimator. Under a shared-covariance Gaussian model, affine logistic discriminants have weight differences $\Sigma^{-1}(\mu_j-\mu_k)$, motivating the filter interpretation. The strong-penalty limit of the implemented estimator approaches the COV image (\cref{prop:penalty}). \emph{LDA}, Fisher's discriminant, is a second filter, computed in whitened form.

\emph{LEACE} is the least-squares-optimal linear eraser; its removed component is \cref{eq:operator}, $P=QG_\Sigma$, with the generalized-least-squares map $G_\Sigma$ of \cref{prop:portable}. Whitening by the total rather than the within-class covariance gives the same map at full rank (\cref{lem:woodbury}; numerically, to a relative \NxWoodbury). \emph{RAND} and \emph{PCA} are label-free references; RAND also supplies chance overlap (\cref{prop:chance}) and the random-subspace reference distribution.

\para{Why compact coordinates.} By \cref{lem:metric}, an L2 probe on $z=Gx$ searches $\operatorname{range}(G^{\top})$ under the penalty $w^{\top}(G^{\top}G)^{\dagger}w$. For orthonormal rows ($G=Q^{\top}$) that penalty is the Euclidean one on the projected input, so a probe on $z$ equals a probe on $Px$: the coordinates change the dimension, not the problem. For centered LEACE inputs, $Px=Q[RU^{\top}Wx]$, so the triangular factor makes $\|z\|=\|Px\|$, and the probe sees the Euclidean geometry of LEACE's ambient output. The complementary eraser uses $I-P$; a constant centering difference is absorbed by the probe's unpenalized intercept.

\para{Same image at full rank.} $\Sigma_{xz}$ has rank at most $K{-}1$, and its column space is spanned by the class-mean offsets. At $M=K{-}1$, $\spn(U)=\spn(W\Sigma_{xz})$, so the LEACE image is $\spn(W^{+}W\Sigma_{xz})=\spn(\Sigma_{xz})$ on the support of $\Sigma$, which is the COV subspace and equals the CPCA subspace. COV and CPCA project orthogonally onto that image. LEACE uses an oblique input map with row space $\spn(\Sigma^{\dagger}\Sigma_{xz})$; on its supported subspace its removed component equals
\begin{equation}
P_\Sigma=Q(Q^{\top}\Sigma^{\dagger}Q)^{-1}Q^{\top}\Sigma^{\dagger}
\label{eq:operator}
\end{equation}
whenever $Q^{\top}\Sigma^{\dagger}Q$ is invertible on its support. A shared image therefore implies neither equal projected observations, nor equal probe geometry (\cref{lem:metric}), nor a shared erasure kernel, nor equal portability (\cref{prop:portable}).

\subsection{Readouts}
\label{app:method-readouts}

\begin{algorithm}[t]
\caption{Read, transfer and erase through a source fit $(Q_e,G_e)$ on target $b$, against the target's own fit $(Q_b,G_b)$.}
\label{alg:readout}
\small
\begin{algorithmic}[1]
\Require fits $(Q_e,G_e)$, $(Q_b,G_b)$; PROBE, VALIDATION and TEST inputs and labels of $e$ and $b$
\State $h\gets$ L2 logistic probe ($C{=}1$) on $\{(G_ex_i,y_i)\}_{i\in b\text{-PROBE}}$; $R_{e\to b}\gets$ accuracy of $h$ on $b$-TEST \Comment{read}
\State $R_{b\to b}$ likewise with $G_b$; $A\gets$ accuracy of the probe on the full $x$; $R_0\gets$ accuracy of the $b$-PROBE majority class
\State $\ssi\gets(R_{e\to b}-R_0)/(R_{b\to b}-R_0)$ if $R_{b\to b}-R_0>2$\pp
\State $h_e\gets$ probe on $\{(G_ex_i,y_i)\}_{i\in e\text{-PROBE}}$, $C$ tuned on $e$-VALIDATION; $\ssi_{\mathrm{zs}}$ of $h_e$ on $b$-TEST, same denominator \Comment{transfer}
\State $a_e\gets$ accuracy of a $b$ probe retrained on $(I-Q_eG_e)x$; $a_b$ likewise with $(Q_b,G_b)$; erase share $\gets(A-a_e)/(A-a_b)$ \Comment{erase}
\State $\sigma_i\gets$ singular values of $Q_e^{\top}Q_b$; shared dimensions $\gets\sum_i\sigma_i^2$; $\omega\gets\frac1M\sum_i\sigma_i^2$ \Comment{overlap}
\State top-1 retrieval of each paired item by cosine after per-condition centering (cross-domain similarity local scaling, CSLS, after Procrustes in \cref{app:alignment}) \Comment{instances}
\end{algorithmic}
\end{algorithm}

\para{Readouts and normalization.} For shared-covariance Gaussian classes, the Bayes rule on $z$ is affine and the logistic model provides a well-specified model for the optimal readout. This motivates the linear probes; \cref{prop:read-erase} analyzes their Bayes-optimal counterparts under its stated assumptions. Probes are trained on labeled target-language PROBE items. RAND compares their performance with what target supervision extracts from a random rank-$M$ projection.

Dividing by the native gain gives $\ssi=0$ at the floor and 1 at native retention. \Cref{prop:read-erase} bounds binary and pairwise-averaged analogs below by $\omega$, motivating the separate multiclass diagnostic. Floors can reorder ratios (\cref{app:theory-metric}), so the primary reading families are also evaluated by balanced accuracy (floor $1/K$) and macro-F1 (\cref{app:balanced,app:idiom-alone}).

\para{Transfer and erasure.} A target-trained readout can recalibrate the supplied directions; a source-trained probe tests whether the source decision rule applies. \Cref{prop:portable} and \cref{lem:filter} show why pattern coordinates can be less sensitive to source covariance than covariance-weighted coordinates. Per-language centering supplies the label-free adaptation assumed by the proposition; English-mean centering omits this adaptation.

By \cref{lem:containment}, removing $S$ orthogonally equalizes class means exactly when $U\subseteq S$. Under its assumptions, \cref{prop:read-erase} bounds binary and pairwise-averaged erase shares by $\omega$. Retraining the probe after removal estimates residual recoverability under the specified linear-probe procedure.

\para{Overlap and retrieval.} Image overlap uses orthogonal projectors:
\[
\sum_i\cos^2\theta_i=\|Q_e^{\top}Q_b\|_F^2
=\tr[(Q_eQ_e^{\top})(Q_bQ_b^{\top})].
\]
Its chance value is $M^2/D$ (\cref{prop:chance}). This geometric quantity enters the bound in \cref{prop:read-erase}. Retrieval asks whether individual items correspond, a question distinct from whether class information occupies shared directions; \cref{app:alignment} uses cross-domain similarity local scaling (CSLS) after Procrustes alignment.

\subsection{Inference}
\label{app:method-inference}

\begin{algorithm}[t]
\caption{Nulls, $p$-values and intervals for a statistic $T$ (SSI or zero-shot SSI) and for paired contrasts.}
\label{alg:inference}
\small
\begin{algorithmic}[1]
\Require observed $T$; draws $B$ (1,999 planned; 999 or 199 exploratory); a family of tests
\For{$j=1,\dots,B$}
\State RAND: draw a random orthonormal $Q$ of rank $M$; $T_j^{\mathrm{R}}\gets T$ with $G=Q^{\top}$
\State PERM: permute labels across SPACE components of equal size; refit $\mathcal{E}$; $T_j^{\mathrm{P}}\gets T$
\EndFor
\State $p\gets(1+\#\{j:T_j\ge T\})/(B+1)$ for each null; Holm-adjust within the family
\State SSI interval: resample TEST components 1,000 times, models fixed; percentile interval
\State paired contrasts: per-item correctness differences summed by component; flip component signs 19,999 times; one-sided $p$ with $+1$; percentile interval from 5,000 component resamples
\end{algorithmic}
\end{algorithm}

\para{Reference distributions.} RAND compares the observed readout with the reference distribution obtained from rank-matched random subspaces. Interpreting its Monte Carlo tail probability as a calibrated null-test $p$-value requires the observed statistic to be exchangeable with, or stochastically dominated by, that reference under the null. Because a target-trained probe can recover class information from a random projection, RAND calibrates target supervision rather than chance-level prediction.

PERM uses a distinct label-permutation null: permuting labels across exchangeable units leaves the estimator's output distribution unchanged under that null. Permuting whole components of equal size preserves dependence among labels attached to records sharing an idiom (77\% of SPACE mass is movable). RAND asks whether the fitted directions outperform random ones; PERM asks whether true labels outperform permuted labels under the same estimator.

\para{Statistics and multiplicity.} Native accuracy and floor are fixed across the observed directions and every null draw. SSI is therefore the same increasing affine function of raw accuracy for every draw, giving identical $p$-values for the two statistics; it rescales effect magnitudes (\cref{tab:app-absolute}). Under the stated null assumptions, the plus-one Monte Carlo estimate is valid and never zero. The smallest attainable planned Holm-adjusted $p$ is $12/2000=.006$. Holm's procedure controls the familywise error rate under arbitrary dependence among tests.

\para{Intervals and paired contrasts.} Bootstrap intervals resample components and condition on the fitted models; resampling individual records would ignore within-component dependence. Wild-sign calibration assumes independent component sums that are symmetric with mean zero at the null boundary, under which sign flips leave their distribution invariant. It is not an exact randomization guarantee for observational inputs.

\para{Planned and exploratory families.} The planned family is fixed in \cref{sec:measures}. For the PERM null, MLR is refitted on permuted labels; matched-control intervals resample components of the paired correctness differences (2,000 draws). The \textit{post hoc} families of \cref{tab:map} are Holm-corrected within each column pair; layer-wise tests (199 draws per null) are unadjusted and descriptive.

\para{Item-level regressions.} We fit logistic regressions of English-direction correctness on standardized covariates: native correctness, counterpart transparency, idiom length, token count, rarity, and cultural, religious, historical and rural flags. Standard errors are clustered by component, the fit uses complete cases ($n=2{,}134$), and Holm correction runs over all 144 coefficients. These regressions are \textit{post hoc} and descriptive.

\FloatBarrier
\section{Theory}
\label{app:theory}

This appendix proves the three main-text propositions and their supporting results. \Cref{app:theory-patterns} derives the strong-penalty pattern limit; \cref{app:theory-portable} characterizes how class-mean and covariance-weighted coordinates can differ in portability; \cref{app:theory-read-erase} bounds reading and erasure; \cref{app:theory-chance} establishes chance overlap; and \cref{app:theory-metric} describes the function class and penalty induced by an input map.

\subsection{Patterns, filters and penalties}
\label{app:theory-patterns}

\begin{proof}[Proof of \cref{prop:penalty}]
Let $F(W)=\frac12\|W\|_F^2+C\,L(W)$ with $L(W)=\sum_{i=1}^n\ell(Wx_i,y_i)$ and $\ell(z,y)=-z_y+\log\sum_k e^{z_k}$. $F$ is strongly convex, so its minimizer $W_C$ is unique and satisfies $\nabla F(W_C)=0$, that is, $W_C=C\,\Gamma(W_C)$ with $\Gamma(W)=\sum_i\big(e_{y_i}-p_i(W)\big)x_i^{\top}$ and $p_i(W)=\operatorname{softmax}(Wx_i)$. The softmax is Lipschitz, so $\|\Gamma(W)-\Gamma(0)\|\le L_\Gamma\|W\|$ near $0$, and $\|W_C\|\le C\big(\|\Gamma(0)\|+L_\Gamma\|W_C\|\big)$ gives $\|W_C\|=O(C)$. Hence $W_C=C\,\Gamma(0)+O(C^2)$. Because $p_i(0)=\mathbf{1}/K$, row $k$ of $\Gamma(0)$ is $\sum_{i:y_i=k}x_i^{\top}-\frac1K\sum_ix_i^{\top}=n\pi_k\mu_k^{\top}$ for centered inputs.

The MLR estimator normalizes the rows of $W_C$ to unit length, which leaves each row's direction and therefore the row space unchanged. The normalized rows converge to $\mu_k/\|\mu_k\|$, a matrix whose rank is $K-1$ when the class means are affinely independent ($\sum_k\pi_k\mu_k=0$ removes one dimension) and whose row space is $\spn\{\mu_k\}=\operatorname{range}(\Sigma_{xz})$, the COV image at $M=K-1$. Since the limiting matrix has a gap between its $(K{-}1)$-th and $K$-th singular values, the $\sin\Theta$ perturbation bound for singular subspaces carries the leading $K-1$ right singular vectors to that image.
\end{proof}

\para{The ridge path.} Ridge regression shows the same transition in closed form: $W_\lambda=(\Sigma+\lambda I)^{-1}\Sigma_{xz}$ converges to $\Sigma^{\dagger}\Sigma_{xz}$ as $\lambda\to0$, while the rescaled coefficients $\lambda W_\lambda$ converge to $\Sigma_{xz}$ as $\lambda\to\infty$. For centered labels, setting the gradient of $\frac1n\|\bar Z-XW\|_F^2+\lambda\|W\|_F^2$ to zero gives $(\Sigma+\lambda I)W_\lambda=\Sigma_{xz}$. As $\lambda\to0$, $W_\lambda\to\Sigma^{\dagger}\Sigma_{xz}$, because $\Sigma_{xz}$ lies in the range of $\Sigma$; as $\lambda\to\infty$, $\lambda W_\lambda=(\Sigma/\lambda+I)^{-1}\Sigma_{xz}\to\Sigma_{xz}$.

\begin{reading}
In the eigenbasis $\Sigma=\sum_js_jv_jv_j^{\top}$, the ridge solution is $W_\lambda=\sum_jv_jv_j^{\top}\Sigma_{xz}/(s_j+\lambda)$. A filter amplifies the low-variance directions by $1/s_j$; a pattern weights all directions equally. Two languages can share their class means exactly and still have different filters, because their low-variance directions differ (\cref{fig:app-theory-patterns}a). Without centering, and with $\mu_k$ still denoting the class offset from the grand mean, row $k$ becomes
\[\Gamma_k(0)=n\pi_k\mu_k^{\top}+n(\pi_k-1/K)\bar x^{\top}.\]
Thus unequal class priors can introduce a grand-mean component into the limit. At finite $C$, the measured path need not reach this limit. The seed-1 path is consistent with the proposition: the MLR subspace's overlap with the COV image rises monotonically in all \FuCells{} model--label cells, reaching \FuOverlapFmin{} of 11 dimensions for F at $C=.001$ (\cref{fig:estimators}e, \cref{fig:app-followup}c).
\end{reading}

How far can a filter stray from its pattern? For a positive definite $\Sigma$ with eigenvalues in $[s_{\min},s_{\max}]$ and condition number $\kappa=s_{\max}/s_{\min}$,
\begin{equation}
\min_{\mu\neq0}\cos\angle\big(\Sigma^{-1}\mu,\mu\big)=\frac{2\sqrt{\kappa}}{1+\kappa}.
\label{eq:antieigen}
\end{equation}
Writing $w_j=(v_j^{\top}\mu)^2/\|\mu\|^2$ and $a_j=1/s_j$, the cosine equals $\sum_jw_ja_j/\big(\sum_jw_ja_j^2\big)^{1/2}$, the ratio of the first moment to the root second moment of a distribution on $[1/s_{\max},1/s_{\min}]$. A two-point distribution on the endpoints minimizes this ratio, which gives \cref{eq:antieigen}. Contextual representations are strongly anisotropic \citep{ethayarajh-2019-contextual,timkey-van-schijndel-2021-bark}, and at the selected layer XLM-R's participation-ratio dimension is 15.2 of 1,024 (\cref{tab:e4}), so the bound allows filters to stand nearly orthogonal to the patterns they discriminate.

\subsection{Portable and covariance-weighted coordinates}
\label{app:theory-portable}

\Cref{prop:portable} treats compact coordinates as estimates. Write a centered input of language $\ell$ as $x=Qa+\varepsilon$, where $a\in\R^M$ locates its class mean along the image and $\varepsilon\sim\mathcal{N}(0,\Sigma_\ell)$ is within-class noise. A linear map $G$ estimates $a$ without bias, $\E[Gx\mid a]=a$, exactly when $GQ=I_M$, which is the condition on every input map of \cref{sec:operations}. The COV map $Q^{\top}$ is then the ordinary least-squares estimate of $a$, and LEACE's map is the generalized least-squares estimate under the covariance it was fitted with.

\begin{proof}[Proof of \cref{prop:portable}]
(a)~Write $m_k=Qa_k$. For any $G$ with $GQ=I_M$, the class means of $z=Gx$ are $Gm_k=a_k$ in both languages. Let $H=G-G_\ell$; since $G_\ell Q=(Q^{\top}\Sigma_\ell^{-1}Q)^{-1}Q^{\top}\Sigma_\ell^{-1}Q=I_M$, we have $HQ=0$. Because $G_\ell\Sigma_\ell=(Q^{\top}\Sigma_\ell^{-1}Q)^{-1}Q^{\top}$, the cross term $G_\ell\Sigma_\ell H^{\top}=(Q^{\top}\Sigma_\ell^{-1}Q)^{-1}(HQ)^{\top}$ vanishes, and $G_\ell\Sigma_\ell G_\ell^{\top}=(Q^{\top}\Sigma_\ell^{-1}Q)^{-1}$. Hence
\[\begin{aligned}G\Sigma_\ell G^{\top}&=(Q^{\top}\Sigma_\ell^{-1}Q)^{-1}+H\Sigma_\ell H^{\top}\\&\succeq(Q^{\top}\Sigma_\ell^{-1}Q)^{-1},\end{aligned}\]
with equality iff $H\Sigma_\ell H^{\top}=0$, that is, $H=0$, since $\Sigma_\ell\succ0$. This is Aitken's generalized Gauss--Markov theorem. LEACE fitted in $\ell$ removes $P_\Sigma=QG_\Sigma$ (\cref{eq:operator}), where $\Sigma$ is the total covariance of $\ell$; its between-class part $\sum_k\pi_km_km_k^{\top}$ lies in $\operatorname{range}(Q)$, so $G_\Sigma=G_\ell$ by \cref{lem:woodbury}. Its compact coordinates $RU^{\top}Wx$ satisfy $Q'[RU^{\top}Wx]=Px$ for the orthonormal basis $Q'$ of the QR step (\cref{alg:estimate}), so they equal $Q'^{\top}QG_\ell x$, an orthogonal transformation of $G_\ell x$ that changes nothing a probe sees. The pattern map $Q^{\top}$ satisfies $Q^{\top}Q=I_M$ and involves no covariance. \Cref{ex:portable} shows that $G_e$ can be arbitrarily poor in $b$.

(b)~With centered inputs and equal priors, the class means are $\pm\delta/2$ in both languages, so for any $w$ with $w^{\top}\delta>0$ the projected classes are $\mathcal{N}(\pm w^{\top}\delta/2,\,w^{\top}\Sigma_bw)$ in $b$, and $\operatorname{sign}(w^{\top}x)$ has accuracy $\Phi\big(f(w)/2\big)$ with $f(w)=w^{\top}\delta/\sqrt{w^{\top}\Sigma_bw}$. For the pattern, $w=\delta$, this is the stated value, free of $\Sigma_e$. For the filter $w=\Sigma_e^{-1}\delta$: first, as $\Sigma_e$ ranges over positive definite matrices, $\Sigma_e^{-1}\delta$ ranges over the whole half-space $\mathcal{H}=\{w:w^{\top}\delta>0\}$. Every $\Sigma_e\succ0$ gives $\delta^{\top}\Sigma_e^{-1}\delta>0$; conversely, for $w\in\mathcal{H}$ the matrix $A=ww^{\top}/(w^{\top}\delta)+I-\delta\delta^{\top}/\|\delta\|^2$ satisfies $A\delta=w$ and is positive definite, since $v^{\top}Av=(w^{\top}v)^2/(w^{\top}\delta)+\|v\|^2-(\delta^{\top}v)^2/\|\delta\|^2$, where the last two terms are nonnegative by Cauchy--Schwarz and vanish only for $v\parallel\delta$, where the first is positive; take $\Sigma_e=A^{-1}$. Second, $f$ is continuous on the connected set $\mathcal{H}$, bounded by $\sqrt{\delta^{\top}\Sigma_b^{-1}\delta}$ (Cauchy--Schwarz in the $\Sigma_b$ inner product) with equality at $w=\Sigma_b^{-1}\delta$, and tends to $0$ as $w$ approaches a nonzero point with $w^{\top}\delta=0$. By the intermediate value theorem $f$ takes every value in $\big(0,\sqrt{\delta^{\top}\Sigma_b^{-1}\delta}\,\big]$, and $\Phi(f/2)$ every value in the stated interval.
\end{proof}

\begin{lemma}[Total and within-class whitening agree]\label{lem:woodbury}
Let $\Sigma=\Sigma_w+QBQ^{\top}$ with $\Sigma_w\succ0$ and $B\succeq0$. Then $(Q^{\top}\Sigma^{-1}Q)^{-1}Q^{\top}\Sigma^{-1}=(Q^{\top}\Sigma_w^{-1}Q)^{-1}Q^{\top}\Sigma_w^{-1}$.
\end{lemma}

\begin{proof}
Let $A=Q^{\top}\Sigma_w^{-1}Q$. For $B\succ0$, the Woodbury identity gives $\Sigma^{-1}=\Sigma_w^{-1}-\Sigma_w^{-1}Q(B^{-1}+A)^{-1}Q^{\top}\Sigma_w^{-1}$, so $Q^{\top}\Sigma^{-1}=TQ^{\top}\Sigma_w^{-1}$ with the invertible $T=I-A(B^{-1}+A)^{-1}=B^{-1}(B^{-1}+A)^{-1}$. Then $Q^{\top}\Sigma^{-1}Q=TA$ and $(Q^{\top}\Sigma^{-1}Q)^{-1}Q^{\top}\Sigma^{-1}=A^{-1}T^{-1}TQ^{\top}\Sigma_w^{-1}$. Both sides are continuous in $B$, so the identity extends to singular $B\succeq0$.
\end{proof}

\begin{lemma}[An English probe returns English's filter]\label{lem:filter}
Let Gaussian classes in $e$ have centered means $m_k$ and shared covariance $\Sigma_e\succ0$, and let $z=Gx$ with $G$ of full row rank. If $\Sigma_e^{-1}m_k\in\operatorname{range}(G^{\top})$ for every $k$, the Bayes linear rule fitted on $z$ in $e$ computes the same scores as $e$'s full-space Bayes rule. The condition holds for LEACE's map $G_e$, for an orthonormal basis of any subspace that contains $e$'s filters, and for the full space. On the pattern map $Q^{\top}$ the rule's weights are $Q(Q^{\top}\Sigma_eQ)^{-1}Q^{\top}m_k$, and for two classes they are proportional to $\delta$.
\end{lemma}

\begin{proof}
The Bayes rule on $z$ scores class $k$ by $v_k^{\top}z-\frac12v_k^{\top}Gm_k+\log\pi_k$ with $v_k=(G\Sigma_eG^{\top})^{-1}Gm_k$. Write $u_k=\Sigma_e^{-1}m_k=G^{\top}c_k$. Then $G\Sigma_eG^{\top}c_k=G\Sigma_eu_k=Gm_k$, so $c_k=v_k$ and $G^{\top}v_k=u_k$: the weights on $x$ are $\Sigma_e^{-1}m_k$, and the offsets $-\frac12u_k^{\top}m_k+\log\pi_k$ are the full-space ones. For $G_e$, $\operatorname{range}(G_e^{\top})=\Sigma_e^{-1}\operatorname{range}(Q)\ni\Sigma_e^{-1}m_k$. For $G=Q^{\top}$, $G^{\top}v_k=Q(Q^{\top}\Sigma_eQ)^{-1}Q^{\top}m_k$; with two classes, $Q=\delta/\|\delta\|$ and the difference between the two class weights is a positive multiple of $\delta$.
\end{proof}

\begin{lemma}[Patterns with partly shared means]\label{lem:partial}
Let two equiprobable classes have centered mean differences $\delta_e$ in $e$ and $\delta_b$ in $b$, with $\delta_e^{\top}\delta_b>0$ and covariance $\Sigma_b\succ0$ in $b$. The English pattern rule $\operatorname{sign}(\delta_e^{\top}x)$ has accuracy $\Phi\!\left(\frac{\delta_e^{\top}\delta_b}{2\sqrt{\delta_e^{\top}\Sigma_b\delta_e}}\right)$ in $b$, which depends on how far the two languages' class means align but not on $\Sigma_e$; the English filter $\operatorname{sign}(\delta_e^{\top}\Sigma_e^{-1}x)$ has accuracy $\Phi\!\left(\frac{\delta_e^{\top}\Sigma_e^{-1}\delta_b}{2\sqrt{\delta_e^{\top}\Sigma_e^{-1}\Sigma_b\Sigma_e^{-1}\delta_e}}\right)$, which depends on $\Sigma_e$ in both terms. Through the pattern map $Q_e^{\top}$, the Bangla class means are $Q_e^{\top}m_k^{b}$: the part of Bangla's class differences that lies in the English image, an average squared projected share $\omega$ when the normalized differences are isotropic (\cref{prop:read-erase}).
\end{lemma}

\begin{proof}
For any nonzero $w$, the Bangla classes project to $\mathcal{N}(\pm w^{\top}\delta_b/2,w^{\top}\Sigma_bw)$, so the classifier $\operatorname{sign}(w^{\top}x)$ has accuracy $\Phi\!\left(\frac{w^{\top}\delta_b}{2\sqrt{w^{\top}\Sigma_bw}}\right)$; take $w=\delta_e$ or $w=\Sigma_e^{-1}\delta_e$. The class means of $Q_e^{\top}x$ are $Q_e^{\top}m_k^{b}$ by linearity, and $Q_e^{\top}$ involves no covariance. The share follows from $\E[c^2]=\omega$ in \cref{prop:read-erase}, with $U$ the span of Bangla's class differences and $S$ the English image.
\end{proof}

\Cref{prop:portable}(b) is the case $\delta_b=\delta_e$. At fixed target means and covariance, the pattern rule is independent of source covariance, whereas a source filter can transfer poorly even when the class means coincide. Across the twelve model--label cells, zero-shot SSI through COV indeed rises with the overlap of the English and Bangla images (Spearman $\rho=\NxOverlapZsRho$; supervised SSI $\rho=\NxOverlapSupRho$), and the rotated directions of \cref{app:alignment}, which raise the overlap, extend the same trend ($\rho=\NxOverlapZsRhoAll$ over 24 cells). The cells are not independent, so these correlations are descriptive.

\begin{example}[Covariance-weighted coordinates fail to travel]\label{ex:portable}
Let $D=2$, $Q=e_1$, $\Sigma_e=\left(\begin{smallmatrix}1&\rho\\\rho&1\end{smallmatrix}\right)$ with $|\rho|<1$, and $\Sigma_b=\operatorname{diag}(1,s^2)$. Then $G_e=(1,\,-\rho)$. In $e$, its noise variance is $1-\rho^2$, below the pattern map's $1$; in $b$, it is $G_e\Sigma_bG_e^{\top}=1+\rho^2s^2$, against $1$ for both $Q^{\top}$ and $b$'s own $G_b=e_1^{\top}$. The English-optimal coordinate gains a factor $1/(1-\rho^2)$ in signal-to-noise in the source language and loses a factor $1+\rho^2s^2$ in the target language, without bound as $s$ grows.
\end{example}

\begin{reading}
For two classes, a probe on coordinates with class means $\pm a/2$ and noise $V$ has Bayes accuracy $\Phi\big(\frac12\sqrt{a^{\top}V^{-1}a}\big)$, which increases as $V$ decreases in the Loewner order. In its own language LEACE's coordinates therefore lose nothing against the full space ($a^{\top}Q^{\top}\Sigma_\ell^{-1}Qa=\delta^{\top}\Sigma_\ell^{-1}\delta$), whereas in the other language they incur additional coordinate noise $H\Sigma_bH^{\top}$ with $H=G_e-G_b$. The measurements are consistent with this possible failure mechanism: English LEACE coordinates read Bangla worse than COV's in \NxLeaceCrossWorse{} of \NxLeaceCrossOf{} cells, and the deficit persists under the tested covariance-estimation changes (\cref{fig:app-leace}). \Cref{lem:filter} supplies a common covariance-shift mechanism consistent with their weak raw-accuracy transfer (\cref{tab:app-zero-shot}); it does not establish a unique explanation. On the pattern subspace, by contrast, the English rule depends on $\Sigma_e$ only through the $M\times M$ matrix $Q^{\top}\Sigma_eQ$. The proposition assumes exactly shared class means.

Measured overlaps are partial: with Bangla offsets $m_k^{b}=Qa_k+r_k$ and $r_k\perp\operatorname{range}(Q)$, the pattern map discards $r_k$ ($Q^{\top}r_k=0$), whereas an oblique map also carries $Gr_k$ into its coordinates, a term that depends on the covariance it uses. This term does not by itself explain the decrease in LEACE retention under pooled or Bangla covariance (\expt{3}). LEACE's native advantage, predicted in the population under the proposition's assumptions, appears in \NxLeaceNativeBetter{} of 24 empirical cells.
\end{reading}

\subsection{Reading \texorpdfstring{\textit{versus}}{versus} erasure}
\label{app:theory-read-erase}

We first state the distribution-free core and then quantify it.

\begin{lemma}[Reading needs an angle; erasure needs containment]\label{lem:containment}
Let $x$ have class-conditional means $\mu_k$ with difference span $U$, and let $S$ be a subspace. Some linear classifier on $P_Sx$ outperforms the optimal constant predictor in expected convex loss only if $P_SU\neq\{0\}$, and removing $S$ orthogonally makes the class-conditional means of $(I-P_S)x$ equal if and only if $U\subseteq S$.
\end{lemma}

\begin{proof}
The class-conditional means of $P_Sx$ are $P_S\mu_k$, which are all equal exactly when $P_S(\mu_j-\mu_k)=0$ for all $j,k$, that is, when $P_SU=\{0\}$. LEACE's authors show that a representation is linearly guarded, so that no linear classifier improves on the optimal constant predictor in expected convex loss, exactly when its class-conditional means are equal. Likewise, the means of $(I-P_S)x$ are equal exactly when $(I-P_S)(\mu_j-\mu_k)=0$ for all $j,k$, that is, when $U\subseteq S$.
\end{proof}

Reading requires a nonzero projection of the class-mean span onto $S$, whereas erasure requires $U\subseteq S$; for subspaces of equal dimension, containment means equality. \Cref{prop:read-erase} quantifies the region between these extremes.

\begin{proof}[Proof of \cref{prop:read-erase}]
Let $x\mid y=\pm1\sim\mathcal{N}(\pm\delta/2,\sigma^2I)$ with equal priors. For an orthogonal projector $P$, the features $Px$ are Gaussian on the range of $P$ with means $\pm P\delta/2$ and isotropic covariance $\sigma^2P$, so the Bayes rule is linear, $\operatorname{sign}(\delta^{\top}Px)$, with accuracy $\Phi(\|P\delta\|/(2\sigma))$. With $d=\|\delta\|/(2\sigma)$, $c=\|P_S\delta\|/\|\delta\|$ and $s=\|(I-P_S)\delta\|/\|\delta\|=\sqrt{1-c^2}$, the native, read and erased accuracies are $\Phi(d)$, $\Phi(cd)$ and $\Phi(sd)$, and chance is $\frac12$.

Let $g(t)=\Phi(t)-\frac12$. On $t\ge0$, $g$ is concave with $g(0)=0$, so $g(\alpha t)\ge\alpha g(t)$ for $\alpha\in[0,1]$. Therefore the read share satisfies $g(cd)/g(d)\ge c$, and the erase share satisfies $\big(g(d)-g(sd)\big)/g(d)\le1-s$. Finally $1-s\le1-s^2=c^2\le c$ because $s,c\in[0,1]$.

For the averaged statement, let $\hat\delta_{jk}$ be the normalized differences of $K$ class means with weights $w_{jk}$ summing to one, and suppose they are isotropic in $U$: $\sum_{j<k}w_{jk}\hat\delta_{jk}\hat\delta_{jk}^{\top}=P_U/M$. (For classes whose means form a regular simplex, equal weights satisfy this.) Then $\E[c^2]=\sum w_{jk}\|P_S\hat\delta_{jk}\|^2=\tr(P_SP_U)/M=\frac1M\sum_i\cos^2\theta_i=\omega$, because the nonzero eigenvalues of $P_UP_SP_U$ are the $\cos^2\theta_i$. Applying the pairwise bounds to every pair and averaging, $\E[\text{read share}]\ge\E[c]\ge\E[c^2]=\omega$ and $\E[\text{erase share}]\le\E[1-s]\le\E[c^2]=\omega$.
\end{proof}

\begin{reading}
Two regimes bracket the bounds. When the native signal is weak ($d\to0$), $g$ is nearly linear and the shares approach $c$ and $1-\sqrt{1-c^2}$; the latter is approximately $c^2/2$ when $c$ is also small. When the signal is strong ($d\to\infty$), any $c>0$ makes the read share approach one while the erase share approaches zero unless $c=1$. The averaged bound uses $\E[c]\ge\E[c^2]$. When every normalized class difference meets $S$ at the same angle, $\E[c]=\sqrt{\omega}$ gives a tighter lower bound. The measured multiclass read shares exceed $\sqrt{\omega}$ for \ThSqrtF{} of 6 encoders on F and \ThSqrtS{} of 6 on S. The proposition assumes isotropic noise.

For anisotropic noise, analogous bounds require whitening the data and defining both projections orthogonally in that whitened space; they do not directly bound the original Euclidean projections. The proposition concerns the removed subspace, not how an eraser finds it. INLP and R-LACE remove orthogonal projections, and an eraser that guards its own language contains that language's class-mean span (\cref{lem:containment}). Under the binary or pairwise assumptions above, overlap with the target class-mean span bounds the corresponding reading and erasure shares. For a uniformly random rank-$R$ removed subspace, expected normalized overlap with a fixed target span is $R/D$ (the same rotation-invariance argument as \cref{prop:chance}). This is a rank-matched reference, not a guarantee for adaptively learned directions. English-fitted INLP removes about a quarter of the space yet achieves limited Bangla erasure (\cref{app:erasers}, \cref{fig:app-inlp}).
\end{reading}

\subsection{Overlap by chance}
\label{app:theory-chance}

\begin{proposition}[Chance overlap]\label{prop:chance}
If $S$ is a uniformly random $M$-dimensional subspace of $\R^D$ and $U$ is a fixed $M$-dimensional subspace, then $\E\big[\sum_i\cos^2\theta_i\big]=M^2/D$, so $\E[\omega]=M/D$.
\end{proposition}

\begin{proof}
$\sum_i\cos^2\theta_i=\|Q_U^{\top}Q_S\|_F^2=\tr(P_UP_S)$. By rotation invariance $\E[P_S]=\frac MDI$, so $\E\,\tr(P_UP_S)=\tr(P_U)M/D=M^2/D$.
\end{proof}

At $M=11$ the expectation is 0.12 for $D=1{,}024$ and 0.16 for $D=768$, the random reference in \cref{fig:gap}d and \cref{fig:estimators}c. MLR subspaces stay near this reference across languages (0.14--0.20), whereas COV subspaces share 1.2--3.6 dimensions.

\subsection{Metrics, oblique erasers and floors}
\label{app:theory-metric}

\begin{lemma}[A probe sees a metric]\label{lem:metric}
An L2-penalized linear probe on $z=Bx$ with weights $v$ computes the score $w^{\top}x$ with $w=B^{\top}v$. Its function class is $\operatorname{range}(B^{\top})$ and its penalty is $\min\{\|v\|^2:B^{\top}v=w\}=w^{\top}(B^{\top}B)^{\dagger}w$.
\end{lemma}

\begin{proof}
The minimum-norm solution of $B^{\top}v=w$ is $v=(B^{\top})^{\dagger}w$, with squared norm $w^{\top}B^{\dagger}(B^{\dagger})^{\top}w=w^{\top}(B^{\top}B)^{\dagger}w$.
\end{proof}

The orthogonal projector $QQ^{\top}$ and LEACE's oblique operator $P_\Sigma=Q(Q^{\top}\Sigma^{\dagger}Q)^{-1}Q^{\top}\Sigma^{\dagger}$ share their image, $\operatorname{range}(Q)$, but their row spaces are $\operatorname{range}(Q)$ and $\operatorname{range}(\Sigma^{\dagger}Q)$. By \cref{lem:metric}, probes on the two outputs search different function classes under different penalties, so changing $\Sigma$ with $Q$ fixed can change retention without changing the image. \expt{3} isolates exactly this effect: with the English image fixed, pooled or Bangla covariance changes LEACE retention in every comparison, and lowers it in all \FuRescueFail{} (\cref{tab:e3}).

\para{Ratios and floors.} SSI divides by the target's native above-floor accuracy, so two groups can order differently by accuracy and by SSI when their floors differ. Transparent-counterpart idioms are read more accurately than opaque ones in every encoder but have a lower SSI, because their majority floor is 23.7\% against 8.6\% (\cref{fig:app-floors}). Reporting floors beside SSI prevents this reversal from being read as a difference in sharing.
\FloatBarrier
\section{Qualitative Analysis}
\label{app:qualitative}

The examples illustrate the input contrasts and class-level results. Bangla is set in Kalpurush; the authors' romanizations and literal translations serve as reading aids. Item-level model outcomes are reported only in \cref{tab:app-qual-cases}.

\begin{table}[t]
\caption{\textbf{One gist, many images.} Bangla idioms from the corpus with their first-listed English counterpart, our literal translation, how far the two share an image (our reading), and the record's figurative class F (all ten are negative in sentiment except \bnrom{nake}, which is positive). Imagery ranges from identical to absent; \cref{tab:classes} gives each class's English-direction recall. Illustrative examples; Bagdhara, CC BY-SA 4.0.}
\label{tab:app-qual-pairs}
\centering
\small
\setlength{\tabcolsep}{4pt}
\begin{tabular}{@{}>{\raggedright\arraybackslash}m{3.4cm}>{\raggedright\arraybackslash}m{3.9cm}>{\raggedright\arraybackslash}m{3.3cm}>{\raggedright\arraybackslash}m{1.65cm}>{\raggedright\arraybackslash}m{2.3cm}@{}}
\toprule
\textbf{Bangla idiom} & \textbf{Literal translation} & \textbf{English counterpart} & \textbf{Image} & \textbf{F class} \\
\midrule
\band{5}{Shared imagery}{the same image, or one element of it} \\
\bn{kata}\newline\bnrom{kata} & \bnlit{kata} & Rubbing salt in the wound & same image & beating \& punishment \\
\bn{ulubon}\newline\bnrom{ulubon} & \bnlit{ulubon} & Casting pearls before swine & shared object & money \& livelihood \\
\bn{dudh}\newline\bnrom{dudh} & \bnlit{dudh} & Nurse a viper in one's bosom & shared animal & beating \& punishment \\
\bn{ghora}\newline\bnrom{ghora} & \bnlit{ghora} & Going over someone's head & shared motion & deception \& secrecy \\
\addlinespace[2.5pt]
\band{5}{Different imagery}{one gist through unrelated images} \\
\bn{gache}\newline\bnrom{gache} & \bnlit{gache} & Counting one's chickens before they hatch & \multirow{10}{=}{different image} & \multirow{6}{=}{work \& tasks} \\
\bn{jotoGorje}\newline\bnrom{jotoGorje} & \bnlit{jotoGorje} & All bark and no bite &  &  \\
\bn{bamon}\newline\bnrom{bamon} & \bnlit{bamon} & Biting off more than one can chew &  &  \\
\bn{otiChalak}\newline\bnrom{otiChalak} & \bnlit{otiChalak} & Hoist with one's own petard &  & folly \& cunning \\
\bn{nake}\newline\bnrom{nake} & \bnlit{nake} & Sleep like a log &  & fear \& agitation \\
\addlinespace[2.5pt]
\band{5}{No imagery}{a literal English counterpart} \\
\bn{chorer}\newline\bnrom{chorer} & \bnlit{chorer} & Holier-than-thou attitude & no image & deception \& secrecy \\
\bottomrule
\end{tabular}
\end{table}

\para{One gist, many images.} \Cref{tab:app-qual-pairs} places ten Bangla idioms beside their English counterparts. The imagery ranges from identical, as with salt in a wound, through a shared object (pearls, a snake) or motion (leaping over a superior), to none at all, as with an oiled mustache against unhatched chickens or a thief's loud-mouthed mother against a holier-than-thou attitude. Semantic alignment could support these pairs despite differences in imagery. The aggregate geometry distinguishes idiom counterparts from explicit glosses: neutral-frame idiom counterparts share a median 2.0 of 11 class-mean dimensions with Bangla idiom spans, against 4.4 for explicit glosses (\cref{fig:gap}d), showing greater class-mean alignment for explicit glosses. The classes of these pairs also span the range of \cref{tab:classes}, from \emph{work \& tasks} (ratio 0.67) to \emph{deception \& secrecy} (0.23) and \emph{folly \& cunning} (0.00).

\begin{table}[!htbp]
\caption{\textbf{Six deterministic illustrations.} The lexicographically first two TEST idioms (by record identifier) that none, three and all six encoders classify correctly under the planned readout (English counterpart MLR directions, F, Bangla idioms in context). Gloss: a clause of the corpus's figurative gloss; counterpart: the first listed, which the experiments use. These cases illustrate the class-level pattern in \cref{tab:classes}.}
\label{tab:app-qual-cases}
\centering
\small
\setlength{\tabcolsep}{4pt}
\begin{tabular}{@{}>{\raggedright\arraybackslash}m{3.45cm}>{\raggedright\arraybackslash}m{2.6cm}>{\raggedright\arraybackslash}m{3.0cm}>{\raggedright\arraybackslash}m{2.75cm}>{\raggedright\arraybackslash}m{2.6cm}@{}}
\toprule
\textbf{Bangla idiom} & \textbf{Literal translation} & \textbf{Figurative gloss} & \textbf{English counterpart} & \textbf{F class; S} \\
\midrule
\band{5}{Correct for no encoder}{0/6} \\
\bn{goroj}\newline\bnrom{goroj} & \bnlit{goroj} & `the importance of necessity' & Desperate times call for desperate measures & money \& livelihood; neutral \\
\bn{dudh}\newline\bnrom{dudh} & \bnlit{dudh} & `to rear a deadly enemy with care' & Nurse a viper in one's bosom & beating \& punishment; negative \\
\addlinespace[2.5pt]
\band{5}{Correct for three of six encoders}{3/6} \\
\bn{konok}\newline\bnrom{konok} & \bnlit{konok} & `glorious full bloom' & In full bloom & residual: time, place, objects; positive \\
\bn{bhikkha}\newline\bnrom{bhikkha} & \bnlit{bhikkha} & `a person of illegitimate or low birth' & Base-born & people \& character; negative \\
\addlinespace[2.5pt]
\band{5}{Correct for all six encoders}{6/6} \\
\bn{tori}\newline\bnrom{tori} & \bnlit{tori} & `various kinds of raw vegetables' & Greens and veggies & residual: time, place, objects; neutral \\
\bn{paka}\newline\bnrom{paka} & \bnlit{paka} & `a shrewd and mature boy' & Street-smart kid & people \& character; neutral \\
\bottomrule
\end{tabular}
\end{table}

\para{Which idioms travel?} \Cref{tab:app-qual-cases} shows six TEST idioms chosen by a deterministic rule from the planned readout's outcomes. The two that every encoder classifies correctly are among the least figurative in the sample: \bn{tori} is a compound for vegetables, and \bn{paka} extends `ripe' to `precocious' much as English \textit{mature} does. Both fall in the classes with the highest English-direction recall. The two that no encoder classifies correctly carry evaluative or culturally specific images (a snake reared on milk and bananas, necessity cast as an affliction) and fall in classes with lower English-direction recall ratios. The middle cases, a golden lotus for glorious completion and a ritual term turned into a slur, divide the encoders evenly. These examples illustrate variation in expression type and class recall; the quantitative relation between evaluative judgments and reading is assessed in \cref{app:human}.

\begin{figure}[t]
\begin{caseboxfloat}{Why a masked context is a strong control}
\footnotesize
\begin{tabular}{@{}l>{\raggedright\arraybackslash}p{13.2cm}@{}}
\cond{bn-fig} & \bn{gacheSent} \\[2pt]
& `Thinking he will get the job, he is already thinking about a new car; he has oiled his mustache with the jackfruit still on the tree.' \\[3pt]
\cond{bn-context} & \bn{gacheSentA}\ \texttt{[MASK]}\ \bn{gacheCtxB} \\[2pt]
& `Thinking he will get the job, he is already thinking about a new car; he has applied \texttt{[MASK]}.' \\
\end{tabular}

\vspace{4pt}
With the idiom hidden, the sentence still states the premature anticipation that the idiom names, so a readout of the masked context can recover the figurative class from the context words alone. The corpus examples were written to illustrate their idioms, making masked context a strong baseline (\cref{sec:rq2}; \cref{fig:app-readability}b). This redundancy motivates the idiom-alone analysis and pooling-matched controls, which separate illustrative context from the idiom's contribution. The span condition pools the idiom's tokens and the masked condition pools the rest, so the two also differ in what is averaged.
\end{caseboxfloat}
\caption{\textbf{A worked example of the context control.} The \cond{bn-fig} and \cond{bn-context} inputs for the running example, in Bangla script with our English rendering. Bagdhara, CC BY-SA 4.0.}
\label{fig:app-qual-context}
\end{figure}

\para{Illustrative contexts reveal the class.} \Cref{fig:app-qual-context} shows why the masked control is strong for the running example. The sentence surrounding \bn{gache} already expresses premature anticipation, while the idiom adds a conventional evaluative framing. The pooling-matched analysis separates the effect of the idiom's presence from the choice of pooled tokens (\cref{app:pooling}).

\section{Data, Split and Inputs}
\label{app:data}

\para{Records.} The Bagdhara export contains 10,359 sense records with 8,877 distinct idiom strings, since some idioms have several senses. Of these records, 10,238 list at least one English counterpart, and we use the first.

\para{Exact spans.} In 8,822 records (85.2\%), the selected Bangla example sentence contains the exact idiom string; the remaining records are excluded from the span-based conditions \cond{bn-fig} and \cond{bn-context}. Only 0.3\% of English translations retain Bangla script.

\para{Split.} We link records that share a normalized idiom form, a listed spelling or form variant, or any identical model input (example sentence, translation, gloss or counterpart), and assign whole connected components to splits. The largest component holds 8.7\% of the records, and no identical input crosses splits.

\para{Pooling.} For span conditions we mean-pool the subword tokens that overlap the character span; for context conditions, all non-\texttt{[MASK]} tokens; for sentences and glosses, all tokens. mBERT marks 13.6\% of the tokens in \cond{en-idiom} inputs as unknown; all of them are the curly quotation marks of the frame, and none falls inside the counterpart span.

\begin{table}[!htbp]
\caption{\textbf{The Bangla inputs (\swatch{bntint}) of \cref{tab:conditions} in Bangla script}, as the models receive them, for the record of \bn{gache} (\bnrom{gache}). \cond{bn-sentence} uses the \cond{bn-fig} sentence but pools all its tokens; \cond{bn-context} replaces only the idiom with \texttt{[MASK]}. Romanizations in the text are reading aids, not experimental inputs. Bagdhara, CC BY-SA 4.0.}
\label{tab:bangla-inputs}
\centering
\small
\setlength{\tabcolsep}{4pt}
\begin{tabular}{@{}l>{\raggedright\arraybackslash}m{13.9cm}@{}}
\toprule
\textbf{Condition} & \textbf{Input} \\
\midrule
\tint{bntint}{l}\cond{bn-fig} & \bn{gacheSent} \\
\tint{bntint}{l}\cond{bn-context} & \bn{gacheSentA}\ \texttt{[MASK]}\ \bn{gacheCtxB} \\
\tint{bntint}{l}\cond{bn-idiom} & \bn{gacheFrame} \\
\tint{bntint}{l}\cond{bn-para} & \bn{gacheGlossFull} \\
\bottomrule
\end{tabular}
\end{table}

\para{Translations.} The \cond{bn-fig} sentence reads, in our rendering, `Thinking he will get the job, he is already thinking about a new car; he has oiled his mustache with the jackfruit still on the tree.' The neutral frame \cond{bn-idiom} reads `Here the expression ``\bn{gache}'' has been used.' The gloss \cond{bn-para} reads `Arranging enjoyment before attainment; imagining enjoyment before obtaining a result; planning as if something unattained had already been obtained.'

\FloatBarrier
\section{The Figurative Taxonomy}
\label{app:taxonomy}

\begin{figure}[!htbp]
\centering
\begin{subfigure}[t]{0.7264\textwidth}
\centering
\includegraphics[width=\linewidth,viewport=0.84 0.84 377.28 215.16,clip]{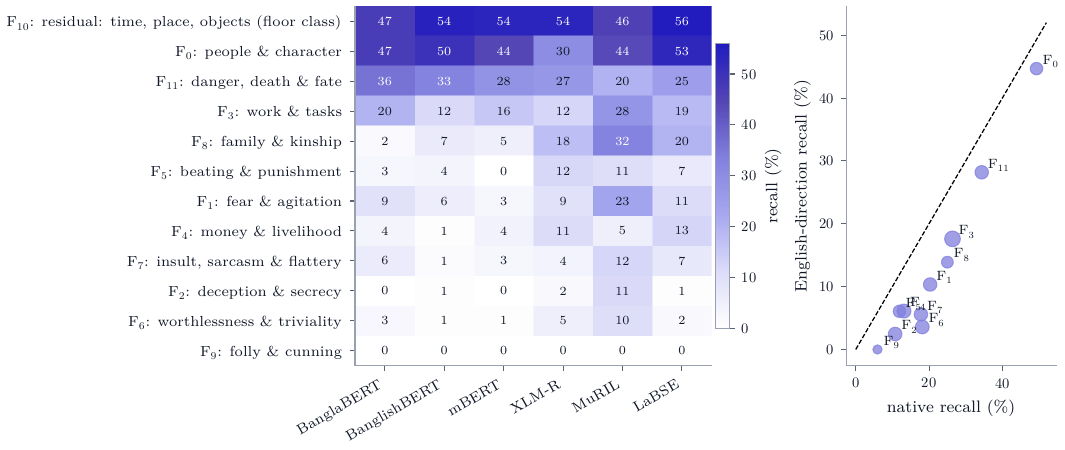}
\caption{English-direction recall per F class (\%)}
\end{subfigure}\hfill
\begin{subfigure}[t]{0.2536\textwidth}
\centering
\includegraphics[width=\linewidth,viewport=379.01 0.84 510.42 215.16,clip]{figures/app_classes.pdf}
\caption{English \textit{vs.} native recall}
\end{subfigure}
\caption{\textbf{Recall by figurative class.} (a)~Recall (\%) of each F class on Bangla idioms in context under the planned readout through English directions, per encoder. (b)~Each class's English-direction recall against its native recall, averaged over encoders (bubble size: TEST $n$; dashed: equality). Classes about people, danger and work lie closest to the diagonal; insult, deception, worthlessness and folly keep a third of their native recall or less. Discovery split; exploratory.}
\label{fig:app-classes}
\end{figure}

\Cref{tab:app-taxonomy} lists the twelve F classes with their sizes and keywords. We rebuilt the taxonomy from the released code and seeds, and its class counts and re-seed ARIs (0.399, 0.432, 0.432) match the campaign exactly. \Cref{fig:app-classes} shows English-direction recall per class and encoder. F agrees only weakly with the corpus's metadata: adjusted mutual information (AMI) is 0.069 with sentiment and 0.080 with usage domain. F is therefore a coarse topical partition, not a validated sense inventory, and the re-seed ARIs show substantial variation across clustering seeds. These ARIs come from the campaign's stability refits, with five initializations; when the pipeline rebuilds the taxonomy with its ten, seeds 1--3 give ARIs of \NyReseedAri{} (\cref{app:taxonomies}).

\para{Which meanings travel.} In the \textit{post hoc} class-level analysis, classes about people, danger, work and family keep \HvTravelTop\% of their native recall through English directions, whereas those named insult, deception, worthlessness and folly keep \mbox{0--\HvTravelBottom\%} (\cref{tab:classes,tab:travel}; \cref{fig:app-classes}). Evaluative function does not consistently distinguish these groups: the volunteers judged insult and folly mostly evaluative, but also people \& character, the non-residual class with the highest cross/native recall ratio (\cref{tab:travel}; \cref{app:human}). The ratios track how well Bangla's own readout separates each class ($\rho=\HvRhoRatioNative$).

\begin{table}[!htbp]
\caption{\textbf{Which meanings travel.} Recall (\%) through Bangla's own (\BN) or English (\EN) directions under the planned MLR readout, averaged over encoders, and their ratio. Good and Eval. give the percentages of the 25 annotated items per class judged good for their assigned class or evaluative, averaged over the two volunteers; Eval. includes \emph{both}. Recall uses the 2,166 exact-span TEST records; \fcl{10} is omitted.}

\label{tab:travel}
\centering
\small
\setlength{\tabcolsep}{10pt}
\begin{tabular}{@{}l rrc rr@{}}
\toprule
\textbf{F class} & \textbf{\BN} & \textbf{\EN} & \textbf{Ratio} & \textbf{Good} & \textbf{Eval.} \\
\midrule
people \& character & 49.3 & 44.7 & .91 & 94 & 72 \\
danger, death \& fate & 34.4 & 28.2 & .82 & 36 & 12 \\
work \& tasks & 26.4 & 17.6 & .67 & 34 & 10 \\
family \& kinship & 25.0 & 13.9 & .56 & 74 & 38 \\
beating \& punishment & 11.9 & 6.1 & .51 & 24 & 44 \\
fear \& agitation & 20.3 & 10.3 & .51 & 28 & 16 \\
money \& livelihood & 13.1 & 6.1 & .46 & 64 & 28 \\
insult, sarcasm \& flattery & 17.8 & 5.6 & .31 & 38 & 82 \\
deception \& secrecy & 10.7 & 2.5 & .23 & 32 & 40 \\
worthlessness \& triviality & 18.2 & 3.6 & .20 & 22 & 36 \\
folly \& cunning & 5.9 & 0.0 & .00 & 62 & 96 \\
\bottomrule
\end{tabular}
\end{table}

\para{Sentiment.} S is the export's \texttt{sentiment} field, used as distributed: negative 4,952 records (47.8\%), neutral 3,531 (34.1\%), positive 1,724 (16.6\%) and mixed 152 (1.5\%). The corpus paper reports 47.1\%, 37.4\% and 15.5\% over three classes for its release of 10,361 records (its \S4.2), so our export, dated 2026-03-01, differs in its labels as well as its size. We apply no recoding, and \emph{mixed} is a fourth class (hence $M{=}3$). The labels are expert-validated rather than purely human: Gemini-2.5-flash and GPT-4o drafted sentiment suggestions (and English translations), which at least two experts reviewed, refined or replaced (its \S3.2).

\begin{table}[t]
\caption{The figurative-class taxonomy F: $k$-means ($k{=}12$) over MiniLM embeddings of English figurative glosses, fitted on SPACE.
SPACE and TEST count all records of each split (4,143 and 2,590); the planned tests use the 2,166 TEST records
with an exact Bangla idiom span (\cref{tab:classes}). Names were assigned by the authors from the keywords after the
experiments. Keywords are ranked by term frequency--inverse document frequency (TF-IDF). Neg.: share of members with negative sentiment.}
\label{tab:app-taxonomy}
\centering
\small
\setlength{\tabcolsep}{4pt}
\begin{tabular}{@{}llrrrm{7.1cm}@{}}
\toprule
& \textbf{Name (\textit{post hoc})} & \textbf{SPACE} & \textbf{TEST} & \textbf{Neg.\,\%} & \textbf{SPACE keywords (top TF-IDF terms of member glosses)} \\
\midrule
\fcl{0} & people \& character & 358 & 148 & 58 & \footnotesize person, extremely, woman, wealthy, man, individual, incompetent, old, worthless, ordinary \\
\fcl{1} & fear \& agitation & 357 & 217 & 58 & \footnotesize restlessness, fear, feel, extremely, intense, anxiety, sound, restless, extreme, enthusiasm \\
\fcl{2} & deception \& secrecy & 295 & 217 & 68 & \footnotesize deceive, deception, cheat, secretly, trick, false, fault, truth, mistake \\
\fcl{3} & work \& tasks & 455 & 401 & 22 & \footnotesize task, work, attention, make, tasks, start, accomplish, set, proceed \\
\fcl{4} & money \& livelihood & 278 & 226 & 49 & \footnotesize money, livelihood, wealth, food, sustenance, means, prosperity, destitute, spending \\
\fcl{5} & beating \& punishment & 354 & 223 & 73 & \footnotesize beat, severely, thrash, beating, punishment, anger, severe, oneself, subdue \\
\fcl{6} & worthlessness \& triviality & 302 & 200 & 37 & \footnotesize insignificant, worthless, trivial, small, good, matter, things, substance, extremely, useless \\
\fcl{7} & insult, sarcasm \& flattery & 397 & 219 & 75 & \footnotesize sarcastically, person, words, arrogance, insult, flattery, manner, sarcastic, ugly \\
\fcl{8} & family \& kinship & 235 & 127 & 28 & \footnotesize family, child, beloved, woman, wife, children, lineage, young, household \\
\fcl{9} & folly \& cunning & 191 & 46 & 76 & \footnotesize foolish, cunning, fool, intelligence, person, lacking, utter, idiot, sarcastically, clever \\
\fcl{10} & residual: time, place, objects & 528 & 360 & 13 & \footnotesize day, long, like, fresh, various, time, place, beautiful, food \\
\fcl{11} & danger, death \& fate & 393 & 206 & 61 & \footnotesize danger, death, end, life, away, enemy, time, destruction, fate \\
\bottomrule
\end{tabular}
\end{table}
\begin{table}[t]
\caption{\textbf{Which meanings travel.} Recall (\%) of each figurative class on Bangla TEST idioms in context,
mean\,{\scriptsize$\pm$\,sd} over the six encoders, under the planned readout through native Bangla or English-counterpart
directions; Ratio is English over native mean recall. Seven classes keep about half of their recall or more
(\swatch{keytint}); insult, deception, worthlessness and folly keep a third or less (\swatch{negtint}).
\Cref{tab:travel} sets the means beside the volunteers' judgments. $n$ counts the
planned population, the 2,166 TEST records with an exact Bangla idiom span; the residual class
\fcl{10} ($n=339$) is omitted. Names are \textit{post hoc} readings of a machine taxonomy.
Exploratory.}
\label{tab:classes}
\centering
\small
\setlength{\tabcolsep}{2.5pt}
\begin{tabular}{@{}lrccc@{}}
\toprule
\textbf{F class (\textit{post hoc} name)} & \textbf{$n$} & \textbf{Native} & \textbf{English} & \textbf{Ratio} \\
\midrule
people \& character & 144 & \mv{49.3}\pmsd{4.2} & \mv{44.7}\pmsd{8.0} & \hlkeyr .91 \\
danger, death \& fate & 178 & \mv{34.4}\pmsd{8.9} & \mv{28.2}\pmsd{5.5} & \hlkeyr .82 \\
work \& tasks & 269 & \mv{26.4}\pmsd{7.3} & \mv{17.6}\pmsd{5.9} & \hlkeyr .67 \\
family \& kinship & 120 & \mv{25.0}\pmsd{11.5} & \mv{13.9}\pmsd{11.6} & \hlkeyr .56 \\
beating \& punishment & 140 & \mv{11.9}\pmsd{5.5} & \mv{6.1}\pmsd{4.8} & \hlkeyr .51 \\
fear \& agitation & 179 & \mv{20.3}\pmsd{8.0} & \mv{10.3}\pmsd{7.1} & \hlkeyr .51 \\
money \& livelihood & 199 & \mv{13.1}\pmsd{5.3} & \mv{6.1}\pmsd{4.7} & \hlkeyr .46 \\
insult, sarcasm \& flattery & 179 & \mv{17.8}\pmsd{5.1} & \mv{5.6}\pmsd{4.0} & \hlnegr .31 \\
deception \& secrecy & 183 & \mv{10.7}\pmsd{5.1} & \mv{2.5}\pmsd{4.5} & \hlnegr .23 \\
worthlessness \& triviality & 191 & \mv{18.2}\pmsd{7.7} & \mv{3.6}\pmsd{3.8} & \hlnegr .20 \\
folly \& cunning & 45 & \mv{5.9}\pmsd{4.1} & \mv{0.0}\pmsd{0.0} & \hlnegr .00 \\
\bottomrule
\end{tabular}
\end{table}

\section{Human Validation}
\label{app:human}

\para{Sample and questions.} The packet holds 300 TEST records, 25 from each F class and one per held-out component, so it covers the classes evenly rather than estimating corpus prevalence. No model output is shown. For each record, volunteers saw the Bangla idiom in its example sentence with an English rendering, its Bangla and English figurative glosses, its English counterpart and its assigned F class, with a codebook giving each class's name, keywords and three SPACE glosses. They answered four questions, each with an explicit \emph{uncertain}: whether the class describes the idiom's main meaning (good, partial, poor), whether the idiom mainly describes (referential) or judges (evaluative; or both), what kind of expression the counterpart is (idiom, proverb, literal paraphrase, other), and whether it preserves the meaning (equivalent, partly, not).

\para{Procedure.} Two computer-science undergraduates, native Bangla speakers fluent in English and not authors, volunteered. They worked independently from brief written instructions and the codebook and entered their answers in spreadsheets. No session times or click records were kept, and disagreements were not adjudicated. The analysis follows the plan written before annotation: unweighted Cohen's $\kappa$ per question with a 2,000-draw item bootstrap, \emph{uncertain} kept as a category. Gwet's AC1 and the class- and item-level associations below are exploratory. Both volunteers' answer sheets are in the repository.

\para{Agreement.} Agreement is substantial for the idiom's function ($\kappa=\HvKappaFunction$, \HvAgreeFunction\%) and the counterpart's kind (\HvKappaType, \HvAgreeType\%), moderate for the class's aptness (\HvKappaClass, \HvAgreeClass\%) and fair for the counterpart's meaning (\HvKappaEquiv, \HvAgreeEquiv\%; \cref{tab:app-human-agreement}). Most disagreements on aptness are between adjacent answers, and most items are judged equivalent, which depresses $\kappa$ for meaning (AC1 \HvAcOneEquiv).

\para{Annotation outcomes.} Both volunteers judged the assigned class apt (good or partial) for \HvAptBoth\% of sampled items and misleading for \HvPoorBoth\%, with substantial differences between classes (\cref{tab:app-human-classes}). Both judged \HvIdiomBoth\% of counterparts to be conventional idioms and, separately, \HvEquivBoth\% to preserve the Bangla meaning. They jointly classified \HvParaphraseBoth\% as literal paraphrases and \HvProverbBoth\% as proverbs. Each judged \HvAtLeastPartly\% at least partly equivalent, and both judged \HvNotEquivBoth\% not equivalent. These marginal judgments characterize distinct properties of the counterparts; their mixture of expression types motivates the input and counterpart-type analyses in \cref{sec:rq2,app:counterpart-type}. Both judged \HvReferentialBoth\% of the Bangla idioms referential and \HvEvaluativeBoth\% evaluative.

\para{Class-level associations.} Classes named for insult and folly are judged mostly evaluative (\HvEvalInsult\% and \HvEvalFolly\%). Across the 12 classes, evaluative share correlates negatively with the cross/native recall ratio ($\rho=\HvRhoEvalTransfer$, $p=\HvPEvalTransfer$; without the floor class, $\rho=\HvRhoEvalTransferCore$). People \& character is also mostly evaluative (\HvEvalPeople\%) but has the highest non-residual ratio (\HvEvalPeopleTransfer). Evaluative share correlates negatively with native recall ($\rho=\HvRhoEvalNative$), while cross/native recall ratios correlate positively with native recall ($\rho=\HvRhoRatioNative$). After controlling for native recall, the evaluative association is near zero (partial $\rho=\HvPartialEval$). The item-level analysis below examines these judgments within classes.

\para{Item-level analysis.} The judged discovery TEST items are matched to their readouts through the packet's hashed identifiers. For each judgment, a linear probability model with F-class and encoder fixed effects estimates the association with correct F classification, pooled over the six encoders. Models for English-direction and zero-shot readouts include native correctness as a covariate (\cref{tab:app-human-items}). Judgments are permuted across items within F class (4,999 draws), intervals resample items within class (2,000 draws), and Holm correction covers the five judgments.

Within classes, counterpart type predicts English-direction reading after accounting for native correctness. Where both volunteers judged the counterpart a conventional idiom, English COV directions classify the item correctly \NzIdiomCpIdiomAbs{} points less often on the idiom alone (\NzIdiomCpIdiomCI, Holm $p=\NzIdiomCpIdiomHolm$) and \NzIdiomCpFigAbs{} points less often in context (\NzIdiomCpFigCI, Holm $p=\NzIdiomCpFigHolm$). Native-readout coefficients for these items are also negative (\NzIdiomCpNativeIdiom{} and \NzIdiomCpNativeFig{} points; Holm $p=\NzIdiomCpNativeIdiomHolm$ and $p=\NzIdiomCpNativeFigHolm$). The zero-shot and MLR coefficients on idioms alone have the same sign (\NzIdiomCpZs{} and \NzIdiomCpMlr), but are not significant after Holm correction.

The evaluative-share coefficients are not distinguishable from zero, either on the idiom alone (\NzEvalIdiom{} points from neither to both volunteers judging it evaluative, \NzEvalIdiomCI, $p=\NzEvalIdiomP$) or in context (\NzEvalFig, \NzEvalFigCI, $p=\NzEvalFigP$). Class aptness predicts native correctness in context (\NzAptNativeFig{} points, \NzAptNativeFigCI); its coefficient for English-direction reading after accounting for native correctness is \NzAptCrossFig{} points and is not distinguishable from zero.

The counterpart-transparency covariate---cosine similarity between the counterpart and English gloss---agrees with the volunteers' expression-type judgments (AUC \NzTranspAuc{} for idioms against literal paraphrases). In its lowest tertile, \NzOpaqueIdiomShare\% of judged counterparts were called idioms by both volunteers; directions estimated from that tertile support weaker reading (\cref{app:counterpart-type}). These item-level analyses are exploratory, and the judgments were not adjudicated.

\begin{table}[t]
\caption{\textbf{Two volunteers' judgments of 300 TEST items} (25 per F class): answers per volunteer and items where
both gave the answer. Beside each question, unweighted Cohen's $\kappa$ (95\% item-bootstrap interval, 2,000 draws)
and raw agreement. \emph{Uncertain} is an answer of its own; three items lack an English counterpart.}
\label{tab:app-human-agreement}
\centering
\small
\setlength{\tabcolsep}{9pt}
\begin{tabular}{@{}l ccc@{}}
\toprule
\textbf{Answer} & \textbf{A} & \textbf{B} & \textbf{Both} \\
\midrule
\band{4}{Class aptness}{$\kappa$ 0.51 [0.43, 0.59], 68\%} \\
good & 154 & 123 & 107 \\
partial & 55 & 90 & 31 \\
poor & 91 & 87 & 67 \\
\addlinespace[2.5pt]
\band{4}{Function}{$\kappa$ 0.63 [0.55, 0.71], 81\%} \\
referential & 157 & 200 & 155 \\
evaluative & 140 & 92 & 89 \\
both & 3 & 6 & 0 \\
uncertain & 0 & 2 & 0 \\
\addlinespace[2.5pt]
\band{4}{Counterpart kind}{$\kappa$ 0.74 [0.66, 0.81], 88\%} \\
idiom & 213 & 203 & 194 \\
proverb & 12 & 10 & 10 \\
literal paraphrase & 69 & 79 & 59 \\
other & 6 & 5 & 1 \\
uncertain & 0 & 3 & 0 \\
\addlinespace[2.5pt]
\band{4}{Counterpart meaning}{$\kappa$ 0.40 [0.30, 0.50], 75\%} \\
equivalent & 251 & 192 & 188 \\
partly equivalent & 41 & 92 & 30 \\
not equivalent & 5 & 13 & 3 \\
uncertain & 3 & 3 & 3 \\
\bottomrule
\end{tabular}
\end{table}
\begin{table}[t]
\caption{\textbf{Human judgments by F class}, 25 items each, averaged over the two volunteers (\emph{evaluative}
includes \emph{both}), beside the class's cross/native recall ratio: mean English-direction recall divided by mean native
recall across encoders (\cref{fig:app-classes}; residual F\textsubscript{10} is the floor class). This ratio
orders the rows. Spearman correlations of the ratio with the evaluative share ($\rho=\HvRhoEvalTransfer$,
$p=\HvPEvalTransfer$) and the share judged good ($\rho=\HvRhoGoodTransfer$, $p=\HvPGoodTransfer$) over the 12 classes
are exploratory; the first is near zero after controlling for native recall (partial $\rho=\HvPartialEval$).}
\label{tab:app-human-classes}
\centering
\small
\setlength{\tabcolsep}{4.4pt}
\begin{tabular}{@{}l ccc cc c@{}}
\toprule
\textbf{\hdtwo{F class}} & \multicolumn{3}{c}{\textbf{The idiom (\%)}} & \multicolumn{2}{c}{\textbf{English counterpart (\%)}} & \textbf{\hdtwo{Ratio}} \\
\cmidrule(lr){2-4}\cmidrule(lr){5-6}
 & \textbf{class good} & \textbf{class poor} & \textbf{evaluative} & \textbf{an idiom} & \textbf{equivalent} & \\
\midrule
F\textsubscript{10} residual: time, place, objects & 46 & 28 & 8 & 54 & 72 & 1.02 \\
F\textsubscript{0} people \& character & 94 & 2 & 72 & 68 & 68 & .91 \\
F\textsubscript{11} danger, death \& fate & 36 & 30 & 12 & 82 & 84 & .82 \\
F\textsubscript{3} work \& tasks & 34 & 44 & 10 & 74 & 82 & .67 \\
F\textsubscript{8} family \& kinship & 74 & 10 & 38 & 60 & 58 & .56 \\
F\textsubscript{5} beating \& punishment & 24 & 28 & 44 & 80 & 78 & .51 \\
F\textsubscript{1} fear \& agitation & 28 & 34 & 16 & 84 & 62 & .51 \\
F\textsubscript{4} money \& livelihood & 64 & 26 & 28 & 70 & 86 & .46 \\
F\textsubscript{7} insult, sarcasm \& flattery & 38 & 36 & 82 & 62 & 80 & .31 \\
F\textsubscript{2} deception \& secrecy & 32 & 44 & 40 & 62 & 60 & .23 \\
F\textsubscript{6} worthlessness \& triviality & 22 & 64 & 36 & 66 & 86 & .20 \\
F\textsubscript{9} folly \& cunning & 62 & 10 & 96 & 70 & 70 & .00 \\
\bottomrule
\end{tabular}
\end{table}
\begin{table}[t]
\caption{\textbf{The volunteers' judgments, item by item.} Change (points; 95\% item-bootstrap interval) in the probability that
a readout classifies an item's F class correctly, from a linear model with F-class and encoder fixed effects over the six
encoders; the English and zero-shot models add the native readout's correctness as a covariate. A judgment holds when
both volunteers gave it (class apt: good or partial); the \emph{evaluative share} of the two runs from 0 to 1.
\swatch{keytint}/\swatch{negtint}\,Holm-adjusted $p\le.05$ over the five judgments, permuting each across items within F
class. Exploratory.}
\label{tab:app-human-items}
\centering
\small
\setlength{\tabcolsep}{4pt}
\begin{tabular}{@{}l ccc cc@{}}
\toprule
 & \multicolumn{3}{c}{\textbf{The idiom alone (300 items)}} & \multicolumn{2}{c}{\textbf{In context (257 items)}} \\
\cmidrule(lr){2-4}\cmidrule(l){5-6}
\textbf{Judgment} & \textbf{English COV} & \textbf{Zero-shot} & \textbf{Native} & \textbf{English COV} & \textbf{Native} \\
\midrule
Class apt & +1.7\,{\scriptsize[\ensuremath{-}2.0, +5.5]} & +2.9\,{\scriptsize[\ensuremath{-}1.4, +6.8]} & +5.4\,{\scriptsize[\ensuremath{-}0.6, +11.2]} & +1.9\,{\scriptsize[\ensuremath{-}2.3, +5.8]} & \hlkeyr +15.6\,{\scriptsize[+8.8, +22.3]} \\
Class good & +0.8\,{\scriptsize[\ensuremath{-}3.0, +4.7]} & +3.6\,{\scriptsize[\ensuremath{-}0.7, +7.9]} & +5.9\,{\scriptsize[\ensuremath{-}1.1, +12.9]} & +0.6\,{\scriptsize[\ensuremath{-}3.9, +5.3]} & \hlkeyr +12.4\,{\scriptsize[+4.6, +20.4]} \\
Evaluative share & \ensuremath{-}3.4\,{\scriptsize[\ensuremath{-}7.4, +0.7]} & \ensuremath{-}2.7\,{\scriptsize[\ensuremath{-}7.4, +1.9]} & +0.3\,{\scriptsize[\ensuremath{-}5.6, +6.5]} & +0.5\,{\scriptsize[\ensuremath{-}4.6, +5.9]} & \ensuremath{-}6.1\,{\scriptsize[\ensuremath{-}14.5, +2.7]} \\
Counterpart an idiom & \hlneg \ensuremath{-}7.3\,{\scriptsize[\ensuremath{-}10.6, \ensuremath{-}3.7]} & \ensuremath{-}2.7\,{\scriptsize[\ensuremath{-}6.7, +1.2]} & \ensuremath{-}6.7\,{\scriptsize[\ensuremath{-}12.4, \ensuremath{-}1.0]} & \hlneg \ensuremath{-}6.9\,{\scriptsize[\ensuremath{-}11.1, \ensuremath{-}2.7]} & \ensuremath{-}7.9\,{\scriptsize[\ensuremath{-}14.6, \ensuremath{-}1.3]} \\
Counterpart equivalent & +1.6\,{\scriptsize[\ensuremath{-}1.6, +4.8]} & +1.7\,{\scriptsize[\ensuremath{-}1.9, +5.3]} & +6.1\,{\scriptsize[+0.4, +11.6]} & +0.4\,{\scriptsize[\ensuremath{-}3.9, +4.7]} & +3.1\,{\scriptsize[\ensuremath{-}3.7, +9.9]} \\
\bottomrule
\end{tabular}
\end{table}

\section{Additional Results}
\label{app:results}

\begin{figure}[t]
\centering
{\centering\includegraphics[width=0.4333\textwidth,viewport=141.38 236.58 369.82 252.00,clip]{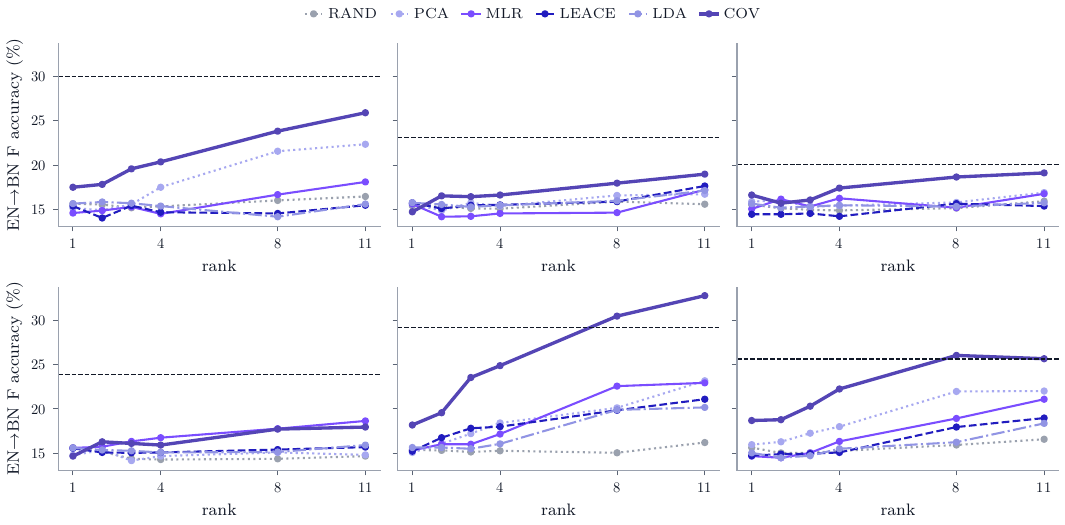}\par}
\par\medskip
\begin{subfigure}[t]{0.3484\textwidth}
\centering
\includegraphics[width=\linewidth,viewport=0.84 117.91 184.56 235.98,clip]{figures/app_rank_curves.pdf}
\caption{BanglaBERT}
\end{subfigure}\hfill
\begin{subfigure}[t]{0.3058\textwidth}
\centering
\includegraphics[width=\linewidth,viewport=186.24 117.91 347.46 235.98,clip]{figures/app_rank_curves.pdf}
\caption{BanglishBERT}
\end{subfigure}\hfill
\begin{subfigure}[t]{0.3058\textwidth}
\centering
\includegraphics[width=\linewidth,viewport=349.14 117.91 510.36 235.98,clip]{figures/app_rank_curves.pdf}
\caption{mBERT}
\end{subfigure}
\par\medskip
\begin{subfigure}[t]{0.3484\textwidth}
\centering
\includegraphics[width=\linewidth,viewport=0.84 0.84 184.56 118.91,clip]{figures/app_rank_curves.pdf}
\caption{XLM-R}
\end{subfigure}\hfill
\begin{subfigure}[t]{0.3058\textwidth}
\centering
\includegraphics[width=\linewidth,viewport=186.24 0.84 347.46 118.91,clip]{figures/app_rank_curves.pdf}
\caption{MuRIL}
\end{subfigure}\hfill
\begin{subfigure}[t]{0.3058\textwidth}
\centering
\includegraphics[width=\linewidth,viewport=349.14 0.84 510.36 118.91,clip]{figures/app_rank_curves.pdf}
\caption{LaBSE}
\end{subfigure}
\caption{\textbf{English-direction F accuracy by estimator and rank} on Bangla idioms in context, one panel per encoder (dashed: native full-space probe). The pattern estimator COV (thick) dominates at almost every rank in five of six encoders; line styles distinguish filters (MLR solid, LEACE dashed, LDA dash-dotted) from unsupervised baselines (dotted). Discovery split; exploratory.}
\label{fig:app-rank}
\end{figure}

\begin{figure}[t]
\centering
\begin{subfigure}[t]{0.4909\textwidth}
\centering
\includegraphics[width=\linewidth,viewport=0.84 0.84 255.21 186.36,clip]{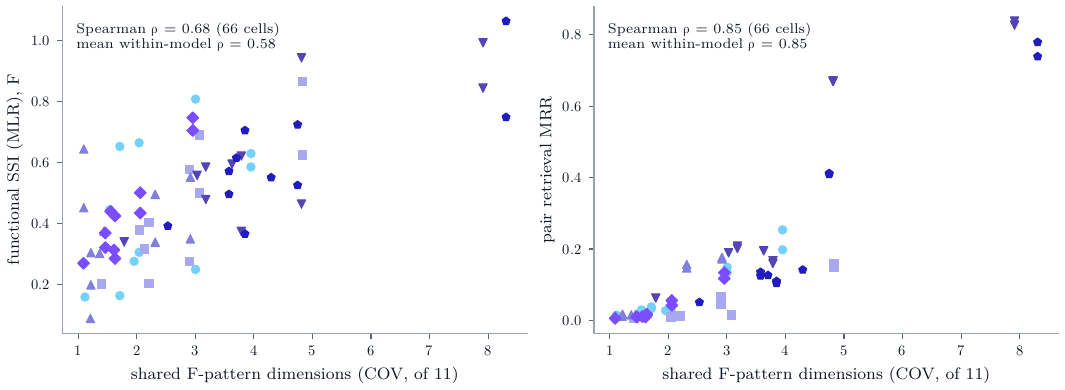}
\caption{functional SSI (MLR)}
\end{subfigure}\hfill
\begin{subfigure}[t]{0.4891\textwidth}
\centering
\includegraphics[width=\linewidth,viewport=256.89 0.84 510.36 186.36,clip]{figures/app_geometry.pdf}
\caption{retrieval mean reciprocal rank (MRR)}
\end{subfigure}
\caption{\textbf{Geometry tracks function.} Across 66 model--pair cells, the dimensions shared by English and Bangla class-mean subspaces track SSI (a) and paired-item retrieval (b). Cells are not independent (reverse pairs share their geometry), so the correlations are descriptive. Encoders: \encoderkey.}
\label{fig:app-geometry}
\end{figure}

\begin{figure}[t]
\centering
\begin{subfigure}[t]{0.3330\textwidth}
\centering
\includegraphics[width=\linewidth,viewport=0.84 0.84 176.39 164.76,clip]{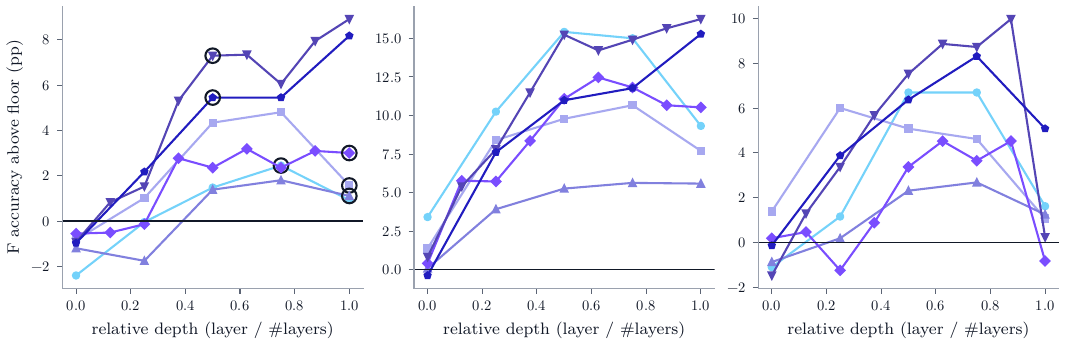}
\caption{EN MLR $\rightarrow$ BN}
\end{subfigure}\hfill
\begin{subfigure}[t]{0.3168\textwidth}
\centering
\includegraphics[width=\linewidth,viewport=178.07 0.84 345.13 164.76,clip]{figures/app_layers.pdf}
\caption{BN MLR $\rightarrow$ BN (native)}
\end{subfigure}\hfill
\begin{subfigure}[t]{0.3102\textwidth}
\centering
\includegraphics[width=\linewidth,viewport=346.81 0.84 510.36 164.76,clip]{figures/app_layers.pdf}
\caption{EN PCA $\rightarrow$ BN}
\end{subfigure}
\caption{\textbf{Sharing by depth.} F accuracy above floor (rank 11) by relative depth: English MLR directions on Bangla (a; circles mark validation-selected layers), native Bangla MLR (b), and English PCA directions (c). Encoders: \encoderkey. \textit{Post hoc} TEST profiles; no layer is reselected on them.}
\label{fig:app-layers}
\end{figure}

\begin{figure}[t]
\centering
{\centering\includegraphics[width=0.9585\textwidth,viewport=7.26 168.47 503.94 194.40,clip]{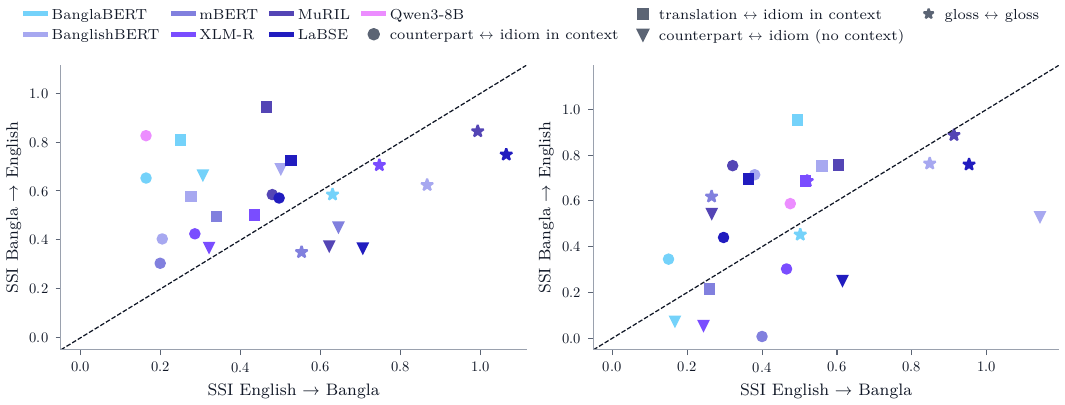}\par}
\par\medskip
\begin{subfigure}[t]{0.4900\textwidth}
\centering
\includegraphics[width=\linewidth,viewport=0.84 0.84 254.76 165.19,clip]{figures/app_asymmetry.pdf}
\caption{F}
\end{subfigure}\hfill
\begin{subfigure}[t]{0.4900\textwidth}
\centering
\includegraphics[width=\linewidth,viewport=256.44 0.84 510.36 165.19,clip]{figures/app_asymmetry.pdf}
\caption{S}
\end{subfigure}
\caption{\textbf{Direction of cross-lingual reading.} SSI from English to Bangla against SSI from Bangla to English, for F (a) and S (b); color marks the encoder and marker shape the pair type (key above the panels). The denominator, floor and item cohort change with the target, so directional differences do not measure a language's private information.}
\label{fig:app-asym}
\end{figure}

\begin{figure}[t]
\centering
\begin{subfigure}[t]{0.3787\textwidth}
\centering
\includegraphics[width=\linewidth,viewport=0.81 0.84 197.21 193.56,clip]{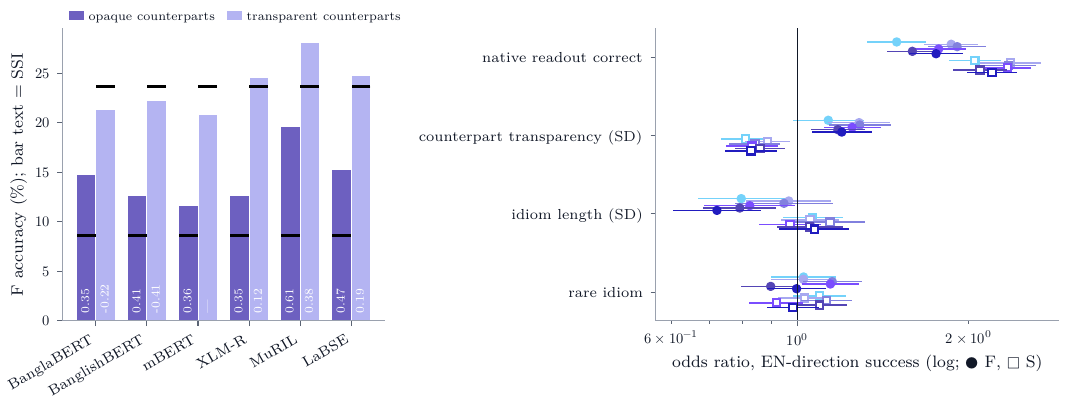}
\caption{Transparency: accuracy $\uparrow$, SSI $\downarrow$ (---: floor)}
\end{subfigure}\hfill
\begin{subfigure}[t]{0.6013\textwidth}
\centering
\includegraphics[width=\linewidth,viewport=198.56 0.84 510.36 193.56,clip]{figures/app_floors.pdf}
\caption{Item-level predictors (\textit{post hoc} GLM)}
\end{subfigure}
\caption{\textbf{Floors and item-level associations.} (a)~Transparent and opaque idioms: transparent items are read more accurately (bars) but have a lower SSI (white numbers), because their majority floor (black ticks) is higher. (b)~Odds ratios per standard deviation from \textit{post hoc} complete-case regressions of English-direction correctness ($\bullet$ F, $\square$ S). Encoders: \encodercolors.}
\label{fig:app-floors}
\end{figure}

\begin{figure}[t]
\centering
{\centering\includegraphics[width=0.4601\textwidth,viewport=136.40 150.18 374.80 165.60,clip]{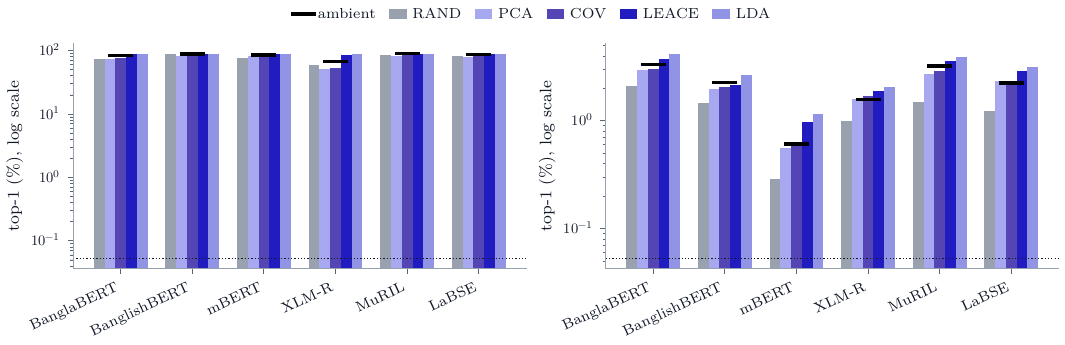}\par}
\par\medskip
\begin{subfigure}[t]{0.4900\textwidth}
\centering
\includegraphics[width=\linewidth,viewport=0.84 0.84 254.76 146.90,clip]{figures/app_identity.pdf}
\caption{Idiom spans (shared surface form)}
\end{subfigure}\hfill
\begin{subfigure}[t]{0.4900\textwidth}
\centering
\includegraphics[width=\linewidth,viewport=256.44 0.84 510.36 146.90,clip]{figures/app_identity.pdf}
\caption{Idiom-masked contexts}
\end{subfigure}
\caption{\textbf{Within-Bangla identity retrieval} of unseen idioms (top-1, log scale; dotted: chance) for idiom spans (a) and idiom-masked contexts (b). Visible spans admit a form-matching shortcut (\cref{app:identity-audit}).}
\label{fig:app-identity}
\end{figure}

\begin{figure}[t]
\centering
\begin{subfigure}[t]{0.5667\textwidth}
\centering
\includegraphics[width=\linewidth,viewport=0.84 0.84 294.53 186.36,clip]{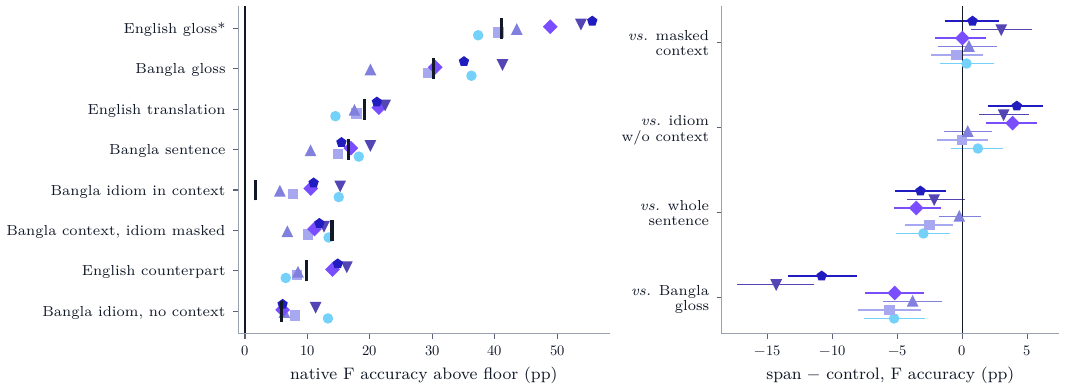}
\caption{Native F readability (| = TF-IDF)}
\end{subfigure}\hfill
\begin{subfigure}[t]{0.4133\textwidth}
\centering
\includegraphics[width=\linewidth,viewport=296.21 0.84 510.36 186.36,clip]{figures/app_readability.pdf}
\caption{Matched controls (95\% component CI)}
\end{subfigure}
\caption{\textbf{Readability by input, and the matched controls.} (a)~Native F readability by input condition (MLR rank 11, pp above floor; black bars: TF-IDF on the same text). (b)~English-direction F accuracy on Bangla idiom spans minus accuracy on each control representation of the same 2,166 records, with unadjusted 95\% paired component-bootstrap intervals; positive values favor the span. Controls differ in input and pooled tokens, and intervals spanning zero do not establish equivalence. Encoders: \encoderkey. Exploratory.}
\label{fig:app-readability}
\end{figure}

\begin{figure}[t]
\centering
{\centering\includegraphics[width=0.7074\textwidth,viewport=72.30 153.05 438.90 169.20,clip]{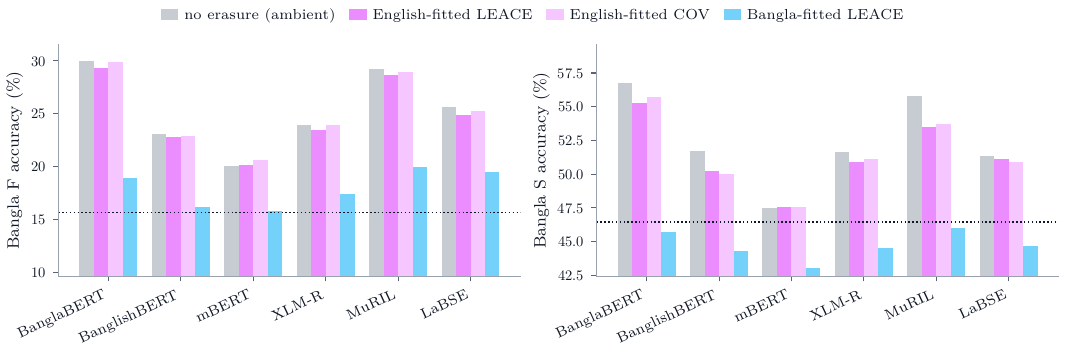}\par}
\par\medskip
\begin{subfigure}[t]{0.4847\textwidth}
\centering
\includegraphics[width=\linewidth,viewport=0.84 0.84 251.99 149.77,clip]{figures/app_erasure_FS.pdf}
\caption{F, rank-11 erasure (dotted: floor)}
\end{subfigure}\hfill
\begin{subfigure}[t]{0.4953\textwidth}
\centering
\includegraphics[width=\linewidth,viewport=253.67 0.84 510.36 149.77,clip]{figures/app_erasure_FS.pdf}
\caption{S, rank-3 erasure (dotted: floor)}
\end{subfigure}
\caption{\textbf{Erasure for F (rank 11) and S (rank 3).} Bangla accuracy after removal and probe retraining, for the ambient representation, English-fitted LEACE and COV, and Bangla-fitted LEACE (dotted: majority floor). Bangla-fitted LEACE pushes S below its floor, a known finite-sample effect of erasure.}
\label{fig:app-erasure}
\end{figure}

\Cref{tab:claims} summarizes the findings; \cref{tab:claims-detail} maps each claim to its evidence. \Cref{tab:planned} gives the planned tests with the null each beats. \Cref{tab:app-pairs,tab:app-isolated,tab:app-estimators,tab:app-erasure,tab:app-readability,tab:app-controls,tab:app-geometry,tab:app-identity} report every other discovery quantity summarized in the main text, and \cref{fig:app-rank,fig:app-geometry,fig:app-layers,fig:app-asym,fig:app-floors,fig:app-identity,fig:app-readability,fig:app-erasure} show results by rank, layer, direction and subgroup. The analyses added after discovery are in \cref{app:newexp}.

\begin{table}[t]
\caption{\textbf{The planned test.} Shared-subspace index (SSI; Eq.~\ref{eq:ssi}): the share of Bangla's native
above-floor accuracy that English-counterpart MLR directions recover on Bangla idioms in context, with 95\% component-bootstrap
intervals, at each encoder's validation-selected layer (\cref{tab:models}).
Null: which of the rank-matched random-subspace (R) and permuted-label (P) nulls, 1,999 draws each, reject at Holm-adjusted
$p\le.05$ over the 12 tests (attainable minimum $.006$); bold and \swatch{keytint} mark both. SSI uses the native-reference definition and cutoff in \cref{sec:operations}.}
\label{tab:planned}
\centering
\small
\setlength{\tabcolsep}{2.8pt}
\begin{tabular}{@{}l cc cc@{}}
\toprule
\textbf{\hdtwo{Model}} & \multicolumn{2}{c}{\textbf{F ($M{=}11$)}} & \multicolumn{2}{c}{\textbf{S ($M{=}3$)}} \\
\cmidrule(lr){2-3}\cmidrule(l){4-5}
 & \textbf{SSI [95\% CI]} & \textbf{Null} & \textbf{SSI [95\% CI]} & \textbf{Null} \\
\midrule
BanglaBERT & .16 [.03, .28] & -- & .15 [.00, .29] & -- \\
BanglishBERT & .20 [$-$.06, .43] & -- & .38 [.15, .58] & P \\
mBERT & .20 [$-$.14, .47] & -- & .40 [.06, .77] & -- \\
XLM-R & \hlkey .29 [.10, .46] & \hlkey \textbf{R+P} & \hlkey .47 [.29, .65] & \hlkeyr \textbf{R+P} \\
MuRIL & \hlkey .48 [.34, .62] & \hlkey \textbf{R+P} & .32 [.11, .52] & -- \\
LaBSE & \hlkey .50 [.33, .66] & \hlkey \textbf{R+P} & .30 [.13, .43] & P \\
\bottomrule
\end{tabular}
\end{table}
\begin{table}[t]
\caption{SSI with the planned MLR estimator for all eleven source--target input pairs at each encoder's validation-selected layer.
$^\star$\swatch{keytint}\,Planned pair (Table~\ref{tab:planned}); other pairs have 19-draw diagnostic nulls only.
Each row's denominator is the native readout through the target's own MLR subspace, and its floor is that target's
majority class, so rows with different targets differ in both ($n$ also differs; absolute gains for the idiom with and
without context: \cref{tab:app-isolated}). SSI is undefined (--) when the native reference is within 2\,pp of the floor.}
\label{tab:app-pairs}
\centering
\footnotesize
\setlength{\tabcolsep}{4pt}
\begin{tabular}{@{}lr cccccc@{}}
\toprule
\textbf{Source $\rightarrow$ target} & \textbf{$n$} & \textbf{BanglaBERT} & \textbf{BanglishBERT} & \textbf{mBERT} & \textbf{XLM-R} & \textbf{MuRIL} & \textbf{LaBSE} \\
\midrule
\band{8}{Figurative class F}{rank 11} \\
\EN{} gloss $\rightarrow$ \BN{} gloss & 2,590 & .63 & .87 & .55 & .75 & .99 & 1.06 \\
\EN{} translation $\rightarrow$ \BN{} idiom in ctx & 2,166 & .25 & .28 & .34 & .43 & .46 & .53 \\
\EN{} counterpart $\rightarrow$ \BN{} sentence & 2,590 & .44 & .38 & .30 & .44 & .56 & .61 \\
\EN{} counterpart $\rightarrow$ \BN{} gloss & 2,590 & .28 & .32 & .38 & .31 & .59 & .55 \\
\EN{} counterpart $\rightarrow$ \BN{} idiom, no ctx & 2,590 & .31 & .50 & .64 & .32 & .62 & .71 \\
\hlkeyl \EN{} counterpart $\rightarrow$ \BN{} idiom in ctx$^\star$ & \hlkey 2,166 & \hlkey .16 & \hlkey .20 & \hlkey .20 & \hlkey .29 & \hlkey .48 & \hlkeyr .50 \\
\EN{} counterpart $\rightarrow$ \BN{} masked ctx & 2,166 & .16 & .20 & .09 & .27 & .34 & .39 \\
\BN{} gloss $\rightarrow$ \EN{} gloss & 2,590 & .59 & .62 & .35 & .71 & .84 & .75 \\
\BN{} idiom in ctx $\rightarrow$ \EN{} translation & 2,590 & .81 & .58 & .50 & .50 & .94 & .72 \\
\BN{} idiom, no ctx $\rightarrow$ \EN{} counterpart & 2,558 & .66 & .69 & .45 & .37 & .37 & .37 \\
\BN{} idiom in ctx $\rightarrow$ \EN{} counterpart & 2,558 & .65 & .40 & .30 & .42 & .59 & .57 \\
\addlinespace[2.5pt]
\band{8}{Sentiment S}{rank 3} \\
\EN{} gloss $\rightarrow$ \BN{} gloss & 2,590 & .50 & .85 & .27 & .52 & .91 & .95 \\
\EN{} translation $\rightarrow$ \BN{} idiom in ctx & 2,166 & .49 & .56 & .26 & .52 & .60 & .36 \\
\EN{} counterpart $\rightarrow$ \BN{} sentence & 2,590 & .17 & .53 & .09 & .34 & .49 & .47 \\
\EN{} counterpart $\rightarrow$ \BN{} gloss & 2,590 & .25 & .52 & .24 & .24 & .50 & .57 \\
\EN{} counterpart $\rightarrow$ \BN{} idiom, no ctx & 2,590 & .17 & 1.14 & -- & .24 & .27 & .61 \\
\hlkeyl \EN{} counterpart $\rightarrow$ \BN{} idiom in ctx$^\star$ & \hlkey 2,166 & \hlkey .15 & \hlkey .38 & \hlkey .40 & \hlkey .47 & \hlkey .32 & \hlkeyr .30 \\
\EN{} counterpart $\rightarrow$ \BN{} masked ctx & 2,166 & .31 & .43 & .05 & .24 & .47 & .30 \\
\BN{} gloss $\rightarrow$ \EN{} gloss & 2,590 & .45 & .76 & .62 & .69 & .89 & .76 \\
\BN{} idiom in ctx $\rightarrow$ \EN{} translation & 2,590 & .95 & .75 & .22 & .69 & .76 & .70 \\
\BN{} idiom, no ctx $\rightarrow$ \EN{} counterpart & 2,558 & .07 & .53 & .13 & .06 & .54 & .25 \\
\BN{} idiom in ctx $\rightarrow$ \EN{} counterpart & 2,558 & .35 & .71 & .01 & .30 & .75 & .44 \\
\bottomrule
\end{tabular}
\end{table}
\begin{table}[t]
\caption{\textbf{The idiom with and without its context.} Percentage points above each target's majority floor for the
planned MLR estimator: the native readout through the target's own subspace (the SSI denominator) and the readout through
English-counterpart directions. In context: the 2,166 TEST records with an exact
Bangla span (floors 15.7\% for F and 46.4\% for S); without context (the idiom alone in a neutral
frame): all 2,590 TEST records (floors 13.9\% and 46.4\%). A higher SSI without context can come from a larger
English gain or a smaller native one. Exploratory (19-draw diagnostic nulls). SSI uses the native-reference definition and cutoff in \cref{sec:operations}.}
\label{tab:app-isolated}
\centering
\small
\setlength{\tabcolsep}{3.4pt}
\begin{tabular}{@{}l ccc ccc ccc ccc@{}}
\toprule
\textbf{\hdtwo{Model}} & \multicolumn{6}{c}{\textbf{Figurative class F ($M{=}11$)}} & \multicolumn{6}{c}{\textbf{Sentiment S ($M{=}3$)}} \\
\cmidrule(lr){2-7}\cmidrule(l){8-13}
 & \multicolumn{3}{c}{\textbf{Idiom in context}} & \multicolumn{3}{c}{\textbf{Idiom, no context}} & \multicolumn{3}{c}{\textbf{Idiom in context}}
 & \multicolumn{3}{c}{\textbf{Idiom, no context}} \\
\cmidrule(lr){2-4}\cmidrule(lr){5-7}\cmidrule(lr){8-10}\cmidrule(l){11-13}
 & \textbf{Native} & \textbf{English} & \textbf{SSI} & \textbf{Native} & \textbf{English} & \textbf{SSI} & \textbf{Native} & \textbf{English} & \textbf{SSI} & \textbf{Native} & \textbf{English} & \textbf{SSI} \\
\midrule
BanglaBERT & 15.0 & 2.4 & .16 & 13.3 & 4.1 & .31 & 13.8 & 2.1 & .15 & 7.8 & 1.3 & .17 \\
BanglishBERT & 7.7 & 1.6 & .20 & 8.0 & 4.0 & .50 & 10.8 & 4.1 & .38 & 4.6 & 5.3 & 1.14 \\
mBERT & 5.6 & 1.1 & .20 & 6.3 & 4.1 & .64 & 4.6 & 1.8 & .40 & 1.7 & 0.3 & -- \\
XLM-R & 10.5 & 3.0 & .29 & 6.0 & 1.9 & .32 & 10.0 & 4.7 & .47 & 3.2 & 0.8 & .24 \\
MuRIL & 15.2 & 7.3 & .48 & 11.3 & 7.0 & .62 & 10.5 & 3.4 & .32 & 8.0 & 2.1 & .27 \\
LaBSE & 11.0 & 5.4 & .50 & 6.0 & 4.2 & .71 & 12.4 & 3.7 & .30 & 4.2 & 2.6 & .61 \\
\bottomrule
\end{tabular}
\end{table}
\begin{table}[t]
\caption{Accuracy (\%) of probes trained on Bangla idiom spans projected onto English- / Bangla-estimated directions, per
estimator, at the selected layer. At full rank COV and CPCA span the same subspace (differences are numerical); LEACE
shares that image but uses a covariance-dependent oblique input map with its own output metric. SSI divides the English
gain over the floor by the gain through the \emph{same estimator's} Bangla subspace, which can exceed the full space
(\textit{e.g.}\ MuRIL 30.9 against 29.2, LaBSE 26.6 against 25.6 for MLR, F), so estimators' SSIs have different
denominators; SSI$^{\mathrm{full}}$ divides by the full-space (ambient) gain, which all share. \swatch{keytint}\,Each
encoder's highest SSI. SSI and $\mathrm{SSI}^{\mathrm{full}}$ are undefined when their respective native gains are at most 2\,pp. Exploratory.}
\label{tab:app-estimators}
\centering
\small
\setlength{\tabcolsep}{4pt}
\begin{tabular}{@{}lcccccc@{}}
\toprule
\textbf{Estimator} & \textbf{BanglaBERT} & \textbf{BanglishBERT} & \textbf{mBERT} & \textbf{XLM-R} & \textbf{MuRIL} & \textbf{LaBSE} \\
\midrule
\band{7}{Figurative class F}{rank 11; floor 15.7\%} \\
RAND & 16.5 / 16.5 & 15.6 / 15.6 & 15.9 / 15.9 & 14.7 / 14.7 & 16.2 / 16.2 & 16.6 / 16.6 \\
PCA & 22.3 / 26.8 & 16.7 / 19.6 & 16.9 / 21.9 & 14.8 / 16.4 & 23.2 / 25.4 & 22.0 / 22.7 \\
MLR & 18.1 / 30.7 & 17.2 / 23.4 & 16.8 / 21.2 & 18.7 / 26.2 & 22.9 / 30.9 & 21.1 / 26.6 \\
LDA & 15.6 / 33.4 & 17.1 / 26.9 & 15.7 / 20.8 & 15.9 / 25.0 & 20.2 / 31.5 & 18.4 / 29.7 \\
LEACE & 15.5 / 34.2 & 17.6 / 27.4 & 15.4 / 22.5 & 15.7 / 26.3 & 21.1 / 32.5 & 19.0 / 30.9 \\
CPCA & 25.9 / 36.9 & 18.9 / 25.7 & 19.1 / 26.6 & 17.9 / 23.2 & 32.8 / 36.6 & 25.7 / 31.0 \\
COV & 25.9 / 36.9 & 19.0 / 25.7 & 19.1 / 26.5 & 18.0 / 23.2 & 32.8 / 36.6 & 25.7 / 31.0 \\
\emph{ambient probe} & 30.0 & 23.1 & 20.1 & 23.9 & 29.2 & 25.6 \\
\cmidrule(l){2-7}
SSI$_{\mathrm{MLR}}$ & .16 & .20 & .20 & .29 & .48 & .50 \\
SSI$_{\mathrm{LEACE}}$ & $-$.01 & .17 & $-$.04 & .00 & .32 & .22 \\
SSI$_{\mathrm{COV}}$ & \hlkey .48 & \hlkey .33 & \hlkey .32 & \hlkey .31 & \hlkey .82 & \hlkeyr .65 \\
SSI$^{\mathrm{full}}_{\mathrm{MLR}}$ & .17 & .21 & .25 & .36 & .54 & .55 \\
SSI$^{\mathrm{full}}_{\mathrm{LEACE}}$ & $-$.01 & .27 & $-$.06 & .01 & .40 & .33 \\
SSI$^{\mathrm{full}}_{\mathrm{COV}}$ & .71 & .45 & .78 & .28 & 1.26 & 1.00 \\
\addlinespace[2.5pt]
\band{7}{Sentiment S}{rank 3; floor 46.4\%} \\
RAND & 49.1 / 49.1 & 47.4 / 47.4 & 46.8 / 46.8 & 46.4 / 46.4 & 46.6 / 46.6 & 46.8 / 46.8 \\
PCA & 49.6 / 49.8 & 48.8 / 48.8 & 47.3 / 49.1 & 46.1 / 46.9 & 49.5 / 52.4 & 50.6 / 49.7 \\
MLR & 48.5 / 60.2 & 50.6 / 57.2 & 48.3 / 51.1 & 51.1 / 56.5 & 49.8 / 56.9 & 50.1 / 58.9 \\
LDA & 47.1 / 62.7 & 51.6 / 58.9 & 46.8 / 52.0 & 48.6 / 55.5 & 48.8 / 58.9 & 50.8 / 55.5 \\
LEACE & 48.2 / 63.9 & 51.6 / 58.4 & 46.6 / 52.1 & 48.7 / 57.9 & 50.4 / 60.8 & 50.8 / 58.6 \\
CPCA & 56.0 / 61.8 & 51.9 / 52.7 & 48.6 / 53.9 & 48.0 / 50.5 & 59.7 / 60.9 & 53.6 / 54.7 \\
COV & 56.0 / 61.8 & 51.9 / 52.7 & 48.5 / 53.9 & 48.0 / 50.5 & 59.7 / 60.9 & 53.6 / 54.7 \\
\emph{ambient probe} & 56.7 & 51.7 & 47.5 & 51.7 & 55.8 & 51.4 \\
\cmidrule(l){2-7}
SSI$_{\mathrm{MLR}}$ & .15 & .38 & \hlkey .40 & \hlkey .47 & .32 & .30 \\
SSI$_{\mathrm{LEACE}}$ & .10 & .43 & .02 & .20 & .27 & .36 \\
SSI$_{\mathrm{COV}}$ & \hlkey .62 & \hlkey .88 & .28 & .38 & \hlkey .92 & \hlkeyr .87 \\
SSI$^{\mathrm{full}}_{\mathrm{MLR}}$ & .20 & .78 & -- & .89 & .36 & .75 \\
SSI$^{\mathrm{full}}_{\mathrm{LEACE}}$ & .17 & .97 & -- & .43 & .42 & .88 \\
SSI$^{\mathrm{full}}_{\mathrm{COV}}$ & .93 & 1.04 & -- & .29 & 1.42 & 1.44 \\
\bottomrule
\end{tabular}
\end{table}
\begin{table}[t]
\caption{\textbf{What erasing a rank-11 (F) or rank-3 (S) subspace removes.} Bangla accuracy (\%) of the probe on the
full representation, then the accuracy points it loses when each subspace is removed and the probe refitted (negative:
accuracy rises). \swatch{keytint}\,The English LEACE eraser's drop as a share of the Bangla LEACE eraser's, at most a
quarter in every row. Exploratory.}
\label{tab:app-erasure}
\centering
\small
\setlength{\tabcolsep}{4.5pt}
\begin{tabular}{@{}lccccccc@{}}
\toprule
\textbf{\hdtwo{Model}} & \textbf{\hdtwo{No erasure}} & \multicolumn{3}{c}{\textbf{English-fitted eraser}} & \multicolumn{2}{c}{\textbf{Bangla-fitted eraser}}
& \textbf{\hdtwo{\EN/\BN{} effect}} \\
\cmidrule(lr){3-5}\cmidrule(lr){6-7}
 & & \textbf{LEACE} & \textbf{COV} & \textbf{MLR} & \textbf{LEACE} & \textbf{COV} & \\
\midrule
\band{8}{Figurative class F}{rank 11; floor 15.7\%} \\
BanglaBERT & 30.0 & 0.69 & 0.09 & $-$0.46 & 11.03 & 11.50 & \hlkeyr 6\% \\
BanglishBERT & 23.1 & 0.28 & 0.18 & $-$0.28 & 6.93 & 7.43 & \hlkeyr 4\% \\
mBERT & 20.1 & $-$0.09 & $-$0.55 & 0.18 & 4.25 & 4.48 & \hlkeyr $-$2\% \\
XLM-R & 23.9 & 0.46 & 0.00 & 0.05 & 6.51 & 7.20 & \hlkeyr 7\% \\
MuRIL & 29.2 & 0.55 & 0.28 & $-$0.51 & 9.23 & 9.97 & \hlkeyr 6\% \\
LaBSE & 25.6 & 0.74 & 0.37 & 0.09 & 6.09 & 6.46 & \hlkeyr 12\% \\
\addlinespace[2.5pt]
\band{8}{Sentiment S}{rank 3; floor 46.4\%} \\
BanglaBERT & 56.7 & 1.43 & 1.02 & 0.09 & 11.03 & 10.85 & \hlkeyr 13\% \\
BanglishBERT & 51.7 & 1.48 & 1.66 & 0.18 & 7.39 & 7.39 & \hlkeyr 20\% \\
mBERT & 47.5 & $-$0.14 & $-$0.09 & 0.37 & 4.39 & 4.48 & \hlkeyr $-$3\% \\
XLM-R & 51.7 & 0.74 & 0.51 & $-$0.42 & 7.11 & 7.11 & \hlkeyr 10\% \\
MuRIL & 55.8 & 2.35 & 2.12 & 0.05 & 9.79 & 9.97 & \hlkeyr 24\% \\
LaBSE & 51.4 & 0.23 & 0.51 & 0.28 & 6.69 & 6.65 & \hlkeyr 3\% \\
\bottomrule
\end{tabular}
\end{table}

\begin{table}[t]
\caption{Native readability: TEST accuracy above the PROBE-majority floor (pp) of MLR readouts trained
and tested within each input condition, and of a TF-IDF classifier on the same text (for idiom conditions, the idiom string only).
The English gloss is the text from which F was derived. \swatch{keytint}\,For F, masked context exceeds the idiom span
in four of six encoders and in the median; TF-IDF also favors the context over the idiom string.}
\label{tab:app-readability}
\centering
\small
\setlength{\tabcolsep}{3.3pt}
\begin{tabular}{@{}lcccccccc@{}}
\toprule
\textbf{Input condition} & \textbf{BanglaBERT} & \textbf{BanglishBERT} & \textbf{mBERT} & \textbf{XLM-R} & \textbf{MuRIL} & \textbf{LaBSE} & \textbf{Median} & \textbf{TF-IDF} \\
\midrule
\band{9}{Figurative class F}{rank 11} \\
\EN{} gloss & 37.3 & 40.6 & 43.5 & 48.9 & 53.8 & 55.6 & 46.2 & 41.1 \\
\BN{} gloss & 36.3 & 29.4 & 20.1 & 30.5 & 41.2 & 35.1 & 32.8 & 30.2 \\
\EN{} translation & 14.5 & 17.8 & 17.5 & 21.4 & 22.4 & 21.1 & 19.5 & 19.1 \\
\BN{} sentence & 18.2 & 14.9 & 10.5 & 16.9 & 20.1 & 15.4 & 16.2 & 16.6 \\
\BN{} idiom in ctx & 15.0 & 7.7 & 5.6 & 10.5 & 15.2 & 11.0 & \hlkey 10.8 & \hlkeyr 1.7 \\
\BN{} masked ctx & 13.4 & 10.1 & 6.8 & 11.1 & 12.7 & 11.9 & \hlkey 11.5 & \hlkeyr 13.9 \\
\EN{} counterpart & 6.5 & 8.3 & 8.5 & 14.0 & 16.3 & 14.9 & 11.2 & 9.8 \\
\BN{} idiom, no ctx & 13.3 & 8.0 & 6.3 & 6.0 & 11.3 & 6.0 & 7.2 & 5.8 \\
\addlinespace[2.5pt]
\band{9}{Sentiment S}{rank 3} \\
\EN{} gloss & 13.8 & 18.4 & 14.6 & 20.5 & 20.5 & 21.1 & 19.4 & 15.1 \\
\BN{} gloss & 19.3 & 16.4 & 10.0 & 13.1 & 18.0 & 17.0 & 16.7 & 12.4 \\
\EN{} translation & 11.5 & 14.9 & 11.1 & 17.1 & 14.7 & 15.7 & 14.8 & 11.7 \\
\BN{} sentence & 18.1 & 14.4 & 8.1 & 12.8 & 12.4 & 12.0 & 12.6 & 12.0 \\
\BN{} idiom in ctx & 13.8 & 10.8 & 4.6 & 10.0 & 10.5 & 12.4 & 10.6 & 3.1 \\
\BN{} masked ctx & 12.7 & 11.7 & 6.6 & 10.6 & 10.2 & 10.3 & 10.5 & 12.7 \\
\EN{} counterpart & 4.2 & 5.7 & 4.8 & 7.7 & 7.5 & 8.5 & 6.6 & 4.6 \\
\BN{} idiom, no ctx & 7.8 & 4.6 & 1.7 & 3.2 & 8.0 & 4.2 & 4.4 & 3.6 \\
\bottomrule
\end{tabular}
\end{table}
\begin{table}[t]
\caption{Matched input controls: English-direction (MLR) accuracy on the Bangla idiom span minus accuracy on a control input
for the same sense records (pp), with 95\% paired component-bootstrap intervals (2,000 draws; \textit{post hoc}, unadjusted).
The span reads better (\swatch{keytint}, interval above zero) or worse (\swatch{negtint}, interval below zero) than the control.}
\label{tab:app-controls}
\centering
\footnotesize
\setlength{\tabcolsep}{2.5pt}
\begin{tabular}{@{}lcccc@{}}
\toprule
\textbf{Model} & \textbf{span $-$ masked context} & \textbf{span $-$ idiom w/o context} & \textbf{span $-$ whole sentence} & \textbf{span $-$ Bangla gloss} \\
\midrule
\band{5}{Figurative class F}{rank 11} \\
BanglaBERT & +0.3 [$-$1.7, +2.4] & +1.2 [$-$0.9, +3.1] & \hlneg $-$3.0 [$-$5.1, $-$0.9] & \hlnegr $-$5.3 [$-$7.6, $-$2.9] \\
BanglishBERT & $-$0.5 [$-$2.4, +1.6] & 0.0 [$-$2.0, +2.0] & \hlneg $-$2.5 [$-$4.4, $-$0.7] & \hlnegr $-$5.6 [$-$8.1, $-$3.2] \\
mBERT & +0.5 [$-$1.9, +2.7] & +0.4 [$-$1.4, +2.3] & $-$0.2 [$-$1.8, +1.5] & \hlnegr $-$3.8 [$-$6.1, $-$1.6] \\
XLM-R & 0.0 [$-$2.1, +1.8] & \hlkey +3.9 [+1.8, +5.8] & \hlneg $-$3.6 [$-$5.3, $-$1.7] & \hlnegr $-$5.2 [$-$7.5, $-$2.9] \\
MuRIL & \hlkey +3.0 [+0.7, +5.3] & \hlkey +3.2 [+1.3, +5.2] & $-$2.2 [$-$4.3, +0.2] & \hlnegr $-$14.4 [$-$17.4, $-$11.5] \\
LaBSE & +0.8 [$-$1.3, +2.8] & \hlkey +4.2 [+2.0, +6.3] & \hlneg $-$3.2 [$-$5.2, $-$1.3] & \hlnegr $-$10.8 [$-$13.4, $-$8.1] \\
\addlinespace[2.5pt]
\band{5}{Sentiment S}{rank 3} \\
BanglaBERT & $-$1.8 [$-$3.9, +0.4] & +1.6 [$-$0.5, +3.6] & $-$0.6 [$-$2.4, +1.2] & \hlnegr $-$2.9 [$-$5.2, $-$0.3] \\
BanglishBERT & $-$0.9 [$-$3.3, +1.4] & $-$0.6 [$-$2.6, +1.5] & \hlneg $-$3.8 [$-$6.4, $-$1.5] & \hlnegr $-$4.7 [$-$7.1, $-$2.4] \\
mBERT & +1.5 [$-$0.3, +3.3] & +1.7 [$-$0.1, +3.5] & +1.3 [$-$0.1, +2.8] & $-$0.2 [$-$2.3, +2.1] \\
XLM-R & +2.1 [$-$0.3, +4.8] & \hlkey +3.6 [+1.4, +6.1] & $-$0.5 [$-$2.5, +1.6] & +1.5 [$-$0.7, +3.6] \\
MuRIL & $-$1.4 [$-$4.3, +1.3] & +0.8 [$-$1.2, +2.9] & $-$2.4 [$-$4.9, +0.1] & \hlnegr $-$5.1 [$-$8.1, $-$1.9] \\
LaBSE & +0.6 [$-$1.8, +3.2] & +1.3 [$-$0.6, +3.2] & $-$1.4 [$-$3.2, +0.3] & \hlnegr $-$4.8 [$-$7.3, $-$2.3] \\
\bottomrule
\end{tabular}
\end{table}
\begin{table}[t]
\caption{Alignment and geometry along the sharing ladder (F, rank 11). Retrieval: nearest-neighbor top-1 among
TEST pairs after per-condition centering (chance 0.04--0.05\%). Shared dimensions: $\sum_i \cos^2\theta_i$ over principal
angles between independently estimated subspaces (random: 0.12--0.16). Linear centered kernel alignment (CKA) \citep{kornblith-etal-2019-similarity}
compares the paired English and Bangla TEST representations. On the planned rung, counterpart $\rightarrow$ idiom
in context, pattern subspaces share 1.2--3.6 dimensions (\swatch{keytint}) and filter subspaces
0.14--0.20, near the random level (\swatch{negtint}).}
\label{tab:app-geometry}
\centering
\small
\setlength{\tabcolsep}{4pt}
\begin{tabular}{@{}lcccccc@{}}
\toprule
\textbf{Rung (English $\rightarrow$ Bangla)} & \textbf{BanglaBERT} & \textbf{BanglishBERT} & \textbf{mBERT} & \textbf{XLM-R} & \textbf{MuRIL} & \textbf{LaBSE} \\
\midrule
\band{7}{Retrieval}{top-1 (\%)} \\
gloss $\rightarrow$ gloss & 12.0 & 10.3 & 11.6 & 9.5 & 75.0 & 65.1 \\
translation $\rightarrow$ idiom in ctx & 7.8 & 4.2 & 8.8 & 3.2 & 57.7 & 31.8 \\
counterpart $\rightarrow$ idiom, no ctx & 1.3 & 0.7 & 0.4 & 0.3 & 10.1 & 5.9 \\
counterpart $\rightarrow$ idiom in ctx & 1.5 & 0.5 & 0.5 & 0.8 & 13.9 & 8.4 \\
counterpart $\rightarrow$ masked ctx & 0.2 & 0.1 & 0.3 & 0.1 & 3.1 & 2.4 \\
\addlinespace[2.5pt]
\band{7}{Shared F-pattern dimensions}{COV} \\
gloss $\rightarrow$ gloss & 3.95 & 4.83 & 2.92 & 2.96 & 7.91 & 8.31 \\
translation $\rightarrow$ idiom in ctx & 3.00 & 2.90 & 2.32 & 2.06 & 4.81 & 4.75 \\
counterpart $\rightarrow$ idiom, no ctx & 2.04 & 3.08 & 1.10 & 1.46 & 3.79 & 3.85 \\
counterpart $\rightarrow$ idiom in ctx & \hlkey 1.71 & \hlkey 2.21 & \hlkey 1.22 & \hlkey 1.63 & \hlkey 3.18 & \hlkeyr 3.58 \\
counterpart $\rightarrow$ masked ctx & 1.12 & 1.40 & 1.21 & 1.09 & 1.79 & 2.53 \\
\addlinespace[2.5pt]
\band{7}{Shared F-filter dimensions}{MLR} \\
gloss $\rightarrow$ gloss & 0.33 & 0.68 & 0.18 & 0.50 & 1.24 & 0.71 \\
translation $\rightarrow$ idiom in ctx & 0.23 & 0.26 & 0.24 & 0.18 & 0.38 & 0.23 \\
counterpart $\rightarrow$ idiom, no ctx & 0.15 & 0.20 & 0.22 & 0.16 & 0.17 & 0.22 \\
counterpart $\rightarrow$ idiom in ctx & \hlneg 0.14 & \hlneg 0.19 & \hlneg 0.20 & \hlneg 0.16 & \hlneg 0.20 & \hlnegr 0.17 \\
counterpart $\rightarrow$ masked ctx & 0.15 & 0.17 & 0.15 & 0.15 & 0.11 & 0.21 \\
\addlinespace[2.5pt]
\band{7}{Linear CKA}{} \\
gloss $\rightarrow$ gloss & 0.454 & 0.206 & 0.279 & 0.112 & 0.554 & 0.539 \\
translation $\rightarrow$ idiom in ctx & 0.248 & 0.038 & 0.200 & 0.027 & 0.355 & 0.258 \\
counterpart $\rightarrow$ idiom, no ctx & 0.114 & 0.024 & 0.079 & 0.005 & 0.171 & 0.137 \\
counterpart $\rightarrow$ idiom in ctx & 0.112 & 0.019 & 0.090 & 0.007 & 0.195 & 0.153 \\
counterpart $\rightarrow$ masked ctx & 0.059 & 0.010 & 0.066 & 0.009 & 0.123 & 0.096 \\
\bottomrule
\end{tabular}
\end{table}
\begin{table}[t]
\caption{Unseen-idiom identity: top-1 (\%) retrieval of a TEST idiom's first context from its second among 1,889 candidates
(chance 0.053\%), in rank-64 subspaces estimated from all 3,125 SPACE pairs (mean of three repetitions). Retrieval in the
ambient space collapses from the visible idiom (\swatch{keytint}) to its masked context (\swatch{negtint}). Exploratory.}
\label{tab:app-identity}
\centering
\small
\setlength{\tabcolsep}{4pt}
\begin{tabular}{@{}lcccccc@{}}
\toprule
\textbf{Subspace} & \textbf{BanglaBERT} & \textbf{BanglishBERT} & \textbf{mBERT} & \textbf{XLM-R} & \textbf{MuRIL} & \textbf{LaBSE} \\
\midrule
\band{7}{Idiom span}{the idiom visible} \\
AMBIENT & \hlkey 83.51 & \hlkey 88.88 & \hlkey 85.65 & \hlkey 67.15 & \hlkey 89.07 & \hlkeyr 87.06 \\
RAND & 74.64 & 88.06 & 75.30 & 58.92 & 84.73 & 83.51 \\
PCA & 74.72 & 83.35 & 82.61 & 50.90 & 81.79 & 80.07 \\
COV & 77.32 & 83.64 & 83.83 & 52.83 & 84.09 & 81.23 \\
LEACE & 89.36 & 88.67 & 87.80 & 85.34 & 88.64 & 88.30 \\
LDA & 89.39 & 88.91 & 87.64 & 87.59 & 88.86 & 88.54 \\
\addlinespace[2.5pt]
\band{7}{Masked context}{the idiom replaced by \texttt{[MASK]}} \\
AMBIENT & \hlneg 3.36 & \hlneg 2.28 & \hlneg 0.61 & \hlneg 1.59 & \hlneg 3.23 & \hlnegr 2.25 \\
RAND & 2.12 & 1.48 & 0.29 & 1.01 & 1.51 & 1.24 \\
PCA & 2.99 & 1.96 & 0.56 & 1.59 & 2.75 & 2.36 \\
COV & 3.04 & 2.06 & 0.58 & 1.69 & 2.94 & 2.33 \\
LEACE & 3.79 & 2.14 & 0.98 & 1.91 & 3.63 & 2.94 \\
LDA & 4.21 & 2.65 & 1.16 & 2.06 & 3.89 & 3.15 \\
\bottomrule
\end{tabular}
\end{table}

\subsection{Further measurement diagnostics}
\label{app:measurement}

\para{Identity has a lexical shortcut.} A separate within-Bangla task retrieves an unseen TEST idiom's first context from its second among 1,889 candidates, using rank-64 subspaces estimated on SPACE (\cref{tab:app-identity}). With the idiom visible, top-1 is 67--89\%, and random projections preserve most of it. Masking the idiom reduces top-1 to 0.6--4.2\% (11--80 times nominal chance). Supervised subspaces outperform random ones, but ties among masked inputs complicate interpretation (\cref{app:identity-audit}). The task thus diagnoses a form-matching shortcut rather than measuring cross-lingual identity transfer.

\para{Depth and direction.} At the embedding layer, five encoders have no SSI because native F retention is too close to its floor. Sharing emerges with contextualization, and in MuRIL and LaBSE the cross-lingual score keeps increasing beyond the layer selected for native validation accuracy (\cref{fig:app-layers}). These \textit{post hoc} TEST profiles are descriptive and do not guide layer selection. They show that the native-validation choice need not maximize cross-lingual reading: the native-validation layer has the highest planned SSI only in \LayerBestOnly{}, and elsewhere it lies \LayerGap{} below the best layer. From half depth on, MuRIL and LaBSE keep F SSIs of \LayerLateTop{} at every layer, and XLM-R keeps \LayerLateXLMR{}. The two studies also select with different estimators (MLR and COV) on different splits, and pattern directions beat MLR at the default penalty under both (\cref{sec:rq3}; \cref{tab:followup}). \Cref{app:layers} repeats the readout at every layer for COV, MLR and LEACE, with nulls. Directional SSI also changes across input pairs (\cref{fig:app-asym}), but because its denominator, floor and item cohort change with the target, directional differences do not quantify a language's private information.

\para{Floors and item-level associations.} Transparent-counterpart groups have higher cross accuracy in every encoder but a lower SSI, because their majority floor is 23.7\% against 8.6\% for opaque groups (\cref{fig:app-floors}; \cref{app:theory-metric}). In \textit{post hoc} complete-case regressions, native correctness is associated with 1.5--2.4 times higher odds of cross correctness. The link is an association between two probes' successes, not evidence about model understanding or a causal pathway.

\FloatBarrier
\subsection{Full evidence ledger}
\label{app:claims}

\begin{table}[!htbp]
\caption{\textbf{Full evidence ledger.} Each main claim with its evidence and inferential status. \emph{Confirmatory}: planned
before the campaign. \emph{Prospective}: frozen before the relevant seed-1 outcomes, in one Holm family of 48 tests (encoders whose
test rejects) or, for the idiom alone, in a later plan frozen before its seed-1 outcomes (tests passing). Seed 1 re-splits
the same corpus (932 of its 2,590 TEST records were discovery TEST records), so prospective tests are internal replications.
\emph{Post hoc}: planned after discovery, with correction within the stated families where applicable. The theoretical
results motivate the comparisons; their multiclass empirical diagnostics are \textit{post hoc}.}
\label{tab:claims-detail}
\centering
\small
\setlength{\tabcolsep}{4pt}
\renewcommand{\arraystretch}{1.12}
\begin{tabular}{@{}>{\raggedright\arraybackslash}m{7.6cm}>{\raggedright\arraybackslash}m{3.5cm}>{\raggedright\arraybackslash}m{4.7cm}@{}}
\toprule
\textbf{Claim} & \textbf{Main evidence} & \textbf{Inferential status} \\
\midrule
English idiom directions retain Bangla figurative classes beyond both nulls in XLM-R, MuRIL and LaBSE, and sentiment in XLM-R & \cref{tab:planned,tab:map,tab:app-absolute} & \hlkeyr confirmatory: the planned family (12 tests, Holm); \NzPlannedSurvive{} of \NzPlannedRejections{} survive RANDCOV (\textit{post hoc}) \\
COV beats RAND and PERM in all 12 contextual balanced-accuracy tests on each split; on discovery it also beats RANDCOV and exceeds PCA in all 12 contextual cells. Reseeded F and TAG analyses support class-mean reading & \cref{tab:map,tab:app-balanced,tab:app-taxonomies,tab:app-matched-nulls,tab:app-refit} & \textit{post hoc} families; prospective \hyp{1}: \FuHoneBothF/3 F, \FuHoneBothS/3 S \\
Patterns read more than filters at the default penalty, and the gap closes as the penalty grows (\cref{prop:penalty}) & \cref{fig:estimators}a, e; \cref{tab:app-estimators,tab:tuned-mlr,tab:app-penalty-path} & prospective \hyp{2}: \FuHtwoNF/6 F, \FuHtwoNS/6 S; penalty path exploratory \\
Explicit glosses share more than idiom counterparts & \cref{fig:gap}; \cref{tab:app-pairs,tab:app-concepts,tab:app-taxonomies} & prospective \hyp{3}: \FuHthreeNF/6 F, \FuHthreeNS/6 S; FBN, DOM and TAG \textit{post hoc} \\
Idiom-alone COV reading passes three nulls in \NzProPass/12 prospective tests, and conditional zero-shot transfer passes both nulls in \NzProZsPass/12; context contrasts are analyzed separately & \cref{tab:map,tab:app-idiom-prospective,tab:app-idiom-alone,tab:app-pooling} & prospective (a later frozen plan, three nulls): \NzProPass/12; context contrasts \textit{post hoc} \\
Less gloss-like counterparts support weaker coarse-class reading; items with idiom counterparts have lower English-direction readout accuracy within class & \cref{tab:app-counterpart-type,tab:app-human-items} & \textit{post hoc}; judgments of two volunteers, not adjudicated \\
Patterns transfer under English-only readout training at target-selected layers; stronger penalties improve filters and LEACE, whose default median accuracy SSIs are below zero & \cref{fig:estimators}b, d; \cref{tab:app-zero-shot,tab:app-penalty-path} & \textit{post hoc} \\
LEACE's deficit persists under the tested covariance-estimation changes and shrinks as ridge regularization moves its operator toward COV & \cref{fig:app-leace}; \cref{tab:e3} & exploratory (\expt{3} and the ablation) \\
English-fitted LEACE and INLP remove substantially less Bangla class information than native erasers; the descriptive multiclass overlap ordering holds in 11/12 unrotated COV cells & \cref{fig:read-erase,fig:app-inlp}; \cref{tab:app-erasure,tab:app-erasers} & prospective \hyp{4}: \FuHfourNF/6 F, \FuHfourNS/6 S; multiclass ordering \textit{post hoc} \\
A rotation fitted on held-out pairs increases image overlap and transferred erasure effects & \cref{fig:read-erase}; \cref{tab:app-heldout,tab:app-alignment} & \textit{post hoc}; map fitted on held-out pairs \\
MLP and codelength analyses support class-mean reading; at half depth MLR preserves the two leading F encoders, while COV rankings vary with depth & \cref{tab:app-nonlinear,tab:app-fixed-depth}; \cref{fig:app-layerwise} & \textit{post hoc} \\
Volunteers judge most sampled class assignments apt, most counterparts idioms, and most counterparts equivalent in meaning, assessed separately; the evaluative association is near zero after controlling native recall and is not detected within class & \cref{tab:travel,tab:app-human-agreement,tab:app-human-classes,tab:app-human-items} & descriptive; two volunteers, $\kappa$ \HvKappaRange \\
\bottomrule
\end{tabular}
\end{table}

\FloatBarrier
\section{Robustness and Mechanism Analyses}
\label{app:newexp}

The following analyses assess sensitivity to metrics, label construction, readout family, input pooling, alignment, layer selection, regularization and estimation uncertainty. They use the discovery split unless otherwise stated; the idiom-alone study in \cref{app:idiom-prospective} uses the prospective seed-1 design, and the human item-level analysis appears in \cref{app:human}. The analyses reuse the corresponding frozen caches, taxonomies, layers and preprocessing; the pooling-matched design adds new forward passes.

Reproduction checks recovered the corresponding baseline results before each extension, with discrepancies documented in the repository. The initial rerun of the planned family differed by at most \NxReproAcc\pp{} in accuracy and \NxReproSSI{} in SSI and reproduced its RAND decisions. Subsequent checks covered alignment, erasure, taxonomy reconstruction, the seed-1 penalty path, input-pair readouts, estimator accuracies and saved predictions.

\subsection{Metrics with rescored nulls}
\label{app:balanced}
The planned test scores SSI by accuracy, which a majority class can dominate (floors 15.7\% for F and 46.4\% for S). We rescored the planned family, and the COV family on both splits, by balanced accuracy (macro recall; floor $1/K$) and macro-F1, recomputing all 1,999 RAND and PERM draws under each metric, so that each metric has its own null distribution and Holm correction over its 12 tests (\cref{tab:app-balanced}). Under balanced accuracy the planned family rejects both nulls in \NxPlanBothBal{} of 12 tests (accuracy \NxPlanBothAcc, macro-F1 \NxPlanBothFone), adding \NxBalNew; every accuracy rejection survives. COV beats both nulls in \NxCovBothAccZero{} (accuracy) and \NxCovBothBalZero{} (balanced) of 12 tests on the discovery split and in \NxCovBothAccOne{} and \NxCovBothBalOne{} on seed 1. Accuracy and balanced SSI agree in rank for COV ($\rho=\NxRhoBalCov$; seed 1 \NxRhoBalOne) but only loosely for the planned MLR family ($\rho=\NxRhoBalPlan$): the filter's accuracy SSIs are the most floor-sensitive.

\begin{table}[t]
\caption{\textbf{SSI under three metrics, with every null draw rescored.} Accuracy (the planned metric), balanced accuracy
(macro recall; floor $1/K$) and macro-F1, each with its own floor, for English-counterpart directions on Bangla idioms in
context. \swatch{keytint}\,Bold: beats both nulls (1,999 draws each) after Holm correction over the 12 tests of its band
and metric. Rank agreement of accuracy and balanced SSI: Spearman $\rho=\NxRhoBalPlan$ (MLR), \NxRhoBalCov{} (COV) and
\NxRhoBalOne{} (seed 1). Accuracy SSIs are the original values (\cref{tab:planned,tab:app-estimators} and the
follow-up); the rerun that rescores every null draw reproduces them to within \NxReproSSIAll. SSIs divide by Bangla's own subspace from the same estimator. SSI uses the native-reference definition and cutoff in \cref{sec:operations}.}
\label{tab:app-balanced}
\centering
\small
\setlength{\tabcolsep}{6pt}
\begin{tabular}{@{}l ccc ccc@{}}
\toprule
\textbf{\hdtwo{Model}} & \multicolumn{3}{c}{\textbf{F ($M{=}11$)}} & \multicolumn{3}{c}{\textbf{S ($M{=}3$)}} \\
\cmidrule(lr){2-4}\cmidrule(l){5-7}
 & \textbf{Acc.} & \textbf{Bal.} & \textbf{F1} & \textbf{Acc.} & \textbf{Bal.} & \textbf{F1} \\
\midrule
\band{7}{Discovery, planned estimator (MLR)}{} \\
BanglaBERT & .16 & .33 & .37 & .15 & .15 & .31 \\
BanglishBERT & .20 & \hlkey \textbf{.47} & .47 & .38 & \hlkey \textbf{.37} & \hlkeyr \textbf{.51} \\
mBERT & .20 & .50 & .53 & .40 & .36 & .51 \\
XLM-R & \hlkey \textbf{.29} & \hlkey \textbf{.47} & \hlkey \textbf{.59} & \hlkey \textbf{.47} & \hlkey \textbf{.43} & \hlkeyr \textbf{.56} \\
MuRIL & \hlkey \textbf{.48} & \hlkey \textbf{.59} & \hlkey \textbf{.66} & .32 & \hlkey \textbf{.42} & \hlkeyr \textbf{.60} \\
LaBSE & \hlkey \textbf{.50} & \hlkey \textbf{.60} & \hlkey \textbf{.62} & .30 & .30 & .45 \\
\band{7}{Discovery, pattern estimator (COV)}{} \\
BanglaBERT & \hlkey \textbf{.48} & \hlkey \textbf{.58} & \hlkey \textbf{.61} & \hlkey \textbf{.62} & \hlkey \textbf{.54} & \hlkeyr \textbf{.67} \\
BanglishBERT & .33 & \hlkey \textbf{.50} & \hlkey \textbf{.56} & \hlkey \textbf{.88} & \hlkey \textbf{.72} & \hlkeyr \textbf{.73} \\
mBERT & \hlkey \textbf{.32} & \hlkey \textbf{.53} & \hlkey \textbf{.65} & .28 & \hlkey \textbf{.38} & \hlkeyr \textbf{.59} \\
XLM-R & \hlkey \textbf{.31} & \hlkey \textbf{.53} & \hlkey \textbf{.71} & .38 & \hlkey \textbf{.49} & \hlkeyr \textbf{.66} \\
MuRIL & \hlkey \textbf{.82} & \hlkey \textbf{.87} & \hlkey \textbf{.91} & \hlkey \textbf{.92} & \hlkey \textbf{.94} & \hlkeyr \textbf{.97} \\
LaBSE & \hlkey \textbf{.65} & \hlkey \textbf{.73} & \hlkey \textbf{.78} & \hlkey \textbf{.87} & \hlkey \textbf{.94} & \hlkeyr \textbf{.98} \\
\band{7}{Seed 1, pattern estimator (COV)}{} \\
BanglaBERT & \hlkey \textbf{.66} & \hlkey \textbf{.66} & \hlkey \textbf{.67} & \hlkey \textbf{.56} & \hlkey \textbf{.57} & \hlkeyr \textbf{.70} \\
BanglishBERT & \hlkey \textbf{.73} & \hlkey \textbf{.75} & \hlkey \textbf{.78} & \hlkey \textbf{.84} & \hlkey \textbf{.88} & \hlkeyr \textbf{.93} \\
mBERT & \hlkey \textbf{.61} & \hlkey \textbf{.67} & \hlkey \textbf{.72} & \hlkey \textbf{.58} & \hlkey \textbf{.56} & \hlkeyr \textbf{.68} \\
XLM-R & \hlkey \textbf{.95} & \hlkey \textbf{.94} & \hlkey \textbf{.93} & \hlkey \textbf{1.22} & \hlkey \textbf{1.09} & \hlkeyr \textbf{1.05} \\
MuRIL & \hlkey \textbf{.89} & \hlkey \textbf{.88} & \hlkey \textbf{.87} & \hlkey \textbf{.91} & \hlkey \textbf{.93} & \hlkeyr \textbf{.95} \\
LaBSE & \hlkey \textbf{.67} & \hlkey \textbf{.69} & \hlkey \textbf{.73} & \hlkey \textbf{1.00} & \hlkey \textbf{.99} & \hlkeyr \textbf{.99} \\
\bottomrule
\end{tabular}
\end{table}

\para{SSI and absolute effects.} A ratio with a small denominator can be unstable, so \cref{tab:app-absolute} restates the planned tests in percentage points. Native gains are \NxPlanNativeGain\pp, far from the 2\pp{} at which SSI is left undefined, and English-direction gains are \NxPlanCrossGain\pp. Within a test the native accuracy and the floor are shared by the observed directions and every null draw, so each test on SSI is exactly a test on raw cross-lingual accuracy: the same null comparison expressed in raw or above-floor accuracy yields identical $p$-values, and the ratio only rescales magnitudes for comparison across cells. SSI's codelength analog ranks cells as accuracy SSI does (\cref{app:nonlinear}).

\begin{table}[t]
\caption{\textbf{The planned tests as absolute effects.} Percentage points above the majority floor on Bangla idioms in
context: the native probe (\emph{Native}, the SSI denominator), English-counterpart MLR directions (\emph{English}), and
the English directions' margin over the mean of each null. Native accuracy and floor are the same for the observed
directions and every null draw, so each null test on SSI is a test on raw cross-lingual accuracy. \swatch{keytint}\,The
null rejects after Holm correction over the 12 planned tests; bold SSI: both do. SSI uses the native-reference definition and cutoff in \cref{sec:operations}.}
\label{tab:app-absolute}
\centering
\footnotesize
\setlength{\tabcolsep}{4.2pt}
\begin{tabular}{@{}l ccccc ccccc@{}}
\toprule
\textbf{\hdtwo{Model}} & \multicolumn{5}{c}{\textbf{F ($M{=}11$)}} & \multicolumn{5}{c}{\textbf{S ($M{=}3$)}} \\
\cmidrule(lr){2-6}\cmidrule(l){7-11}
 & \textbf{Native} & \textbf{English} & \textbf{\textit{vs.}\ RAND} & \textbf{\textit{vs.}\ PERM} & \textbf{SSI} & \textbf{Native} & \textbf{English} & \textbf{\textit{vs.}\ RAND} & \textbf{\textit{vs.}\ PERM} & \textbf{SSI} \\
\midrule
BanglaBERT & 15.0 & 2.4 & +0.3 & +1.3 & .16 & 13.8 & 2.1 & +0.6 & +0.9 & .15 \\
BanglishBERT & 7.7 & 1.6 & +1.4 & +1.5 & .20 & 10.8 & 4.1 & +2.8 & \hlkey +3.0 & .38 \\
mBERT & 5.6 & 1.1 & +1.2 & +0.7 & .20 & 4.6 & 1.8 & +1.4 & +1.2 & .40 \\
XLM-R & 10.5 & 3.0 & \hlkey +4.0 & \hlkey +3.7 & \hlkey \textbf{.29} & 10.0 & 4.7 & \hlkey +4.2 & \hlkey +4.1 & \hlkeyr \textbf{.47} \\
MuRIL & 15.2 & 7.3 & \hlkey +6.7 & \hlkey +7.2 & \hlkey \textbf{.48} & 10.5 & 3.4 & +2.4 & +2.5 & .32 \\
LaBSE & 11.0 & 5.4 & \hlkey +4.6 & \hlkey +4.6 & \hlkey \textbf{.50} & 12.4 & 3.7 & +2.7 & \hlkey +2.8 & .30 \\
\bottomrule
\end{tabular}
\end{table}

\subsection{Zero-shot readout}
\label{app:zero-shot}
The subspace is fitted on English SPACE as before, but the probe is trained on English PROBE items in that subspace, with its penalty tuned on English VALIDATION, and applied unchanged to Bangla TEST items. Readout fitting and penalty selection use no Bangla labels; the layer remains selected by within-Bangla validation accuracy (\cref{sec:models}), and the global RMS scale uses unlabeled SPACE rows from all conditions (\cref{app:method-inputs}). Bangla inputs are centered on their own unlabeled SPACE mean (the main setting) or on the English mean. The zero-shot SSI divides the zero-shot gain over the floor by the supervised native gain, so it shares a scale with the supervised SSI. The nulls are 999 random subspaces and, for COV, 999 estimators fitted to permuted labels, each read by an English-trained probe; Holm runs over the 12 tests of each estimator and centering.

Through COV, zero-shot SSIs beat both nulls in \NxZsBoth{} of 12 tests (\NxZsBothSrc{} with English-mean centering). Because the nulls' zero-shot accuracies lie below the floor on average, we also test the floor itself: zero-shot accuracy exceeds it in \NxZsAboveFloor{} of 12 tests after Holm correction, \NxZsBothAndFloor{} of them also beating both nulls. The paired supervised-minus-zero-shot accuracy difference is \NxZsSupGap\pp, with an interval that includes zero in \NxZsSupInside{} of 12 cells, every MuRIL and LaBSE cell among them. Through MLR, LEACE and the full space, median accuracy-based zero-shot SSIs are below zero (medians \NxZsMlrMedian, \NxZsLeaceMedian, \NxZsAmbMedian) and beat the RAND null in only \NxZsMlrRand{} (MLR) and \NxZsLeaceRand{} (LEACE) of 12 tests. For S, the English full-space classifier falls below the majority floor in all \NxZsAmbBelowS{} encoders (\NxZsAmbSAcc\pp; \NxZsAmbSrcSAcc\pp{} with English-mean centering). Balanced accuracy shows what the filters keep: zero-shot MLR retains \NxZsMlrBalF{} of the native balanced F signal and beats the rescored RAND null in \NxZsMlrBalRand{} of 12 tests. Transfer is not specific to the planned pair: through COV, English glosses transfer zero-shot to Bangla glosses with F SSIs of \NxZsGlossCovF, and translations transfer to idioms in context with \NxZsTransCovF, while full-space classifiers trained on translations reach \NxZsTransAmbF.

\subsection{Concepts not built from English}
\label{app:concepts}
F is built from English glosses, which favors English-gloss sources. FBN, built from the Bangla glosses that nearly translate them, favors the Bangla-gloss target instead, so in the gloss contrast below only DOM is free of both gloss fields. FBN rebuilds the taxonomy from the \emph{Bangla} figurative glosses: $k$-means ($k=12$) over mean-pooled NLLB-200 encoder embeddings, fitted on SPACE only (re-seed ARI \NxFBNAri). DOM takes the corpus's usage-domain metadata, drops register descriptors that apply to most idioms (\emph{everyday conversation}, \emph{colloquial} and the like), keeps the 12 most frequent remaining SPACE tags and labels \NxDOMCoverage\% of records; its TEST majority class holds \NxDOMMajority\%. Both agree only weakly with F (AMI \NxFBNAmiF{} and \NxDOMAmiF), so they are distinct partitions rather than relabelings. Random subspaces already recover much of FBN (null mean SSI \NxFBNRandMean), which makes its nulls demanding.

English-counterpart COV directions beat both nulls for FBN in \NxFBNCOVBoth{} encoders (SSI \NxFBNCovSSI{} across all six) and for DOM in \NxDOMCOVBoth{} (MLR: \NxFBNMLRBoth{} and \NxDOMMLRBoth); \NxThreeConcepts{} pass for F, FBN and DOM alike (\cref{tab:app-concepts}). The gloss advantage of \hyp{3} reappears without English-built labels: English-gloss directions beat English-counterpart directions on Bangla glosses for FBN in \NxGlossFBN{} of \NxGlossFBNOf{} encoder--estimator cells (\NxGlossFBNpp\pp), and for DOM in \NxGlossDOM{}, with all \NxGlossDOMPos{} estimates positive.

\begin{table}[t]
\caption{\textbf{Two concepts not built from English glosses.} SSI [95\% CI] of English-counterpart directions on Bangla
idioms in context. FBN clusters SPACE's \emph{Bangla} figurative glosses ($k=12$) in the embedding space of NLLB-200's
never-probed encoder; DOM keeps the 12 most frequent topical usage-domain tags in SPACE (literature, emotions, politics, family, relationships, religion, finance, criticism, business, education, emotional expression, health) after
removing register descriptors such as \emph{everyday conversation}, and labels \NxDOMCoverage\% of records.
\swatch{keytint}\,Bold: beats both nulls (999 draws) after Holm correction over the six encoders. Last columns: COV
accuracy on Bangla glosses through English-gloss minus English-counterpart directions, the \hyp{3} contrast for the new
concepts; \swatch{keytint}\,Holm-adjusted $p\le.05$. SSIs divide by Bangla's own subspace from the same estimator. SSI uses the native-reference definition and cutoff in \cref{sec:operations}.}
\label{tab:app-concepts}
\centering
\small
\setlength{\tabcolsep}{3.5pt}
\begin{tabular}{@{}l cc cc cc@{}}
\toprule
 & \multicolumn{2}{c}{\textbf{FBN: Bangla-gloss clusters}} & \multicolumn{2}{c}{\textbf{DOM: usage domain}} & \multicolumn{2}{c}{\textbf{Gloss $-$ counterpart (pp)}} \\
\cmidrule(lr){2-3}\cmidrule(lr){4-5}\cmidrule(l){6-7}
\textbf{Model} & \textbf{MLR} & \textbf{COV} & \textbf{MLR} & \textbf{COV} & \textbf{FBN} & \textbf{DOM} \\
\midrule
BanglaBERT & .47\,{\scriptsize[.37, .58]} & .74\,{\scriptsize[.66, .82]} & .11\,{\scriptsize[$-$.04, .27]} & .10\,{\scriptsize[$-$.11, .30]} & \hlkey +8.0 & \hlkeyr +6.8 \\
BanglishBERT & .50\,{\scriptsize[.38, .63]} & .58\,{\scriptsize[.49, .66]} & $-$.57\,{\scriptsize[$-$1.05, $-$.16]} & -- & \hlkey +5.1 & +2.4 \\
mBERT & .73\,{\scriptsize[.57, .90]} & .42\,{\scriptsize[.30, .53]} & .03\,{\scriptsize[$-$.44, .37]} & .25\,{\scriptsize[$-$.11, .56]} & \hlkey +13.1 & +1.2 \\
XLM-R & \hlkey \textbf{.67}\,{\scriptsize[.52, .84]} & \hlkey \textbf{.63}\,{\scriptsize[.52, .76]} & .09\,{\scriptsize[$-$.07, .29]} & \hlkey \textbf{.40}\,{\scriptsize[.00, .90]} & \hlkey +9.7 & +3.3 \\
MuRIL & \hlkey \textbf{.76}\,{\scriptsize[.64, .89]} & \hlkey \textbf{.86}\,{\scriptsize[.79, .93]} & \hlkey \textbf{.43}\,{\scriptsize[.23, .59]} & \hlkey \textbf{.73}\,{\scriptsize[.55, .92]} & \hlkey +9.6 & \hlkeyr +4.0 \\
LaBSE & .79\,{\scriptsize[.63, .98]} & \hlkey \textbf{.91}\,{\scriptsize[.82, .99]} & \hlkey \textbf{.34}\,{\scriptsize[.07, .56]} & \hlkey \textbf{.52}\,{\scriptsize[.25, .76]} & \hlkey +7.2 & +1.7 \\
\bottomrule
\end{tabular}
\end{table}

\subsection{The LEACE ablation}
\label{app:leace}
\begin{figure}[t]
\centering
\begin{subfigure}[t]{0.3617\textwidth}
\centering
\includegraphics[width=\linewidth,viewport=0.84 0.84 169.89 146.76,clip]{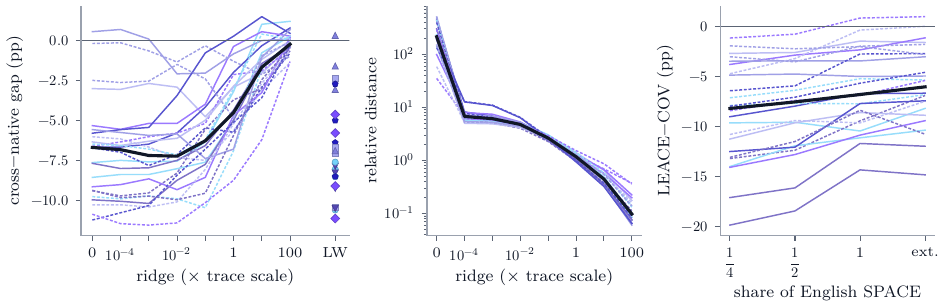}
\caption{ridge path}
\end{subfigure}\hfill
\begin{subfigure}[t]{0.2994\textwidth}
\centering
\includegraphics[width=\linewidth,viewport=171.57 0.84 311.49 146.76,clip]{figures/app_leace_ablation.pdf}
\caption{LEACE operator to COV}
\end{subfigure}\hfill
\begin{subfigure}[t]{0.2989\textwidth}
\centering
\includegraphics[width=\linewidth,viewport=313.10 0.84 452.82 146.76,clip]{figures/app_leace_ablation.pdf}
\caption{rows for $\Sigma$, fixed image}
\end{subfigure}
\caption{\textbf{What drives LEACE's cross-lingual deficit.} (a)~LEACE$-$COV accuracy through English directions minus the same difference through Bangla directions, on Bangla idioms in context, along the whitening ridge (thin: split--model--label cells, solid F, dashed S; thick: median; markers: Ledoit--Wolf). (b)~Relative distance of the LEACE operator to the COV projection along the same path. (c)~With the English image fixed, LEACE$-$COV accuracy on Bangla as $\Sigma$ is estimated from more English rows (ext.: SPACE, PROBE and VALIDATION inputs). Among the tested changes, the deficit closes as ridge regularization moves LEACE toward the COV projection. Encoders: \encoderkey. Exploratory.}
\label{fig:app-leace}
\end{figure}

\expt{3} showed that covariance matching does not improve LEACE retention. We examine covariance weighting and estimation effects by varying regularization, the number of covariance-estimation rows and the probe penalty, while tracking retention and the operator's distance from COV (\cref{fig:app-leace}). Each analysis reads English- and Bangla-fitted LEACE on contextual Bangla idioms on both splits. \emph{Ridge}: adding $\lambda\cdot\tr(\Sigma)/D$ to $\Sigma$ before whitening, for $\lambda$ from 0 (the original Moore--Penrose setting) to 100, or Ledoit--Wolf shrinkage \citep{ledoit-wolf-2004-well}, interpolates from the oblique eraser to the orthogonal COV projection. \emph{Rows for $\Sigma$}: with the English COV image fixed, $\Sigma$ is estimated from 25\% or 50\% of English SPACE (three component subsamples each), all of it, or SPACE with the PROBE and VALIDATION inputs added (labels unused). \emph{Probe penalty}: $C\in\{.01,\dots,100\}$ for the COV and LEACE readouts. 

\para{Retention effects.} The cross-minus-native LEACE$-$COV difference at the original setting is \NxLeaceDeficit\pp, negative in \NxLeaceDeficitNeg{} of \NxLeaceDeficitOf{} cells and with intervals below zero in \NxLeaceDeficitCI. Ledoit--Wolf shrinkage changes the deficit by a median \NxLeaceLWShift\pp{} (at most \NxLeaceLWShiftMax). The fixed-image LEACE$-$COV difference on Bangla is a median \NxSigmaQuarter, \NxSigmaFull{} and \NxSigmaExt\pp{} as rows grow, \NxSigmaShrink\% smaller with four times the rows, and the probe penalty moves the deficit by a median \NxProbeCSpread\pp{} (at most \NxProbeCSpreadMax). Among the tested changes, the deficit closes as ridge regularization moves LEACE toward COV: the operator's relative distance to the COV projection falls from \NxLeaceDistZero{} at $\lambda=0$ (\NxLeaceDistLW{} under Ledoit--Wolf) to \NxLeaceDistHundred{} at $\lambda=100$, where the deficit is \NxLeaceDeficitHundred\pp. 

\para{Erasure and operator checks.} On held-out English, LEACE removes \NxEnglishEraseShareF{} of the above-floor F signal, against \NxCrossEraseShareF{} on Bangla. Total- and within-class whitening give the same operators (\cref{lem:woodbury}; relative difference at most \NxWoodbury), and all fixed-image operators are idempotent to \NxIdempotence.

\subsection{The pooling-matched design}
\label{app:pooling}
Masked contexts differ from idiom spans in two ways: the input (idiom masked or not) and the pooled tokens (context or span). Two new conditions complete a $2{\times}2$ design on the same sentence: \cond{bn-figctx} pools the context tokens of the sentence with the idiom present, and \cond{bn-maskslot} pools the mask token of the masked sentence. All cells were re-extracted with the original pinned model revisions, layers, maximum length and normalization (minimum cosine to the cached features \NxPoolCos), and readouts are restricted to the \NxPoolN{} records valid in all four cells, so every contrast is paired. On average, the context cells differ by at most \NxPoolTokDiff{} pooled tokens. \Cref{tab:app-pooling} gives the contrasts. With the context tokens fixed, the idiom's presence raises native F accuracy by \NxPoolCtxAmbF\pp{} (significant in \NxPoolCtxAmbFSig{} encoders) but changes accuracy through English COV directions by only \NxPoolCtxCovF\pp{} (\NxPoolCtxCovFSig{} significant). With the idiom present, native probes read the context tokens better than the span (span minus context \NxPoolSpanAmbF\pp), while English directions read both about equally (\NxPoolSpanCovF\pp). For S, the span position carries more than the mask slot through English directions (\NxPoolSlotCovS\pp, significant in \NxPoolSlotCovSSig{} encoders).

\begin{table}[t]
\caption{\textbf{The pooling-matched $2{\times}2$ design.} Paired accuracy differences (pp) on the same \NxPoolN{}
records. \emph{Idiom$-$mask (context pooled)}: context tokens pooled from the sentence with the idiom (\cond{bn-figctx}) minus
the same positions with the idiom masked (\cond{bn-context}). \emph{Span$-$context}: idiom span (\cond{bn-fig}) minus the
context tokens of the same sentence. \emph{Idiom$-$mask (span pooled)}: \cond{bn-fig} minus the masked slot
(\cond{bn-maskslot}). Native: full-space Bangla probe; English: English-counterpart COV directions.
\swatch{keytint}/\swatch{negtint}\,two-sided component wild-sign $p\le.05$, unadjusted.}
\label{tab:app-pooling}
\centering
\footnotesize
\setlength{\tabcolsep}{4pt}
\begin{tabular}{@{}l cc cc c@{}}
\toprule
 & \multicolumn{2}{c}{\textbf{Idiom$-$mask (context pooled)}} & \multicolumn{2}{c}{\textbf{Span$-$context (idiom present)}}
 & \textbf{Idiom$-$mask (span pooled)} \\
\cmidrule(lr){2-3}\cmidrule(lr){4-5}\cmidrule(l){6-6}
\textbf{Model} & \textbf{Native} & \textbf{English} & \textbf{Native} & \textbf{English} & \textbf{English} \\
\midrule
\band{6}{Figurative classes F}{} \\
BanglaBERT & \hlkey +4.9 & +1.2 & $-$1.0 & $-$0.9 & \hlkeyr +9.6 \\
BanglishBERT & \hlkey +4.3 & +0.9 & \hlneg $-$5.2 & $-$1.3 & $-$0.3 \\
mBERT & +0.4 & $-$0.2 & $-$1.6 & $-$1.8 & $-$1.5 \\
XLM-R & +1.6 & $-$0.2 & \hlneg $-$3.3 & $-$0.3 & \hlnegr $-$4.9 \\
MuRIL & \hlkey +3.9 & +0.7 & $-$1.9 & +1.9 & \hlkeyr +6.0 \\
LaBSE & \hlkey +2.5 & +0.9 & \hlneg $-$3.1 & \hlneg $-$3.6 & +1.7 \\
\band{6}{Sentiment S}{} \\
BanglaBERT & \hlkey +6.7 & \hlkey +2.4 & \hlneg $-$4.0 & $-$1.9 & \hlkeyr +9.2 \\
BanglishBERT & \hlkey +3.8 & +1.1 & \hlneg $-$6.0 & $-$1.5 & \hlkeyr +3.8 \\
mBERT & +1.1 & +0.5 & $-$1.5 & +2.0 & \hlkeyr +3.0 \\
XLM-R & $-$0.6 & $-$0.7 & \hlneg $-$3.8 & $-$0.6 & +2.0 \\
MuRIL & \hlkey +4.2 & \hlkey +2.9 & $-$2.2 & +0.6 & \hlkeyr +9.6 \\
LaBSE & \hlkey +1.9 & +0.6 & \hlneg $-$3.9 & $-$1.1 & \hlkeyr +5.5 \\
\bottomrule
\end{tabular}
\end{table}

\subsection{Nonlinear readouts and description length}
\label{app:nonlinear}
\para{MLP probes.} One-hidden-layer MLPs (hidden size 64 or 256 and $L_2$ strength chosen on VALIDATION by mean accuracy over three seeds) replace the linear probe for English-fitted and Bangla-fitted MLR, COV and LEACE subspaces. Within Bangla they gain \NxMlpGainF\pp{} over linear probes on the full representation for F, but the SSI moves beyond its bootstrap interval in only \NxMlpSig{} of \NxMlpCells{} subspace cells, with a median absolute change of \NxMlpCovDiff{} for COV (\cref{tab:app-nonlinear}). The class-mean reading pattern persists with the tested MLP readouts.

\para{Description length.} We compute online codelength on PROBE and VALIDATION items in a component-respecting order (blocks from 0.1\% to 100\%), sending the first block with a uniform code and each later block with an L2 softmax probe over all $K$ classes, mixed with a Krichevsky--Trofimov label-frequency code \citep{krichevsky-trofimov-1981-performance}. The probe penalty and mixture weight are chosen per block from a $5\times4$ grid whose index is charged, so the code stays decodable. Every block's predictive distribution covers all $K$ classes, and the frequency mixture keeps classes absent from an early training prefix encodable. Under this code, the full representation compresses the labels by \NxMdlPrior{} times relative to the frequency code (above 1 in \NxMdlPriorAbove{} of \NxMdlCells{} cells) and \NxMdlUniform{} times relative to a uniform code. A bits-based share, the codelength saving of English directions over the frequency code relative to the saving of Bangla's own, is \NxBitsCovF{} for COV on F and \NxBitsCovS{} on S, \NxBitsLeace{} for LEACE, and rank-correlates with accuracy SSI ($\rho=\NxBitsRho$).

\begin{table}[t]
\caption{\textbf{Nonlinear readouts and description length.} \emph{MLP $-$ linear}: TEST accuracy gain of a
one-hidden-layer MLP over the linear probe on the full Bangla representation (idioms in context). \emph{SSI}: English
COV directions read by the linear probe (as in \cref{tab:app-estimators}) and by the MLP (hidden size and $L_2$ strength
chosen on VALIDATION; mean of three seeds); \swatch{keytint}\,the bootstrap interval of the MLP's difference from the
same run's linear readout (which reproduces the linear SSI to within \NxMlpLinRepro) excludes zero. \emph{Compression}:
the Krichevsky--Trofimov frequency code's label codelength over the full representation's online codelength
(above 1: information beyond label frequencies). \emph{Bits-based share}: English directions' codelength saving over the
frequency code as a share of the target's own directions' saving, an information-theoretic analog of SSI. SSIs divide by Bangla's own subspace from the same estimator. SSI uses the native-reference definition and cutoff in \cref{sec:operations}.}
\label{tab:app-nonlinear}
\centering
\footnotesize
\setlength{\tabcolsep}{4pt}
\begin{tabular}{@{}l cc cc cc cccc@{}}
\toprule
 & \multicolumn{2}{c}{\textbf{MLP $-$ linear (pp)}} & \multicolumn{2}{c}{\textbf{SSI, linear/MLP}} & \multicolumn{2}{c}{\textbf{Compression}}
 & \multicolumn{4}{c}{\textbf{Bits-based share}} \\
\cmidrule(lr){2-3}\cmidrule(lr){4-5}\cmidrule(lr){6-7}\cmidrule(l){8-11}
\textbf{Model} & \textbf{F} & \textbf{S} & \textbf{F} & \textbf{S} & \textbf{F} & \textbf{S} & \textbf{F COV} & \textbf{F LEACE} & \textbf{S COV} & \textbf{S LEACE} \\
\midrule
BanglaBERT & +3.4 & +8.7 & .48/\hlkey .55 & .62/.53 & 1.19 & 1.27 & .54 & .01 & .29 & .00 \\
BanglishBERT & +3.6 & +5.2 & .33/.34 & .88/.72 & 1.08 & 1.11 & .24 & .09 & .51 & .14 \\
mBERT & +5.3 & +4.1 & .32/.31 & .28/.29 & 1.06 & 1.05 & .24 & .00 & .15 & $-$.32 \\
XLM-R & +2.1 & +4.0 & .31/.32 & .38/.45 & 1.06 & 1.07 & .25 & $-$.01 & .18 & $-$.02 \\
MuRIL & +4.8 & +7.6 & .82/.82 & .92/.93 & 1.19 & 1.16 & .77 & .19 & .90 & .03 \\
LaBSE & +5.9 & +7.0 & .65/.71 & .87/.78 & 1.13 & 1.08 & .69 & .14 & .95 & .07 \\
\bottomrule
\end{tabular}
\end{table}

\subsection{Hubness correction and alignment}
\label{app:alignment}
Paired retrieval in the main text uses nearest-neighbor cosine after per-condition centering. We add CSLS ($k=10$) and an orthogonal Procrustes map fitted on row-normalized paired SPACE records, so no map sees a TEST pair; a variant restricts the map to the top 128 principal directions of the SPACE pairs. In the four weaker encoders, the map with CSLS lifts top-1 retrieval of translation pairs from \NxRetTransBefore{} to \NxRetTransAfter{} and of gloss pairs from \NxRetGlossBefore{} to \NxRetGlossAfter, but idiom pairs move from \NxRetIdiomBefore{} to \NxRetIdiomAfter{} (at most \NxRetIdiomMax{} in any setting; \cref{tab:app-alignment}). The map fits SPACE pairs (mean pair cosine \NxAlignFitSpace) but not TEST pairs (\NxAlignFitTest, against \NxAlignFitTestBefore{} before), so it does not generalize at the instance level.

Carried into Bangla coordinates by the same map, English COV directions share \NxAlignDimsAfter{} of 11 dimensions with Bangla's image (\NxAlignDimsBefore{} before), read \NxAlignSSIAfter{} of the native signal (\NxAlignSSIBefore{} before; \NxAlignReducedSSI{} with the 128-direction map) and erase \NxAlignEraseF{} of the native effect for F and \NxAlignEraseS{} for S. The analogous multiclass ordering is observed in \NxAlignOK{} of \NxAlignCells{} aligned cells; the exceptions, \NxAlignFail, exceed the overlap by \NxAlignFailGap. Zero-shot, an English classifier reads Bangla through rotated COV directions with SSIs of \NxAlignZsCovF{} for F and \NxAlignZsCovS{} for S (higher than unaligned in \NxAlignZsCovUp{} of 12 cells), but through rotated LEACE coordinates at \NxAlignZsLeace{} (median \NxAlignZsLeaceMedian), consistent with the covariance-dependent transfer mechanism allowed by \cref{prop:portable}: aligning images alone does not determine classifier transfer. Removing rotated MLR directions changes Bangla accuracy by at most \NxAlignEraseMlrMax\pp, an empirical result; \cref{lem:containment} characterizes complete mean removal but does not bound this accuracy change. Because this map is fitted on the SPACE records that also define the class means, \cref{app:heldout} refits it on held-out pairs, and the main text reports the held-out map.

\begin{table}[t]
\caption{\textbf{Hubness correction and a learned alignment.} Left: top-1 retrieval of the paired TEST item (fractions in $[0,1]$) among all
candidates after per-condition centering (NN: nearest-neighbor cosine, the original setting) and after an orthogonal
Procrustes map fitted on paired SPACE records, with CSLS scoring ($k=10$). Right: English COV directions before and after
the same map carries them into Bangla coordinates: dimensions shared with the Bangla COV image, the share of native
above-floor accuracy they read, and the share of the native erasure effect their removal achieves (raw: original values,
\cref{tab:app-estimators,tab:app-erasure}). The map is fitted on the SPACE records that also define the class means, so
the aligned columns diagnose how far a rotation recovers the shared geometry; they are not evidence of unaligned sharing.}
\label{tab:app-alignment}
\centering
\footnotesize
\setlength{\tabcolsep}{3.2pt}
\begin{tabular}{@{}l cc cc cc cc cc cc@{}}
\toprule
 & \multicolumn{6}{c}{\textbf{Top-1 retrieval, English $\rightarrow$ Bangla}} & \multicolumn{6}{c}{\textbf{English COV directions, F ($M{=}11$)}} \\
\cmidrule(lr){2-7}\cmidrule(l){8-13}
 & \multicolumn{2}{c}{\textbf{Translations}} & \multicolumn{2}{c}{\textbf{Glosses}} & \multicolumn{2}{c}{\textbf{Idiom pairs}}
 & \multicolumn{2}{c}{\textbf{Shared dims}} & \multicolumn{2}{c}{\textbf{Read share (SSI)}} & \multicolumn{2}{c}{\textbf{Erase share}} \\
\cmidrule(lr){2-3}\cmidrule(lr){4-5}\cmidrule(lr){6-7}\cmidrule(lr){8-9}\cmidrule(lr){10-11}\cmidrule(l){12-13}
\textbf{Model} & \textbf{NN} & \textbf{P+CSLS} & \textbf{NN} & \textbf{P+CSLS} & \textbf{NN} & \textbf{P+CSLS} & \textbf{raw} & \textbf{aligned} & \textbf{raw} & \textbf{aligned} & \textbf{raw} & \textbf{aligned} \\
\midrule
BanglaBERT & .32 & .64 & .12 & .43 & .02 & .02 & 1.7 & 6.0 & .48 & .89 & .01 & .49 \\
BanglishBERT & .16 & .27 & .10 & .21 & .01 & .01 & 2.2 & 5.1 & .33 & .83 & .02 & .27 \\
mBERT & .18 & .38 & .12 & .23 & .00 & .01 & 1.2 & 5.2 & .32 & .91 & $-$.12 & .17 \\
XLM-R & .13 & .29 & .10 & .20 & .01 & .01 & 1.6 & 4.8 & .31 & 1.05 & .00 & .40 \\
MuRIL & .89 & .93 & .75 & .83 & .14 & .11 & 3.2 & 7.4 & .82 & .96 & .03 & .56 \\
LaBSE & .86 & .91 & .65 & .75 & .08 & .07 & 3.6 & 7.0 & .65 & .96 & .06 & .63 \\
\bottomrule
\end{tabular}
\end{table}

\subsection{Layers}
\label{app:layers}
\begin{figure}[t]
\centering
\begin{subfigure}[t]{0.3372\textwidth}
\centering
\includegraphics[width=\linewidth,viewport=0.84 131.85 157.63 240.36,clip]{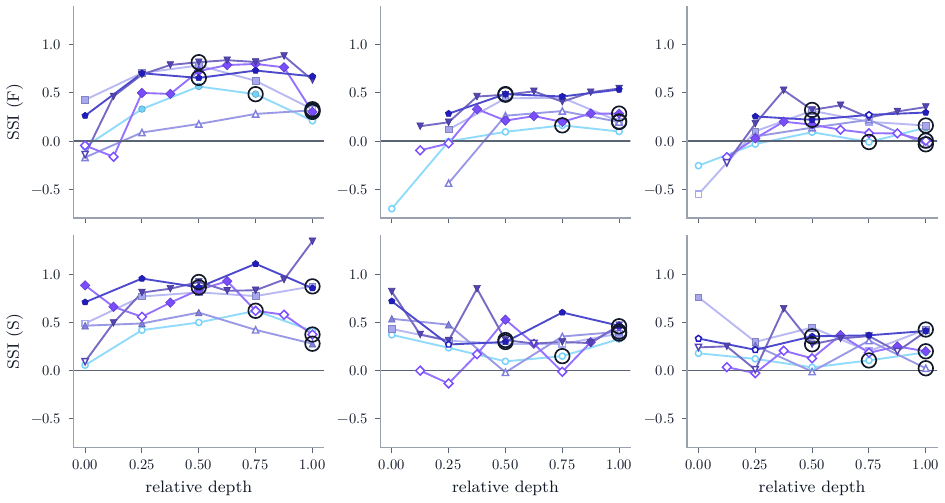}
\caption{COV, F}
\end{subfigure}\hfill
\begin{subfigure}[t]{0.3114\textwidth}
\centering
\includegraphics[width=\linewidth,viewport=160.04 131.85 304.83 240.36,clip]{figures/app_layerwise.pdf}
\caption{MLR, F}
\end{subfigure}\hfill
\begin{subfigure}[t]{0.3114\textwidth}
\centering
\includegraphics[width=\linewidth,viewport=307.24 131.85 452.03 240.36,clip]{figures/app_layerwise.pdf}
\caption{LEACE, F}
\end{subfigure}
\par\medskip
\begin{subfigure}[t]{0.3371\textwidth}
\centering
\includegraphics[width=\linewidth,viewport=0.84 0.84 158.39 130.17,clip]{figures/app_layerwise.pdf}
\caption{COV, S}
\end{subfigure}\hfill
\begin{subfigure}[t]{0.3114\textwidth}
\centering
\includegraphics[width=\linewidth,viewport=160.04 0.84 305.59 130.17,clip]{figures/app_layerwise.pdf}
\caption{MLR, S}
\end{subfigure}\hfill
\begin{subfigure}[t]{0.3114\textwidth}
\centering
\includegraphics[width=\linewidth,viewport=307.24 0.84 452.79 130.17,clip]{figures/app_layerwise.pdf}
\caption{LEACE, S}
\end{subfigure}
\caption{\textbf{Sharing by depth, for three estimators.} SSI of English-counterpart directions on Bangla idioms in context by relative depth, for COV (a, d), MLR (b, e) and LEACE (c, f), F above and S below. Filled markers beat the RAND null (and, for COV, the PERM null) at $p\le.05$, unadjusted, with 199 draws each; open markers do not; rings mark the validation-selected layer. Encoders: \encoderkey. \textit{Post hoc}; no layer is reselected on these profiles. SSIs divide by Bangla's own subspace from the same estimator.}
\label{fig:app-layerwise}
\end{figure}

\Cref{fig:app-layerwise} repeats the planned readout at every evaluated layer for COV, MLR and LEACE, with 199 RAND draws per layer and, for COV, 199 PERM draws; these tests are unadjusted and descriptive. COV beats the RAND null at \NxLayerRand{} of \NxLayerCells{} non-embedding layer--label cells and both nulls at \NxLayerBoth, and its SSI exceeds MLR's at \NxLayerCovMlrPos{} of \NxLayerCovMlrCells{} layer--label cells (\NxLayerCovMlrSig{} with paired intervals above zero, \NxLayerCovMlrNeg{} below). Native-validation selection does not generally maximize observed cross-lingual SSI: the selected layer lies below the median layer's SSI in \NxLayerBelowMedian{} of \NxLayerSelCells{} COV cells, and for XLM-R F it gives \NxLayerXLMRSel{} where the best layer reaches \NxLayerXLMRMax.

\para{Fixed relative depth.} The selected layer is not stable across splits: seed 1 selected it by COV rather than MLR validation accuracy and chose a different layer in every encoder (\textit{e.g.}\ XLM-R 24 and 15, MuRIL 12 and 15, LaBSE 6 and 12). Because encoder rankings carry interpretive weight, \cref{tab:app-fixed-depth} reads every encoder at the same fractions of its depth, all of which are cached. At half depth, the planned estimator preserves MuRIL and LaBSE as the top two encoders for F (\NxDepthMlrTop; others \NxDepthMlrRest); the complete ranking correlates with that at the selected layers (Spearman $\rho=\NxDepthMlrRho$). COV directions beat both nulls at half depth in \NxDepthCovBoth{} of 6 encoders and read more there than at the selected layer in \NxDepthCovHigher{} encoders selected elsewhere (XLM-R: \NxDepthXLMRCov), so their ranking moves with depth ($\rho=\NxDepthCovRho$): COV's ranking varies with depth, whereas MLR preserves the leading pair in this comparison.

\begin{table}[t]
\caption{\textbf{SSI at fixed relative depth.} English-counterpart directions on Bangla idioms in context at a quarter,
half and three quarters of each encoder's depth and at its last layer. \underline{Underlined}: the validation-selected
layer, with its original SSI. \swatch{keytint}\,Beats the random-subspace null (MLR) or both nulls (COV) at $p\le.05$,
uncorrected (199 draws each). Exploratory. SSIs divide by Bangla's own subspace from the same estimator. SSI uses the native-reference definition and cutoff in \cref{sec:operations}.}
\label{tab:app-fixed-depth}
\centering
\small
\setlength{\tabcolsep}{5pt}
\begin{tabular}{@{}l cccc cccc@{}}
\toprule
\textbf{\hdtwo{Model}} & \multicolumn{4}{c}{\textbf{MLR (planned)}} & \multicolumn{4}{c}{\textbf{COV}} \\
\cmidrule(lr){2-5}\cmidrule(l){6-9}
 & \textbf{$\frac14$} & \textbf{$\frac12$} & \textbf{$\frac34$} & \textbf{1} & \textbf{$\frac14$} & \textbf{$\frac12$} & \textbf{$\frac34$} & \textbf{1} \\
\midrule
\band{9}{Figurative class F ($M{=}11$)}{} \\
BanglaBERT & .00 & .10 & \underline{.16} & .10 & \hlkey .33 & \hlkey .57 & \hlkey \underline{.48} & .21 \\
BanglishBERT & \hlkey .12 & \hlkey .44 & \hlkey .45 & \underline{.20} & \hlkey .71 & \hlkey .78 & \hlkey .63 & \underline{.33} \\
mBERT & $-$.44 & .26 & .31 & \underline{.20} & .09 & .18 & .28 & \hlkeyr \underline{.32} \\
XLM-R & $-$.02 & \hlkey .21 & \hlkey .20 & \hlkey \underline{.29} & \hlkey .50 & \hlkey .73 & \hlkey .80 & \hlkeyr \underline{.31} \\
MuRIL & \hlkey .20 & \hlkey \underline{.48} & \hlkey .41 & \hlkey .55 & \hlkey .69 & \hlkey \underline{.82} & \hlkey .82 & \hlkeyr .63 \\
LaBSE & \hlkey .28 & \hlkey \underline{.50} & \hlkey .46 & \hlkey .53 & \hlkey .70 & \hlkey \underline{.65} & \hlkey .73 & \hlkeyr .67 \\
\addlinespace[2.5pt]
\band{9}{Sentiment S ($M{=}3$)}{} \\
BanglaBERT & .24 & .10 & \underline{.15} & \hlkey .33 & .42 & .50 & \hlkey \underline{.62} & .41 \\
BanglishBERT & \hlkey .31 & .28 & .28 & \hlkey \underline{.38} & \hlkey .77 & \hlkey .81 & \hlkey .77 & \hlkeyr \underline{.88} \\
mBERT & \hlkey .48 & $-$.02 & .35 & \underline{.40} & \hlkey .49 & \hlkey .60 & .42 & \hlkeyr \underline{.28} \\
XLM-R & $-$.13 & \hlkey .53 & $-$.01 & \hlkey \underline{.47} & .56 & \hlkey .84 & .62 & \underline{.38} \\
MuRIL & \hlkey .31 & \hlkey \underline{.32} & \hlkey .30 & \hlkey .47 & \hlkey .81 & \hlkey \underline{.92} & \hlkey .83 & \hlkeyr 1.34 \\
LaBSE & .27 & \hlkey \underline{.30} & \hlkey .60 & \hlkey .47 & \hlkey .96 & \hlkey \underline{.87} & \hlkey 1.11 & \hlkeyr .86 \\
\bottomrule
\end{tabular}
\end{table}

\subsection{Rotations fitted on held-out pairs}
\label{app:heldout}
The rotation of \cref{app:alignment} was fitted on SPACE pairs, the records that also define the class means, so its gains could be partly circular. We refit the same two maps on the unlabeled paired inputs of PROBE, of VALIDATION or of both (\NyAlignPairs{} pairs), none of which enters an estimator, and on three shuffled pairings of PROBE and VALIDATION inputs as a null. Refitted on SPACE, the maps reproduce \cref{app:alignment} exactly. \Cref{tab:app-heldout} gives medians over encoders. The held-out map raises the overlap of the English and Bangla COV images for F from \NyAlignNoneDimsF{} to \NyAlignPVDimsF{} of 11 dimensions, the read share to \NyAlignPVReadF{} and the erase share to \NyAlignPVEraseF; shuffled pairs give \NyAlignShufDimsF, \NyAlignShufReadF{} and \NyAlignShufEraseF, and the SPACE map gives \NyAlignSpaceDimsF, \NyAlignSpaceReadF{} and \NyAlignSpaceEraseF. The held-out map thus keeps \NyAlignKeepDims\% of the SPACE map's overlap gain and \NyAlignKeepRead\% of its read gain, and the analogous multiclass ordering is observed in all \NyAlignCells{} cells. A map fitted on VALIDATION alone, which has seen neither PROBE nor SPACE inputs, gains less (\NyAlignValDimsF{} dimensions) but in the same direction.

\begin{table}[t]
\caption{\textbf{Rotations fitted on held-out pairs.} English COV directions carried into Bangla coordinates by an
orthogonal Procrustes map fitted on unlabeled paired inputs of each split (medians over six encoders; shuffled: three
draws pairing each English row with a random Bangla row). \emph{Dims}: dimensions shared with Bangla's COV image (of
$M$); \emph{read}: SSI of a Bangla-trained probe; \emph{zero-shot}: SSI of an English-trained probe; \emph{erase}:
erase share; \emph{Ordering}: cells with erase share $\le\omega\le$ read share (a multiclass diagnostic motivated by Prop.~3). PROBE and VALIDATION pairs share no
record with the SPACE inputs that define the class means. \swatch{keytint}\,The held-out map of
\cref{fig:read-erase}. Discovery split; \textit{post hoc}. SSI uses the native-reference definition and cutoff in \cref{sec:operations}.}
\label{tab:app-heldout}
\centering
\footnotesize
\setlength{\tabcolsep}{2.7pt}
\begin{tabular}{@{}l r ccccc ccccc@{}}
\toprule
\textbf{\hdtwo{Map fitted on}} & \textbf{\hdtwo{Pairs}} & \multicolumn{5}{c}{\textbf{F ($M{=}11$)}} & \multicolumn{5}{c}{\textbf{S ($M{=}3$)}} \\
\cmidrule(lr){3-7}\cmidrule(l){8-12}
 & & \textbf{Dims} & \textbf{Read} & \textbf{Zero-shot} & \textbf{Erase} & \textbf{Order} & \textbf{Dims} & \textbf{Read} & \textbf{Zero-shot} & \textbf{Erase} & \textbf{Order} \\
\midrule
\band{12}{Orthogonal map}{} \\
no map (original) & -- & 2.0 & .41 & .23 & .02 & 6/6 & 0.5 & .74 & .60 & .09 & 5/6 \\
SPACE (class-mean records) & 3,476 & 5.6 & .93 & .64 & .44 & 6/6 & 1.5 & .91 & .78 & .47 & 4/6 \\
PROBE & 2,197 & 3.0 & .71 & .46 & .16 & 6/6 & 0.9 & .70 & .68 & .20 & 4/6 \\
VALIDATION & 895 & 2.3 & .57 & .39 & .07 & 6/6 & 0.6 & .73 & .67 & .04 & 6/6 \\
PROBE + VALIDATION & 3,092 & \hlkey 3.2 & \hlkey .68 & \hlkey .50 & \hlkey .17 & \hlkey 6/6 & \hlkey 0.9 & \hlkey .75 & \hlkey .71 & \hlkey .18 & \hlkeyr 6/6 \\
shuffled pairs (null) & 3,092 & 0.7 & .18 & $-$.31 & .01 & 13/18 & 0.0 & .19 & $-$.20 & .00 & 14/18 \\
\addlinespace[2.5pt]
\band{12}{Orthogonal map in the top-128 principal plane}{} \\
SPACE (class-mean records) & 3,476 & 4.7 & .79 & .57 & .25 & 6/6 & 1.4 & .76 & .70 & .21 & 6/6 \\
PROBE & 2,197 & 3.1 & .61 & .45 & .17 & 6/6 & 0.9 & .64 & .62 & .17 & 6/6 \\
VALIDATION & 895 & 2.6 & .55 & .37 & .10 & 6/6 & 0.7 & .76 & .69 & .10 & 6/6 \\
PROBE + VALIDATION & 3,092 & 3.3 & .60 & .45 & .17 & 6/6 & 0.9 & .75 & .66 & .24 & 5/6 \\
shuffled pairs (null) & 3,092 & 1.0 & .20 & $-$.34 & .02 & 13/18 & 0.1 & .18 & $-$.39 & .00 & 12/18 \\
\bottomrule
\end{tabular}
\end{table}

\subsection{Other erasers and nonlinear readouts}
\label{app:erasers}
LEACE and orthogonal projections onto estimated images are not the only way to erase. We add INLP and R-LACE, each fitted on English SPACE inputs and, as the reference, on Bangla SPACE inputs. INLP removes, at each step, the row space of an $L_2$ multinomial logistic classifier fitted to the inputs with earlier directions removed ($K{-}1$ directions per step), until accuracy on the fitting language's VALIDATION inputs is no more than one percentage point above the majority rate. The operating point is the first successful step, or the final step of the configured 264-direction path if the criterion is not reached. R-LACE plays a linear adversary against a rank-$M$ projection relaxed to the Fantope, with its learning rate chosen per fit on a pilot. Each eraser's subspace is removed and the Bangla probe retrained, as in \cref{sec:rq4}, and MLP probes (\cref{app:nonlinear}) repeat the readouts.

\begin{figure}[t]
\centering
{\centering\includegraphics[width=0.9907\textwidth,viewport=0.00 107.68 471.00 123.84,clip]{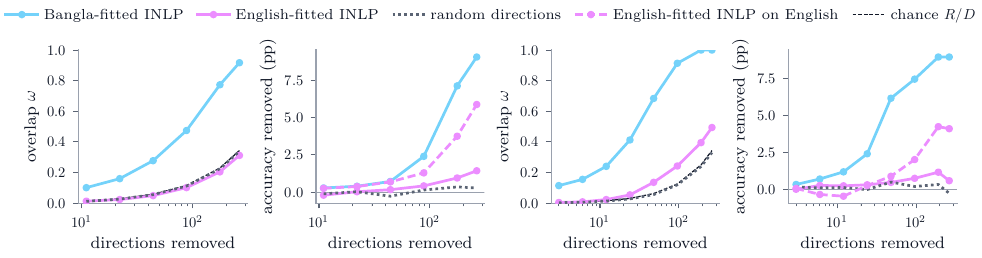}\par}
\par\medskip
\begin{subfigure}[t]{0.4770\textwidth}
\centering
\includegraphics[width=\linewidth,viewport=9.54 0.84 120.77 107.71,clip]{figures/app_inlp_path.pdf}
\caption{overlap, F}
\end{subfigure}\hfill
\begin{subfigure}[t]{0.4810\textwidth}
\centering
\includegraphics[width=\linewidth,viewport=122.45 0.84 234.66 107.71,clip]{figures/app_inlp_path.pdf}
\caption{accuracy removed, F}
\end{subfigure}
\par\vspace{10pt}
\begin{subfigure}[t]{0.4770\textwidth}
\centering
\includegraphics[width=\linewidth,viewport=236.34 0.84 347.57 107.71,clip]{figures/app_inlp_path.pdf}
\caption{overlap, S}
\end{subfigure}\hfill
\begin{subfigure}[t]{0.4810\textwidth}
\centering
\includegraphics[width=\linewidth,viewport=349.25 0.84 461.46 107.71,clip]{figures/app_inlp_path.pdf}
\caption{accuracy removed, S}
\end{subfigure}
\caption{\textbf{English-fitted INLP has limited erasure transfer to Bangla.} Along INLP's path ($K{-}1$ directions per step), medians over six encoders: (a, c)~overlap of the removed subspace with Bangla's class-mean image and (b, d)~accuracy removed, for F (a, b) and S (c, d). English-fitted INLP removes Bangla's class-mean span and accuracy about as fast as random directions of the same rank, while removing several points of English accuracy. Discovery split; \textit{post hoc}.}
\label{fig:app-inlp}
\end{figure}

INLP guards its own language in \NyInlpGuarded{} of \NyInlpFits{} fits, removing a median \NyInlpRankF{} directions for F (\NyInlpRankRangeF, about \NyInlpRankShareF\% of the space; \cref{tab:app-erasers}). English-fitted INLP removes a median \NyInlpDropF\pp{} of Bangla F accuracy (at most \NyInlpDropMaxF), Bangla-fitted INLP \NyInlpNativeF\pp, and random subspaces of the same rank \NyInlpRandDropF\pp; for S, \NyInlpDropS{} against \NyInlpNativeS\pp. The English INLP subspace contains \NyInlpOmegaEnF{} of English's class-mean span but \NyInlpOmegaF{} of Bangla's, at chance (\NyInlpChanceF) for F; for S the overlap is \NyInlpOmegaS{} against a chance of \NyInlpChanceS. Along INLP's path, the overlap with Bangla's image and the Bangla accuracy removed grow about as fast as for random directions (\cref{fig:app-inlp}). Erase shares stay below the overlap in \NyInlpPropOK{} of \NyInlpPropCells{} INLP cells, and the held-out rotation of \cref{app:heldout} raises INLP's erase share for F only to \NyRotInlpEraseF{} (COV \NyRotCovEraseF, LEACE \NyRotLeaceEraseF).

R-LACE at the concept's rank meets the three-percentage-point validation guard criterion in \NyRlaceGuarded{} of \NyRlaceFits{} fits, removes a median \NyRlaceEnglishShareF{} of English's above-floor F signal from held-out English inputs, and shares \NyRlaceOmegaEnF{} of English's class-mean span: it defeats the adversary on SPACE without removing the class-mean directions, so it erases too little, even in English, to test transfer. MLP probes see the same cross-lingual asymmetry as linear ones (erase shares \NyMlpShareF{} for COV, LEACE and INLP, against \NyLinShareF{} for linear probes), but after Bangla's own LEACE or INLP they still read \NyMlpLeftF\pp{} above the F floor, whereas linear probes read \NyLinLeftF: these erasers guard linear readouts only.

\subsection{Re-seeded and tag-built taxonomies}
\label{app:taxonomies}
F is one $k$-means partition of English glosses. We rebuilt it with $k$-means seeds 1--3 through the same pipeline (seed 0 reproduces F on every record; ARI with F \NyReseedAri), and built TAG by applying F's recipe to the corpus's topical tags after removing tags of form, register and polarity (\emph{figurative}, \emph{colloquial}, \emph{negative} and the like). TAG's 12 classes (among them arrogance and flattery, poverty and money, danger and misfortune) label \NyTagCoverage\% of records, are stable under re-seeding (ARI \NyTagAri) and agree only weakly with F (AMI \NyTagAmiF). \Cref{tab:app-taxonomies} repeats the planned test under each. Class-mean directions beat both nulls under balanced accuracy in all \NyTaxCells{} tests (\NyTaxCovAcc{} by accuracy). The planned MLR test depends on the seed: it passes in \NyTaxMlrAcc{} encoders under the three re-seeds (XLM-R in \NyTaxMlrXLMR, MuRIL in \NyTaxMlrMuRIL, LaBSE in \NyTaxMlrLaBSE) and in \NyTaxMlrTag{} under TAG. The gloss advantage of \rqn{2} holds under every re-seed and under TAG, which uses neither gloss field, in \NyGlossCovSig{} of \NyGlossCells{} encoder--taxonomy cells for COV and \NyGlossMlrSig{} for MLR. The advantage shrinks under TAG (COV \NyGlossCovTag{} against \NyGlossCovReseed\pp; MLR \NyGlossMlrTag{} against \NyGlossMlrReseed), so part of the original advantage is built into gloss-based taxonomies.

\begin{table}[t]
\caption{\textbf{The planned test under other taxonomies.} SSI of English-counterpart directions on Bangla idioms in context.
F is the original taxonomy; F1--F3 rebuild it with $k$-means seeds 1--3 (seed 0 reproduces F exactly); TAG applies F's
recipe to the corpus's topical tags without form, register and polarity tags, so it uses neither gloss field (12
classes, \NyTagCoverage\% of records). \swatch{keytint}\,Bold: beats both nulls after Holm correction over the six
encoders (999 draws each; F: the original families). \emph{Both nulls}: encoders that do so by accuracy (balanced
accuracy). \emph{Gloss $-$ counterpart}: median accuracy advantage of English-gloss over English-counterpart directions
on Bangla glosses (encoders significant after Holm). Discovery split; \textit{post hoc}. SSIs divide by Bangla's own subspace from the same estimator. SSI uses the native-reference definition and cutoff in \cref{sec:operations}.}
\label{tab:app-taxonomies}
\centering
\small
\setlength{\tabcolsep}{4.2pt}
\begin{tabular}{@{}l ccccc ccccc@{}}
\toprule
\textbf{\hdtwo{Model}} & \multicolumn{5}{c}{\textbf{MLR (planned estimator)}} & \multicolumn{5}{c}{\textbf{COV (patterns)}} \\
\cmidrule(lr){2-6}\cmidrule(l){7-11}
 & \textbf{F} & \textbf{F1} & \textbf{F2} & \textbf{F3} & \textbf{TAG} & \textbf{F} & \textbf{F1} & \textbf{F2} & \textbf{F3} & \textbf{TAG} \\
\midrule
BanglaBERT & .16 & $-$.03 & .03 & $-$.05 & .00 & \hlkey \textbf{.48} & \hlkey \textbf{.45} & \hlkey \textbf{.52} & \hlkey \textbf{.59} & \hlkeyr \textbf{.52} \\
BanglishBERT & .20 & .18 & \hlkey \textbf{.36} & .14 & .04 & .33 & \hlkey \textbf{.35} & .24 & .34 & .18 \\
mBERT & .20 & .14 & .02 & $-$.03 & -- & \hlkey \textbf{.32} & \hlkey \textbf{.35} & \hlkey \textbf{.33} & .08 & .11 \\
XLM-R & \hlkey \textbf{.29} & \hlkey \textbf{.27} & \hlkey \textbf{.31} & \hlkey \textbf{.32} & .21 & \hlkey \textbf{.31} & \hlkey \textbf{.13} & \hlkey \textbf{.40} & \hlkey \textbf{.19} & \hlkeyr \textbf{.20} \\
MuRIL & \hlkey \textbf{.48} & \hlkey \textbf{.33} & \hlkey \textbf{.49} & .15 & \hlkey \textbf{.30} & \hlkey \textbf{.82} & \hlkey \textbf{.78} & \hlkey \textbf{.77} & \hlkey \textbf{.82} & \hlkeyr \textbf{.76} \\
LaBSE & \hlkey \textbf{.50} & .22 & \hlkey \textbf{.50} & .25 & .12 & \hlkey \textbf{.65} & \hlkey \textbf{.72} & \hlkey \textbf{.84} & \hlkey \textbf{.66} & \hlkeyr \textbf{.89} \\
\midrule
Both nulls, acc.\ (bal.) & 3 (4) & 2 (4) & 4 (4) & 1 (3) & 1 (3) & 5 (6) & 6 (6) & 5 (6) & 4 (6) & 4 (6) \\
ARI with F & 1 & .52 & .48 & .63 & .13 & 1 & .52 & .48 & .63 & .13 \\
Gloss $-$ counterpart (pp) & -- & 13.8 (6) & 15.7 (6) & 14.9 (6) & 4.2 (5) & -- & 6.0 (6) & 6.2 (6) & 7.0 (6) & 5.1 (6) \\
\bottomrule
\end{tabular}
\end{table}

\subsection{Transfer along the penalty path}
\label{app:penalty-path}
\Cref{prop:penalty} makes the filter--pattern distinction a matter of degree. We refit MLR at estimator penalties $C$ from 10 to $10^{-4}$ and LEACE with a whitening ridge from 0 to 100 times the trace scale, on both splits, and read each subspace with a Bangla-trained probe and zero-shot with an English-trained probe, against 999 random subspaces (\cref{tab:app-penalty-path}). As $C$ falls, the MLR subspace approaches the COV image (\NyPathDimsDefaultF{} shared dimensions at $C{=}1$, \NyPathDimsBestF{} at $10^{-3}$ and \NyPathDimsLimitF{} at $10^{-4}$) and zero-shot transfer appears gradually. Along the path, overlap and zero-shot SSI correlate at $\rho={}$\NyPathRho. LEACE transfers once its whitening is regularized. Choosing $C$ or the ridge by the English probe's accuracy on English VALIDATION inputs, without Bangla labels in this choice, selects $C\le.01$ in \NySelMlrSmallC{} of 24 MLR fits and a ridge of 1 in \NySelLeaceRidge{} of 24 LEACE fits, and these classifiers beat the random-subspace null zero-shot in \NySelMlr{} of \NySelMlrOf{} and \NySelLeace{} of \NySelLeaceOf{} tests after Holm correction. The default penalty overfits English too: at $C{=}10^{-3}$ the English probe's accuracy on English VALIDATION inputs is \NyPathEnValGainF{} points higher for F.

\begin{table}[t]
\caption{\textbf{Transfer along the penalty path.} English-counterpart subspaces on Bangla idioms in context, medians
over six encoders. \emph{Dims}: dimensions the English subspace shares with the English COV image (of $M$; LEACE has
COV's image at every ridge). \emph{Supervised}: SSI of a Bangla-trained probe, against the same setting's Bangla
subspace; \emph{zero-shot}: SSI of an English-trained probe applied unchanged. \emph{Beats RAND}: encoders whose
zero-shot accuracy beats 999 random subspaces after Holm correction over the six encoders. \swatch{keytint}\,Bold: all
six do. The last rows choose $C$ or the ridge per encoder by the English probe's accuracy on English VALIDATION inputs,
on both splits. Discovery split unless stated; \textit{post hoc}. SSIs divide by Bangla's own subspace from the same estimator. SSI uses the native-reference definition and cutoff in \cref{sec:operations}.}
\label{tab:app-penalty-path}
\centering
\footnotesize
\setlength{\tabcolsep}{4pt}
\begin{tabular}{@{}l cccc cccc@{}}
\toprule
\textbf{\hdtwo{Setting}} & \multicolumn{4}{c}{\textbf{F ($M{=}11$)}} & \multicolumn{4}{c}{\textbf{S ($M{=}3$)}} \\
\cmidrule(lr){2-5}\cmidrule(l){6-9}
 & \textbf{Dims} & \textbf{Supervised} & \textbf{Zero-shot} & \textbf{Beats RAND} & \textbf{Dims} & \textbf{Supervised} & \textbf{Zero-shot} & \textbf{Beats RAND} \\
\midrule
\band{9}{MLR, estimator penalty $C$}{} \\
$10$ & 0.5 & .23 & .03 & 3/6 & 0.2 & .33 & $-$.12 & 0/6 \\
$1$ (default) & 0.6 & .24 & .02 & 3/6 & 0.2 & .35 & $-$.05 & 0/6 \\
$10^{-1}$ & 1.3 & .32 & \hlkey \textbf{.18} & 6/6 & 0.4 & .37 & .22 & 3/6 \\
$10^{-2}$ & 3.3 & .39 & \hlkey \textbf{.36} & 6/6 & 0.8 & .46 & \hlkey \textbf{.43} & 6/6 \\
$10^{-3}$ & 7.5 & .54 & \hlkey \textbf{.42} & 6/6 & 1.3 & .53 & \hlkey \textbf{.49} & 6/6 \\
$10^{-4}$ & 10.3 & .46 & \hlkey \textbf{.23} & 6/6 & 1.6 & .62 & \hlkey \textbf{.55} & 6/6 \\
\addlinespace[2.5pt]
\band{9}{LEACE, whitening ridge (trace scale)}{} \\
$0$ (LEACE) & 11.0 & .09 & $-$.21 & 1/6 & 3.0 & .24 & $-$.08 & 0/6 \\
$10^{-2}$ & 11.0 & .20 & $-$.04 & 3/6 & 3.0 & .35 & .11 & 1/6 \\
$1$ & 11.0 & .50 & \hlkey \textbf{.38} & 6/6 & 3.0 & .54 & \hlkey \textbf{.54} & 6/6 \\
$100$ & 11.0 & .46 & \hlkey \textbf{.26} & 6/6 & 3.0 & .68 & \hlkey \textbf{.56} & 6/6 \\
\addlinespace[2.5pt]
\band{9}{COV, the limit}{} \\
COV & 11.0 & .41 & \hlkey \textbf{.23} & 6/6 & 3.0 & .74 & .60 & 5/6 \\
\addlinespace[2.5pt]
\band{9}{Penalty chosen on English VALIDATION accuracy}{} \\
MLR, discovery & -- & .42 & \hlkey \textbf{.36} & 6/6 & -- & .50 & \hlkey \textbf{.46} & 6/6 \\
LEACE, discovery & -- & .50 & \hlkey \textbf{.38} & 6/6 & -- & .54 & \hlkey \textbf{.54} & 6/6 \\
MLR, seed 1 & -- & .70 & \hlkey \textbf{.62} & 6/6 & -- & .60 & .48 & 4/6 \\
LEACE, seed 1 & -- & .67 & \hlkey \textbf{.58} & 6/6 & -- & .61 & .58 & 5/6 \\
\bottomrule
\end{tabular}
\end{table}

\subsection{The idiom without its context}
\label{app:idiom-alone}
The planned target is the idiom in an example sentence written to illustrate it. \Cref{tab:app-idiom-alone} gives the planned design on the idiom alone in a neutral frame (\cond{bn-idiom}): MLR and COV, F and S, six encoders, 1,999 RAND and 1,999 PERM draws, accuracy, balanced accuracy and macro-F1, Holm-corrected within each family of 12. COV directions beat both nulls in \NyIsoCovModels{} for F and S (SSI \NyIsoCovTopSSI), and in \NyIsoCovBalF{} encoders for F and \NyIsoCovBalS{} for S under balanced accuracy. MLR directions do so for F only in \NyIsoMlrModelsF{} (\NyIsoMlrBalF{} encoders under balanced accuracy). Zero-shot, English classifiers through COV directions beat both nulls in \NyIsoZsBoth{} of 12 tests, all four MuRIL and LaBSE tests among them (SSI \NyIsoZsTop). The English-direction gain on the idiom alone (median \NyIsoGainIdiomF{} points for COV on F) matches that in context (\NyIsoGainCtxF), while the native gain falls from \NyIsoNativeCtxF{} to \NyIsoNativeIdiomF. On the same records, dropping the context costs native readouts \NyCtxNativeLossF{} points and English-direction readouts \NyCtxCrossLossF. As these tests were specified after their point estimates were known, \cref{app:idiom-prospective} repeats them prospectively.

\begin{table}[t]
\caption{\textbf{The planned design on the idiom without its context.} English-counterpart directions on the Bangla idiom alone
in a neutral frame (\cond{bn-idiom}): SSI of MLR and COV directions read by a Bangla-trained probe, and of COV directions
read zero-shot by an English-trained one. \swatch{keytint}\,Bold: beats both nulls after Holm correction over the 12
tests of its family (1,999 draws; zero-shot 999). $\Delta$: accuracy (points) of the idiom alone minus the idiom in
context on the 2,166 records with both inputs, through Bangla's own COV image (\emph{native}) or English COV directions
(\emph{English}); \swatch{negtint}\,two-sided $p\le.05$. The family was specified after its point estimates
(\cref{tab:app-pairs}) but before its nulls were drawn. Discovery split; \textit{post hoc}. SSIs divide by Bangla's own subspace from the same estimator. SSI uses the native-reference definition and cutoff in \cref{sec:operations}.}
\label{tab:app-idiom-alone}
\centering
\footnotesize
\setlength{\tabcolsep}{3.1pt}
\begin{tabular}{@{}l ccc cc ccc cc@{}}
\toprule
\textbf{\hdtwo{Model}} & \multicolumn{5}{c}{\textbf{F ($M{=}11$)}} & \multicolumn{5}{c}{\textbf{S ($M{=}3$)}} \\
\cmidrule(lr){2-6}\cmidrule(l){7-11}
 & \textbf{MLR} & \textbf{COV} & \textbf{Zero-shot} & \textbf{$\Delta$ native} & \textbf{$\Delta$ English} & \textbf{MLR} & \textbf{COV} & \textbf{Zero-shot} & \textbf{$\Delta$ native} & \textbf{$\Delta$ English} \\
\midrule
BanglaBERT & .31 & .69 & \hlkey \textbf{.49} & \hlneg $-$8.7 & \hlneg $-$2.8 & .17 & .72 & .70 & \hlneg $-$7.4 & \hlnegr $-$3.6 \\
BanglishBERT & .50 & .74 & .46 & \hlneg $-$5.0 & 0.2 & \hlkey \textbf{1.14} & 1.02 & .89 & $-$2.3 & $-$1.2 \\
mBERT & .64 & .39 & .10 & \hlneg $-$7.4 & \hlneg $-$3.8 & -- & .43 & .44 & \hlneg $-$3.3 & $-$0.1 \\
XLM-R & .32 & .63 & .13 & \hlneg $-$5.0 & $-$1.6 & .24 & .50 & .39 & $-$1.6 & $-$0.5 \\
MuRIL & \hlkey \textbf{.62} & \hlkey \textbf{1.05} & \hlkey \textbf{.93} & \hlneg $-$8.7 & \hlneg $-$3.6 & .27 & \hlkey \textbf{1.09} & \hlkey \textbf{1.09} & \hlneg $-$3.7 & \hlnegr $-$1.9 \\
LaBSE & .71 & \hlkey \textbf{1.08} & \hlkey \textbf{1.09} & \hlneg $-$8.3 & \hlneg $-$2.4 & .61 & \hlkey \textbf{1.13} & \hlkey \textbf{1.15} & \hlneg $-$3.1 & $-$0.6 \\
\midrule
Both nulls, acc.\ (bal.) & 1 (3) & 2 (4) & 3 &  &  & 1 (1) & 2 (4) & 2 &  &  \\
\bottomrule
\end{tabular}
\end{table}

\subsection{The idiom alone, prospectively}
\label{app:idiom-prospective}
Before any seed-1 outcome for the idiom alone was computed, a plan committed to the repository fixed the design, family, metric, nulls and decision rule, and the analysis checked that plan against its SHA-256 (prefix \texttt{\NzPlanHash}) when it ran. The plan is frozen and internal, not a public preregistration. The primary family repeats \cref{app:idiom-alone} on the follow-up split at each encoder's follow-up layer: English counterparts (\cond{en-idiom}) to the Bangla idiom alone (\cond{bn-idiom}), COV with a Bangla-trained probe, F at $M{=}11$ and S at $M{=}3$ in six encoders, scored by balanced accuracy against 1,999 RAND, PERM and RANDCOV draws (\cref{app:matched-nulls}), with Holm correction within each null over the 12 tests. A test passes when it beats all three nulls. Secondary tests apply the same rule to COV by accuracy and to MLR at $C{=}1$, and test zero-shot transfer through COV against 999 RAND and PERM draws. The plan expected MuRIL and LaBSE to pass, made no prediction for the other four encoders, and commits to reporting every result. Before any test, seed-1 COV and MLR accuracies on \cond{bn-fig} had to reproduce the follow-up's within .005, and did.

\Cref{tab:app-idiom-prospective} gives every result. COV passes in \NzProPass{} of 12 tests: for F and S in \NzProEncoders{} encoders, beyond the plan's expectation, while \NzProFail{} beats RAND and RANDCOV but not PERM for either concept. The median SSI is \NzProSsiMedian{} relative to each encoder's native COV reference, and the passing tests gain \NzProGain{} points of balanced accuracy over the floor. By accuracy, \NzProAccPass{} tests pass. MLR passes only LaBSE's F test (\cref{app:matched-nulls}). An English classifier through COV directions, fitted without Bangla readout labels at the target-selected layer, beats both zero-shot nulls in \NzProZsPass{} of 12 tests (\NzProZsPassAcc{} by accuracy; median F SSI \NzProZsMedianF). Without the sentence written to illustrate it, the idiom's own representation thus shares its coarse class with its English counterpart in every encoder but one. Seed 1 re-partitions the same corpus, so the test is internal, not independent external validation.

\subsection{A covariance-matched null and PCA}
\label{app:matched-nulls}
RAND draws directions uniformly, so a subspace that merely meets English's high-variance directions could beat it without using the labels, whereas PERM keeps the estimator's pull toward those directions but not the class structure. RANDCOV isolates that pull: each of its 1,999 draws spans $X_c^{\top}g$ for the centered English SPACE inputs $X_c$ and a Gaussian matrix $g$, so its directions follow the English covariance. PCA, English SPACE's top principal directions of the same rank, is the label-free subspace that keeps the most variance. We add both to the discovery-split design for MLR and COV, on idioms in context and alone (\cref{tab:app-matched-nulls}). The reruns reproduce the original MLR, COV and PCA accuracies (largest difference below .001). In context, COV beats RANDCOV in all \NzCovRandcovFigBal{} tests under balanced accuracy and in \NzCovRandcovFigAcc{} by accuracy, and it reads more than PCA in all \NzCovPcaFigBal{} cells, by \NzCovPcaRange{} points of balanced accuracy (by accuracy, \NzCovPcaFigAcc{} of 12 paired wild-sign tests reject after Holm correction). MLR at the default penalty beats RANDCOV in \NzMlrRandcovFigAcc{} tests by accuracy and \NzMlrRandcovFigBal{} under balanced accuracy, and reads more than PCA in \NzMlrPcaFigBal{} cells (\NzMlrPcaFigAcc{} by accuracy). In BanglaBERT, its F accuracy minus PCA's is \NzMlrPcaBanglaBERT{} points, with an interval below zero. Of the planned family's \NzPlannedRejections{} rejections, \NzPlannedSurvive{} survive RANDCOV; \NzPlannedLost{} does not. On the idiom alone, COV beats RANDCOV in \NzCovRandcovIdiomBal{} of 12 tests under balanced accuracy (\NzCovRandcovIdiomAcc{} by accuracy) and reads more than PCA in \NzCovPcaIdiomBal{} cells; MLR does so in \NzMlrRandcovIdiomBal{} and \NzMlrPcaIdiomBal. The covariance-matched null accounts for much of the default filter's reading advantage, whereas COV retains additional class-specific signal.

\subsection{Which counterparts carry the signal}
\label{app:counterpart-type}
English counterparts range from conventional English idioms to literal paraphrases of the gloss (\cref{app:human}). The pipeline's transparency covariate, the cosine between a record's counterpart and its English figurative gloss in the taxonomy's never-probed sentence encoder, splits SPACE into tertiles. No probe or subspace is fitted on this covariate. We re-estimate English COV directions from the most idiom-like third (lowest cosine), from the most paraphrase-like third (highest) and from \NzSubsetDraws{} random SPACE subsets of the same size (\NzSubsetSize{} records), leaving the Bangla side, layers and probes unchanged, so every SSI keeps the original denominator and floor (\cref{tab:app-counterpart-type}). The covariate tracks the volunteers' judgments (\cref{app:human}). On idioms in context, directions from the idiom-like third read Bangla's classes less well than the random subsets do on average in all \NzOpaqueBelowFig{} cells (by a median \NzOpaqueDropFig{} in SSI; below every random subset in \NzOpaqueBelowAllFig), and directions from the paraphrase-like third read them better in \NzTranspAboveFig. On the idiom alone the counts are \NzOpaqueBelowIdiom{} (\NzOpaqueBelowAllIdiom{} below every subset) and \NzTranspAboveIdiom{} of \NzCellsIdiom{} defined cells. Directions estimated from less gloss-like counterparts generally support weaker coarse-class reading than size-matched random subsets, while more gloss-like counterparts generally support stronger reading. The within-class human-judgment analysis provides convergent evidence for this association (\cref{app:human}).

\begin{table}[t]
\caption{\textbf{Which English counterparts carry the shared signal.} SSI under balanced accuracy of English COV directions
estimated from all SPACE counterparts, from the idiom-like third (lowest cosine between counterpart and English gloss),
from the paraphrase-like third (highest) and from 20 random subsets of the same size
(mean\,$\pm$\,sd); discovery split. \swatch{negtint}/\swatch{keytint}\,Below/above every random subset.
\textit{Post hoc}. SSI uses the native-reference definition and cutoff in \cref{sec:operations}.}
\label{tab:app-counterpart-type}
\centering
\small
\setlength{\tabcolsep}{3.7pt}
\begin{tabular}{@{}l cccc cccc@{}}
\toprule
 & \multicolumn{4}{c}{\textbf{F ($M{=}11$)}} & \multicolumn{4}{c}{\textbf{S ($M{=}3$)}} \\
\cmidrule(lr){2-5}\cmidrule(l){6-9}
\textbf{Model} & \textbf{All} & \textbf{Idiom-like} & \textbf{Paraphrase-like} & \textbf{Random} & \textbf{All} & \textbf{Idiom-like} & \textbf{Paraphrase-like} & \textbf{Random} \\
\midrule
\band{9}{Idioms in context (\cond{bn-fig})}{} \\
BanglaBERT & .58 & \hlneg .46 & .57 & .53\,{\scriptsize$\pm$.03} & .54 & \hlneg .31 & .53 & .47\,{\scriptsize$\pm$.04} \\
BanglishBERT & .50 & .44 & .58 & .49\,{\scriptsize$\pm$.05} & .72 & .44 & .52 & .56\,{\scriptsize$\pm$.08} \\
mBERT & .53 & \hlneg .27 & .51 & .45\,{\scriptsize$\pm$.04} & .38 & \hlneg .18 & .40 & .35\,{\scriptsize$\pm$.05} \\
XLM-R & .53 & \hlneg .33 & \hlkey .65 & .46\,{\scriptsize$\pm$.05} & .49 & .26 & \hlkey .86 & .43\,{\scriptsize$\pm$.18} \\
MuRIL & .87 & \hlneg .48 & \hlkey .86 & .77\,{\scriptsize$\pm$.03} & .94 & \hlneg .51 & \hlkey 1.02 & .91\,{\scriptsize$\pm$.04} \\
LaBSE & .73 & \hlneg .50 & \hlkey .87 & .74\,{\scriptsize$\pm$.03} & .94 & \hlneg .75 & 1.02 & .94\,{\scriptsize$\pm$.05} \\
\addlinespace[2.5pt]
\band{9}{The idiom alone (\cond{bn-idiom})}{} \\
BanglaBERT & .70 & \hlneg .58 & .68 & .67\,{\scriptsize$\pm$.04} & .61 & \hlneg .38 & .68 & .59\,{\scriptsize$\pm$.05} \\
BanglishBERT & .75 & \hlneg .54 & \hlkey .76 & .68\,{\scriptsize$\pm$.05} & 1.03 & \hlneg .64 & 1.03 & .96\,{\scriptsize$\pm$.07} \\
mBERT & .43 & .55 & .56 & .45\,{\scriptsize$\pm$.07} & .53 & .43 & .56 & .51\,{\scriptsize$\pm$.08} \\
XLM-R & .54 & .56 & .66 & .53\,{\scriptsize$\pm$.08} & -- & -- & -- & -- \\
MuRIL & .96 & \hlneg .57 & \hlkey 1.08 & .90\,{\scriptsize$\pm$.04} & 1.11 & \hlneg .65 & \hlkey 1.22 & 1.04\,{\scriptsize$\pm$.07} \\
LaBSE & 1.03 & \hlneg .55 & \hlkey 1.06 & .93\,{\scriptsize$\pm$.05} & 1.09 & \hlneg .73 & 1.10 & 1.01\,{\scriptsize$\pm$.07} \\
\bottomrule
\end{tabular}
\end{table}

\subsection{Intervals that refit the subspaces}
\label{app:refit}
The original intervals resample TEST components with the subspaces and probes held fixed, so they omit the error of estimating the subspaces and probes. For the planned pair on the discovery split, each of \NzRefitDraws{} draws resamples components with replacement within SPACE (the same components for both languages), PROBE, VALIDATION and TEST, refits the English and Bangla estimators and both probes (penalty tuned on the resampled VALIDATION rows) and scores the resampled TEST rows. A draw on the original rows agrees with the original predictions on at least \NzRefitAgree\% of items. \Cref{tab:app-refit} sets the percentile intervals beside the conditional ones; the refitted intervals are a median \NzRefitWidth{} times as wide. Under balanced accuracy they lie above zero for COV in all \NzRefitCovBal{} cells and for MLR in \NzRefitMlrBal{} (conditional: \NzCondCovBal{} and \NzCondMlrBal); by accuracy, in \NzRefitCovAcc{} and \NzRefitMlrAcc{} (conditional: \NzCondCovAcc{} and \NzCondMlrAcc). Estimation error thus leaves the class-mean share's sign intact but widens every interval, so rankings of encoders within a family are descriptive.

\begin{table}[t]
\caption{\textbf{Intervals that refit the subspaces.} SSI under balanced accuracy for the planned pair on the discovery split,
with two 95\% intervals: the original \emph{conditional} one, which resamples TEST components with the fitted subspaces
and probes held fixed, and a \emph{refitted} one from 200 draws that resample
components in every split and refit both estimators and both probes. \swatch{keytint}\,Refitted interval above zero.
\textit{Post hoc}. SSIs divide by Bangla's own subspace from the same estimator. SSI uses the native-reference definition and cutoff in \cref{sec:operations}.}
\label{tab:app-refit}
\centering
\small
\setlength{\tabcolsep}{5pt}
\begin{tabular}{@{}l ccc ccc@{}}
\toprule
 & \multicolumn{3}{c}{\textbf{F ($M{=}11$)}} & \multicolumn{3}{c}{\textbf{S ($M{=}3$)}} \\
\cmidrule(lr){2-4}\cmidrule(l){5-7}
\textbf{Model} & \textbf{SSI} & \textbf{Conditional} & \textbf{Refitted} & \textbf{SSI} & \textbf{Conditional} & \textbf{Refitted} \\
\midrule
\band{7}{COV (class means)}{} \\
BanglaBERT & .58 & [.51, .64] & \hlkey [.49, .71] & .54 & [.47, .62] & \hlkeyr [.38, .63] \\
BanglishBERT & .50 & [.41, .60] & \hlkey [.38, .66] & .72 & [.57, .90] & \hlkeyr [.16, 1.14] \\
mBERT & .53 & [.43, .63] & \hlkey [.32, .65] & .38 & [.24, .52] & \hlkeyr [.12, .64] \\
XLM-R & .53 & [.42, .65] & \hlkey [.27, .77] & .49 & [.28, .75] & \hlkeyr [.05, 1.41] \\
MuRIL & .87 & [.80, .95] & \hlkey [.73, .95] & .94 & [.84, 1.04] & \hlkeyr [.73, 1.05] \\
LaBSE & .73 & [.65, .81] & \hlkey [.65, .98] & .94 & [.81, 1.09] & \hlkeyr [.72, 1.21] \\
\addlinespace[2.5pt]
\band{7}{MLR (the planned estimator)}{} \\
BanglaBERT & .33 & [.26, .41] & \hlkey [.20, .45] & .15 & [.09, .23] & \hlkeyr [.01, .31] \\
BanglishBERT & .47 & [.36, .59] & \hlkey [.30, .63] & .37 & [.27, .48] & \hlkeyr [.10, .58] \\
mBERT & .50 & [.38, .66] & \hlkey [.32, .81] & .36 & [.20, .54] & [\ensuremath{-}.01, .66] \\
XLM-R & .47 & [.36, .58] & \hlkey [.34, .71] & .43 & [.33, .54] & \hlkeyr [.17, .70] \\
MuRIL & .59 & [.50, .68] & \hlkey [.46, .72] & .42 & [.30, .54] & \hlkeyr [.28, .72] \\
LaBSE & .60 & [.51, .71] & \hlkey [.39, .81] & .30 & [.21, .39] & \hlkeyr [.16, .70] \\
\bottomrule
\end{tabular}
\end{table}

\FloatBarrier
\section{Internal Replication and Mechanism Checks}
\label{app:followup}

\para{Prospective plan and scope.} The repository holds the plan frozen before seed-1 outcomes were computed, its digest, the code and the complete aggregate results. The plan was local, not publicly preregistered. We reuse the original corpus and hidden-state caches but refit the split, taxonomy, preprocessing, estimators and probes. Split-0 and split-1 TEST populations overlap (932 of 2,590 records) and can exchange training and evaluation roles, so the replication is an internal sensitivity analysis. COV layer selection uses native F validation accuracy, averaged across supported configured ranks; MLR and LEACE use that same layer. A second plan, frozen later on the same split for the idiom without its context, is reported in \cref{app:idiom-prospective}.

\para{Hypotheses and multiplicity.} \hyp{1} comprises COV against RAND and PERM nulls for F and S in BanglaBERT, MuRIL and LaBSE (12 tests). \hyp{2} compares COV with MLR on the primary pair, \hyp{3} compares English-gloss with English-counterpart COV directions on fixed Bangla-gloss targets, and \hyp{4} tests bounded erasure transfer (12 tests each). One new Holm family covers all 48, with undefined hypotheses occupying their positions conservatively. \hyp{1} uses 1,999 matched-rank random subspaces and 1,999 permutations that move label bundles between components of equal size, with plus-one Monte Carlo $p$-values, and is distinct from the original MLR family.

\para{Paired tests.} For \hyp{2} and \hyp{3} we compute per-item paired correctness differences. \hyp{4} tests the one-sided contrast $\tfrac14(a_0-a_B)-(a_0-a_E)>0$, where $a_0,a_B,a_E$ are correctness without erasure, after Bangla erasure and after English erasure. A positive contrast corresponds to an English effect below one quarter of a positive native effect, but it does not establish a zero effect. For \hyp{2}--\hyp{4}, 19,999 wild-sign draws flip component score sums, using the plus-one one-sided tail probability. This calibration requires independent components and sign symmetry at the null boundary and is not an exact randomization guarantee for observational inputs. Paired 95\% percentile intervals use 5,000 component resamples and are unadjusted. Erasure-fraction intervals omit draws with nonpositive native effect and report the valid count. \hyp{1} SSI intervals use 1,000 component resamples. All intervals condition on the fitted models.

\begin{table}[t]
\caption{\textbf{Prospective internal replication.} Encoders whose seed-1 test rejects after one Holm correction over all 48 planned tests (\swatch{keytint} all; \swatch{negtint} none), for figurative classes (F) and sentiment (S), with the range of F estimates: SSI, paired accuracy differences on matched items, or the English erasure effect as a share of the native one. $^\dagger$\hyp{1} nominates BanglaBERT, MuRIL and LaBSE; \hyp{2}--\hyp{4} include all six encoders. The table summarizes every prespecified hypothesis; the following tables report estimates and unadjusted intervals. SSIs divide by Bangla's own subspace from the same estimator. SSI uses the native-reference definition and cutoff in \cref{sec:operations}.}
\label{tab:followup}
\centering
\small
\setlength{\tabcolsep}{3.2pt}
\begin{tabular}{@{}lccr@{}}
\toprule
\textbf{Seed-1 hypothesis} & \textbf{F} & \textbf{S} & \textbf{F estimates} \\
\midrule
\hyp{1}\enspace COV beats both nulls$^\dagger$ & \hlkey 3/3 & 2/3 & SSI 0.66--0.89 \\
\hyp{2}\enspace COV beats MLR ($C{=}1$) & \hlkey 6/6 & 5/6 & 4.6--15.1\,pp \\
\hyp{3}\enspace gloss beats counterpart & \hlkey 6/6 & \hlneg 0/6 & 2.9--10.3\,pp \\
\hyp{4}\enspace erasure below $\tfrac14$ native & 4/6 & 3/6 & $-$7 to 19\% \\
\bottomrule
\end{tabular}
\end{table}
\begin{table}[t]
\caption{\hyp{1}: seed-1 COV reading. $p_R$ and $p_P$ are RAND/PERM values adjusted in the full 48-test family; intervals are unadjusted component-bootstrap intervals. Each null has 1,999 draws. SSI above one is allowed. \swatch{negtint}\,A null that is not rejected. SSI uses the native-reference definition and cutoff in \cref{sec:operations}.}
\label{tab:followup-h1}
\centering
\small
\setlength{\tabcolsep}{3.5pt}
\begin{tabular}{@{}llrcccc@{}}
\toprule
\textbf{Model} & \textbf{Label} & \textbf{$n$} & \textbf{SSI [95\% CI]} & \textbf{Cross/native/floor (\%)} & \textbf{$p_R$} & \textbf{$p_P$} \\
\midrule
\multirow{2}{*}{BanglaBERT} & F & 2,213 & 0.66 [0.57, 0.74] & 29.6/37.7/14.1 & .0140 & .0140 \\
 & S & 2,214 & 0.56 [0.40, 0.69] & 53.6/60.0/45.6 & .0140 & \hlnegr .1550 \\
\addlinespace[1.5pt]
\multirow{2}{*}{MuRIL} & F & 2,213 & 0.89 [0.82, 0.97] & 36.2/39.0/14.1 & .0140 & .0140 \\
 & S & 2,214 & 0.91 [0.79, 1.02] & 59.6/61.0/45.6 & .0140 & .0140 \\
\addlinespace[1.5pt]
\multirow{2}{*}{LaBSE} & F & 2,213 & 0.67 [0.58, 0.74] & 28.7/36.1/14.1 & .0140 & .0140 \\
 & S & 2,214 & 1.00 [0.82, 1.24] & 55.8/55.8/45.6 & .0140 & .0140 \\
\bottomrule
\end{tabular}
\end{table}
\begin{table}[t]
\caption{\hyp{2}: paired COV minus MLR cross accuracy on contextual Bangla spans. Unadjusted 5,000-draw paired component intervals; one-sided wild-sign $p$ values receive Holm-48 adjustment. All estimators share a COV-validation-selected layer and fixed probe $C=1$, so the test does not compare separately optimized estimators. \swatch{negtint}\,A test that does not reject.}
\label{tab:followup-h2}
\centering
\small
\setlength{\tabcolsep}{3.5pt}
\begin{tabular}{@{}lcccc@{}}
\toprule
\textbf{Model} & \textbf{F $\Delta$ pp [95\% CI]} & \textbf{$p$} & \textbf{S $\Delta$ pp [95\% CI]} & \textbf{$p$} \\
\midrule
BanglaBERT & 11.5 [8.9, 14.1] & .0024 & 6.0 [3.6, 8.4] & .0024 \\
BanglishBERT & 7.8 [5.8, 9.9] & .0024 & 8.1 [5.6, 10.8] & .0024 \\
mBERT & 4.6 [2.5, 6.7] & .0024 & \hlneg 2.6 [0.4, 4.8] & \hlnegr .1276 \\
XLM-R & 8.0 [5.3, 10.6] & .0024 & 5.8 [3.5, 8.3] & .0024 \\
MuRIL & 15.1 [12.8, 17.5] & .0024 & 9.8 [7.1, 12.8] & .0024 \\
LaBSE & 4.7 [2.5, 6.9] & .0024 & 4.0 [1.7, 6.3] & .0140 \\
\bottomrule
\end{tabular}
\end{table}
\begin{table}[t]
\caption{\hyp{3}: paired English-gloss minus English-counterpart COV accuracy on Bangla glosses. Unadjusted 5,000-draw paired component intervals; one-sided wild-sign $p$ values receive Holm-48 adjustment. All estimators share a COV-validation-selected layer and fixed probe $C=1$, so the test does not compare separately optimized estimators. \swatch{negtint}\,A test that does not reject.}
\label{tab:followup-h3}
\centering
\small
\setlength{\tabcolsep}{3.5pt}
\begin{tabular}{@{}lcccc@{}}
\toprule
\textbf{Model} & \textbf{F $\Delta$ pp [95\% CI]} & \textbf{$p$} & \textbf{S $\Delta$ pp [95\% CI]} & \textbf{$p$} \\
\midrule
BanglaBERT & 4.1 [2.3, 5.9] & .0024 & \hlneg 1.3 [$-$0.5, 3.3] & \hlnegr .5981 \\
BanglishBERT & 7.1 [5.1, 9.0] & .0024 & \hlneg 1.2 [$-$0.6, 2.9] & \hlnegr .5981 \\
mBERT & 5.4 [3.7, 7.1] & .0024 & \hlneg 0.0 [$-$1.6, 1.5] & \hlnegr 1.0000 \\
XLM-R & 2.9 [1.0, 4.8] & .0225 & \hlneg $-$0.5 [$-$2.4, 1.3] & \hlnegr 1.0000 \\
MuRIL & 4.3 [2.1, 6.6] & .0031 & \hlneg $-$2.5 [$-$4.4, $-$0.5] & \hlnegr 1.0000 \\
LaBSE & 10.3 [8.0, 12.6] & .0024 & \hlneg 1.3 [$-$0.1, 2.8] & \hlnegr .3648 \\
\bottomrule
\end{tabular}
\end{table}
\begin{table}[t]
\caption{\hyp{4}: erasure effects and the contrast $\frac14(a_0-a_B)-(a_0-a_E)$ in pp. Positive values favor an English effect below 25\% of the native effect. Intervals are unadjusted 95\% component intervals; $p$ is Holm-48. Negative effects mean erasure improved accuracy. The contrast tests a bound, not a zero effect or complete erasure. \swatch{negtint}\,A test that does not reject.}
\label{tab:followup-h4}
\centering
\small
\setlength{\tabcolsep}{3.5pt}
\begin{tabular}{@{}llccccc@{}}
\toprule
\textbf{Model} & \textbf{Label} & \textbf{\EN{} effect} & \textbf{\BN{} effect} & \textbf{\EN/\BN{} (\%) [CI]} & \textbf{Bound contrast [CI]} & \textbf{$p$} \\
\midrule
\multirow{2}{*}{BanglaBERT} & F & 0.3 & 12.5 & 2 [$-$7, 11] & 2.8 [1.8, 3.9] & .0024 \\
 & S & 0.3 & 8.3 & 4 [$-$11, 18] & 1.8 [0.6, 2.9] & .0280 \\
\addlinespace[1.5pt]
\multirow{2}{*}{BanglishBERT} & F & 0.6 & 9.0 & 7 [$-$3, 17] & 1.6 [0.7, 2.5] & .0031 \\
 & S & 0.9 & 9.0 & 10 [$-$2, 22] & \hlneg 1.4 [0.3, 2.4] & \hlnegr .1104 \\
\addlinespace[1.5pt]
\multirow{2}{*}{mBERT} & F & $-$0.4 & 5.6 & $-$7 [$-$25, 7] & 1.8 [1.0, 2.7] & .0024 \\
 & S & 0.6 & 3.6 & 18 [$-$3, 43] & \hlneg 0.3 [$-$0.5, 1.0] & \hlnegr 1.0000 \\
\addlinespace[1.5pt]
\multirow{2}{*}{XLM-R} & F & 0.9 & 9.1 & 10 [0, 19] & 1.4 [0.5, 2.3] & .0200 \\
 & S & 1.1 & 8.9 & 12 [1, 22] & \hlneg 1.1 [0.3, 2.0] & \hlnegr .0780 \\
\addlinespace[1.5pt]
\multirow{2}{*}{MuRIL} & F & 2.1 & 11.2 & 19 [9, 28] & \hlneg 0.7 [$-$0.3, 1.8] & \hlnegr .5981 \\
 & S & 0.1 & 10.7 & 1 [$-$10, 11] & 2.6 [1.5, 3.8] & .0024 \\
\addlinespace[1.5pt]
\multirow{2}{*}{LaBSE} & F & 1.9 & 11.7 & 16 [7, 24] & \hlneg 1.1 [0.1, 2.1] & \hlnegr .1782 \\
 & S & 0.5 & 10.5 & 5 [$-$6, 14] & 2.1 [1.1, 3.2] & .0043 \\
\bottomrule
\end{tabular}
\end{table}

\begin{figure}[t]
\centering
\begin{subfigure}[t]{0.3231\textwidth}
\centering
\includegraphics[width=\linewidth,viewport=0.84 0.84 151.78 139.56,clip]{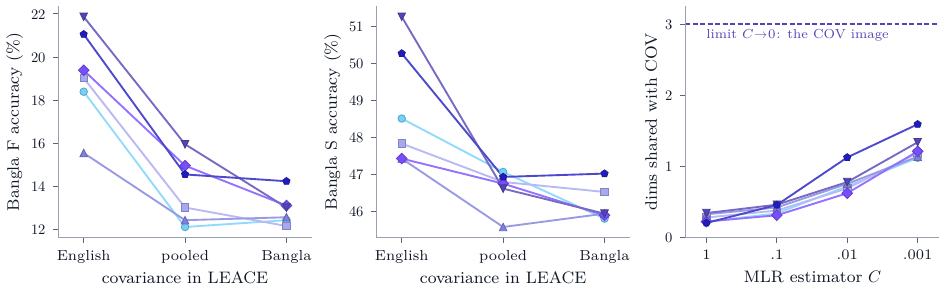}
\caption{fixed English image, F}
\end{subfigure}\hfill
\begin{subfigure}[t]{0.3231\textwidth}
\centering
\includegraphics[width=\linewidth,viewport=153.53 0.84 304.47 139.56,clip]{figures/app_followup_mechanisms.pdf}
\caption{fixed English image, S}
\end{subfigure}\hfill
\begin{subfigure}[t]{0.3137\textwidth}
\centering
\includegraphics[width=\linewidth,viewport=306.21 0.84 452.76 139.56,clip]{figures/app_followup_mechanisms.pdf}
\caption{MLR meets COV, S}
\end{subfigure}
\caption{\textbf{Seed-1 mechanism diagnostics.} (a--b)~LEACE retention of Bangla F and S with the English class-mean image fixed and the covariance in \cref{eq:operator} taken from English, pooled or Bangla SPACE features: matching the covariance lowers retention in every encoder and label. (c)~Dimensions that the English MLR subspace shares with the English COV image for S as the estimator's penalty grows; \cref{prop:penalty} predicts the limit of 3 (dashed). Encoders: \encoderkey. Exploratory; every setting is shown.}
\label{fig:app-followup}
\end{figure}

\para{Exploratory mechanism diagnostics.} \expt{2} changes only the English MLR estimator's $C\in\{.001,.01,.1,1\}$, holding the probe's $C=1$. \expt{3} fixes the English class-mean image and changes the covariance in \cref{eq:operator} to English, equal-weight pooled within-language, or Bangla SPACE covariance. Because these variants have target-feature access, they are diagnostic adaptations, not strict source-only transfer. Image agreement, idempotence and source-operator reconstruction are checked numerically. \expt{4} refits XLM-R COV and MLR after SPACE-only dimension scaling or removal of the leading 1, 3 or 10 principal components of pooled source and target SPACE features, without selecting a new layer, and measures effective dimension by the covariance participation ratio $(\sum_i\lambda_i)^2/\sum_i\lambda_i^2$. No confirmatory $p$-values are attached to \expt{2}--\expt{4}. \Cref{fig:app-followup} plots \expt{3} and the S path of \expt{2}; \cref{fig:estimators} plots the F path.

\para{Validation-selected regularization.} After inspecting the \expt{2} path, we add an explicitly \textit{post hoc} diagnostic that chooses among its four fitted settings using saved primary-pair VALIDATION accuracy only, breaking ties toward stronger regularization. The rule was frozen before these validation scores were read, but the path's TEST outcomes had already been inspected, so the diagnostic is not confirmatory. We compare the selected MLR predictions with COV on matched TEST items, with unadjusted 5,000-component intervals and no new $p$-values (\cref{tab:tuned-mlr}). All six F settings choose $C=.001$; neither estimator is separately optimized over layers.

Under pooled and target covariance, retention decreases in all 12 model--label comparisons, whereas erasure changes are small and mixed; source-operator reconstruction and idempotence errors stay below $10^{-5}$. Stronger MLR regularization improves all 12 endpoint comparisons, and the intermediate F improvements are monotonic in five encoders, with a small reversal in mBERT. Removing ten PCs raises the participation-ratio dimension of XLM-R from 15.2 to 243.6, yet COV F readout accuracy falls from 29.8\% to 24.3\%: better isotropy alone does not guarantee better cross-lingual reading.

\begin{table}[t]
\caption{\expt{2}: cross accuracy (\%) over the English MLR estimator regularization path; readout $C=1$ is fixed. COV is the observed \expt{1} comparator. Every setting is shown; no test-set optimum is selected. \swatch{keytint}\,The strongest penalty, more accurate than the default $C=1$ in every row. These diagnostics are exploratory.}
\label{tab:e2}
\centering
\small
\setlength{\tabcolsep}{9pt}
\begin{tabular}{@{}lccccc@{}}
\toprule
\textbf{\hdtwo{Model}} & \multicolumn{4}{c}{\textbf{MLR estimator $C$}} & \textbf{\hdtwo{COV}} \\\cmidrule(lr){2-5} & \textbf{$.001$} & \textbf{$.01$} & \textbf{$.1$} & \textbf{$1$} & \\
\midrule
\band{6}{Figurative class F}{rank 11} \\
BanglaBERT & \hlkey 30.0 & 24.7 & 20.9 & 18.1 & 29.6 \\
BanglishBERT & \hlkey 28.3 & 24.7 & 21.0 & 19.7 & 27.6 \\
mBERT & \hlkey 20.5 & 16.8 & 15.7 & 15.8 & 20.3 \\
XLM-R & \hlkey 30.6 & 26.5 & 22.9 & 21.9 & 29.8 \\
MuRIL & \hlkey 35.2 & 29.9 & 23.9 & 21.1 & 36.2 \\
LaBSE & \hlkey 30.7 & 29.2 & 26.0 & 24.0 & 28.7 \\
\addlinespace[2.5pt]
\band{6}{Sentiment S}{rank 3} \\
BanglaBERT & \hlkey 55.9 & 52.4 & 49.2 & 47.6 & 53.6 \\
BanglishBERT & \hlkey 58.4 & 52.8 & 49.9 & 49.4 & 57.5 \\
mBERT & \hlkey 47.7 & 46.4 & 46.8 & 46.8 & 49.4 \\
XLM-R & \hlkey 56.0 & 51.5 & 49.2 & 48.6 & 54.4 \\
MuRIL & \hlkey 59.9 & 54.1 & 51.1 & 49.8 & 59.6 \\
LaBSE & \hlkey 58.3 & 59.0 & 57.2 & 51.9 & 55.8 \\
\bottomrule
\end{tabular}
\end{table}
\begin{table}[t]
\caption{Exploratory baseline: estimator $C$ chosen from the four \expt{2} values by saved primary-pair validation accuracy alone (ties: smaller $C$); accuracies (\%) with unadjusted 5,000-draw component-bootstrap intervals. This \textit{post hoc} diagnostic lies outside the 48-test family and is not an equivalence test. Every F selection is $.001$, and most large default-COV advantages disappear: only two intervals exclude zero, one favoring COV (\swatch{keytint}) and one MLR (\swatch{negtint}).}
\label{tab:tuned-mlr}
\centering
\small
\setlength{\tabcolsep}{3.5pt}
\begin{tabular}{@{}llccccc@{}}
\toprule
\textbf{Model} & \textbf{Label} & \textbf{Chosen $C$} & \textbf{Val.\ MLR} & \textbf{TEST MLR} & \textbf{TEST COV} & \textbf{COV$-$MLR pp [CI]} \\
\midrule
\multirow{2}{*}{BanglaBERT} & F & .001 & 28.2 & 30.0 & 29.6 & $-$0.4 [$-$2.3, 1.4] \\
 & S & .001 & 58.0 & 55.9 & 53.6 & $-$2.3 [$-$4.5, 0.1] \\
\addlinespace[1.5pt]
\multirow{2}{*}{BanglishBERT} & F & .001 & 30.1 & 28.3 & 27.6 & $-$0.8 [$-$2.3, 0.9] \\
 & S & .001 & 60.5 & 58.4 & 57.5 & $-$0.9 [$-$3.1, 1.2] \\
\addlinespace[1.5pt]
\multirow{2}{*}{mBERT} & F & .001 & 21.8 & 20.5 & 20.3 & $-$0.1 [$-$1.8, 1.7] \\
 & S & .01 & 51.0 & 46.4 & 49.4 & \hlkeyr 3.0 [0.9, 5.1] \\
\addlinespace[1.5pt]
\multirow{2}{*}{XLM-R} & F & .001 & 27.5 & 30.6 & 29.8 & $-$0.8 [$-$2.4, 0.7] \\
 & S & .001 & 57.0 & 56.0 & 54.4 & $-$1.6 [$-$3.5, 0.5] \\
\addlinespace[1.5pt]
\multirow{2}{*}{MuRIL} & F & .001 & 33.9 & 35.2 & 36.2 & 1.0 [$-$0.6, 2.7] \\
 & S & .001 & 62.4 & 59.9 & 59.6 & $-$0.3 [$-$2.0, 1.5] \\
\addlinespace[1.5pt]
\multirow{2}{*}{LaBSE} & F & .001 & 31.9 & 30.7 & 28.7 & \hlnegr $-$1.9 [$-$3.6, $-$0.3] \\
 & S & .1 & 61.4 & 57.2 & 55.8 & $-$1.4 [$-$3.5, 0.9] \\
\bottomrule
\end{tabular}
\end{table}
\begin{table}[t]
\caption{\expt{3}: fixed-English-image LEACE accuracy (\%) with English, pooled or Bangla covariance, for retention and erasure. Covariance uses SPACE only. Target covariance does not rescue retention: English covariance retains the most in every row (\swatch{keytint}); these variants do not change the English class-mean image. Exploratory.}
\label{tab:e3}
\centering
\small
\setlength{\tabcolsep}{8pt}
\begin{tabular}{@{}lcccccc@{}}
\toprule
\textbf{\hdtwo{Model}} & \multicolumn{3}{c}{\textbf{Read through the image}} & \multicolumn{3}{c}{\textbf{Erase the image}} \\\cmidrule(lr){2-4}\cmidrule(l){5-7} & \textbf{\EN} & \textbf{pooled} & \textbf{\BN} & \textbf{\EN} & \textbf{pooled} & \textbf{\BN} \\
\midrule
\band{7}{Figurative class F}{rank 11} \\
BanglaBERT & \hlkey 18.4 & 12.1 & 12.4 & 31.6 & 31.4 & 31.9 \\
BanglishBERT & \hlkey 18.9 & 13.0 & 12.2 & 27.5 & 28.1 & 27.7 \\
mBERT & \hlkey 15.5 & 12.4 & 12.6 & 21.2 & 20.7 & 20.7 \\
XLM-R & \hlkey 19.2 & 15.0 & 13.1 & 27.4 & 27.9 & 28.2 \\
MuRIL & \hlkey 22.0 & 16.0 & 13.0 & 29.4 & 30.0 & 30.7 \\
LaBSE & \hlkey 21.1 & 14.6 & 14.2 & 28.3 & 27.8 & 28.0 \\
\addlinespace[2.5pt]
\band{7}{Sentiment S}{rank 3} \\
BanglaBERT & \hlkey 48.5 & 47.1 & 45.8 & 52.1 & 52.0 & 52.2 \\
BanglishBERT & \hlkey 47.9 & 46.8 & 46.5 & 51.2 & 51.1 & 51.6 \\
mBERT & \hlkey 47.4 & 45.6 & 45.9 & 46.9 & 47.3 & 47.1 \\
XLM-R & \hlkey 47.4 & 46.7 & 45.9 & 52.6 & 52.3 & 52.6 \\
MuRIL & \hlkey 51.3 & 46.6 & 45.9 & 55.1 & 55.6 & 55.8 \\
LaBSE & \hlkey 50.2 & 46.9 & 47.0 & 56.8 & 57.0 & 56.7 \\
\bottomrule
\end{tabular}
\end{table}
\begin{table}[t]
\caption{\expt{4}: XLM-R English-direction accuracy (\%) at fixed seed-1 layer 15, with pooled-SPACE covariance diagnostics. Higher isotropy is not monotonically related to cross-lingual reading: removing ten PCs makes the space the most isotropic (\swatch{keytint}) yet gives the lowest accuracy for every readout (\swatch{negtint}). Every setting is retained; exploratory.}
\label{tab:e4}
\centering
\small
\setlength{\tabcolsep}{3.5pt}
\begin{tabular}{@{}lcccccc@{}}
\toprule
\textbf{XLM-R setting} & \textbf{Eff.\ dim.} & \textbf{Top PC (\%)} & \textbf{F COV} & \textbf{F MLR} & \textbf{S COV} & \textbf{S MLR} \\
\midrule
baseline & 15.2 & 19.6 & 29.8 & 21.9 & 54.4 & 48.6 \\
standardized & 30.8 & 15.4 & 30.8 & 21.0 & 56.0 & 48.5 \\
remove\_pc\_1 & 23.5 & 19.3 & 29.0 & 22.2 & 54.3 & 50.5 \\
remove\_pc\_3 & 147.3 & 3.7 & 30.6 & 22.3 & 55.3 & 49.0 \\
remove\_pc\_10 & \hlkey 243.6 & \hlkey 1.5 & \hlneg 24.3 & \hlneg 15.7 & \hlneg 48.9 & \hlnegr 46.6 \\
\bottomrule
\end{tabular}
\end{table}

\FloatBarrier
\section{Decoder}
\label{app:decoder}

\begin{figure}[!htb]
\centering
\includegraphics[width=0.5\linewidth]{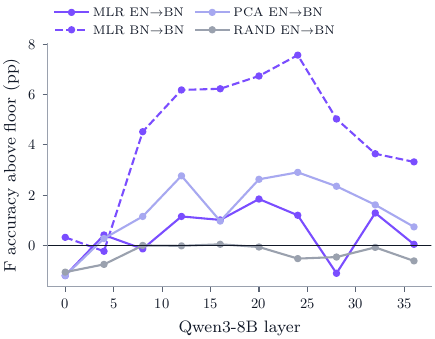}
\caption{\textbf{Qwen3-8B by layer.} F accuracy above floor (rank 3) for English-direction MLR, PCA and RAND subspaces on Bangla idioms in context, and native Bangla MLR (dashed). Exploratory; 3--9 null draws.}
\label{fig:app-decoder}
\end{figure}

We extracted hidden states of the decoder Qwen3-8B \citep{yang-etal-2025-qwen3} at every fourth layer in bf16 and selected layer 16 on VALIDATION. The estimators were RAND, PCA, MLR and LEACE at ranks 1--3, with 9 RAND and PERM draws for the planned pair and 3 for other pairs. The minimum attainable Holm-adjusted $p$ across the two decoder primary tests is therefore 0.2, and no decoder test can reject. These decoder results are descriptive because the null-draw budget cannot resolve significance at the stated familywise threshold. \Cref{tab:app-decoder,fig:app-decoder} report the results, in which the ladder and the direction asymmetry match those of the encoders.

\begin{table}[!htb]
\caption{Qwen3-8B, layer 16, MLR rank 3. Pairs: gloss = \EN{} gloss $\rightarrow$ \BN{} gloss; translation and
counterpart $\rightarrow$ \BN{} idiom in context; reverse = \BN{} idiom in context $\rightarrow$ \EN{} counterpart.
Exploratory (9 null draws for the planned pair, 3 otherwise).}
\label{tab:app-decoder}
\centering
\small
\setlength{\tabcolsep}{2.5pt}
\begin{tabular}{@{}lrrrc@{}}
\toprule
\textbf{Pair} & \textbf{Cross (\%)} & \textbf{Native (\%)} & \textbf{Floor (\%)} & \textbf{SSI [95\% CI]} \\
\midrule
\band{5}{Figurative class F}{} \\
gloss & 34.3 & 35.1 & 13.9 & .96 [.86, 1.09] \\
translation & 18.7 & 21.9 & 15.7 & .48 [.25, .74] \\
counterpart & 16.7 & 21.9 & 15.7 & .16 [$-$.06, .37] \\
reverse & 23.9 & 26.1 & 13.3 & .83 [.69, 1.01] \\
\addlinespace[2.5pt]
\band{5}{Sentiment S}{} \\
gloss & 68.1 & 70.3 & 46.3 & .91 [.86, .96] \\
translation & 53.3 & 57.7 & 46.4 & .61 [.39, .87] \\
counterpart & 51.8 & 57.7 & 46.4 & .48 [.34, .65] \\
reverse & 57.7 & 65.3 & 46.9 & .59 [.46, .74] \\
\bottomrule
\end{tabular}
\end{table}

\FloatBarrier
\section{Robustness and Negative Results}
\label{app:robustness}

\para{Readout family and description length.} Within Bangla, MLP probes on the full representation improve accuracy over linear probes by several points. The primary SSI uses linear probes on projected inputs and therefore measures the linearly accessible share of class information. The complementary MLP and online-codelength analyses in \cref{app:nonlinear} support the class-mean reading pattern.

\para{Literal labels.} We machine-translated literal glosses with NLLB-200 and assigned them to the F taxonomy. One class receives 55.6\% of TEST literal labels, so the literal label is too imbalanced to support claims about literal--figurative competition. On records whose literal and figurative codes differ, literal-label error inside F subspaces (38.3--38.7\%) matches the conditional majority error (38.3\%).

\para{Convergence.} All retention probes in the reported grids converged. Of 50,426 grid rows, 52 have an unconverged leakage probe, mostly in mBERT and XLM-R, while the selected LEACE and COV erasure comparisons converge. The follow-up audit checks 24,720 stored readouts, including nulls, and all report successful solver stopping. The separate isotropy readouts also pass. These flags allow objective-stagnation stopping and do not certify every gradient against the nominal tolerance.

\FloatBarrier
\section{Identity Retrieval Audit}
\label{app:identity-audit}
Identity experiments use two Bangla contexts and so do not test English--Bangla identity sharing. Because repeated masked-context strings can produce exact nearest-neighbor ties, raw top-1 figures alone can depend on deterministic candidate order; a separate audit in the repository supplies unique-context and tie-aware results. The visible-span result likewise admits a lexical shortcut, because the same idiom form occurs in both contexts. Neither result establishes the absence of cross-lingual idiom identity information.

\FloatBarrier
\section{Reproducibility and Resources}
\label{app:repro}

\para{Pipeline and protocol.} The public code repository (\coderepo) contains the analysis pipeline, configurations and derived tables; original corpora and model weights are obtained from their cited providers. It also holds the code, run instructions and aggregate results of the analyses of \cref{app:newexp}, with the frozen plan of \cref{app:idiom-prospective}, and the item-level tests of \cref{app:human}. Discovery: split seed 0 and taxonomy seed 0. Follow-up: split seed 1, taxonomy seed 0, refitted on the new SPACE subset. All encoders use layers at a step of 3 and ranks $\{1,2,3,4,8,11\}$ (F) and $\{1,2,3\}$ (S). Figures are drawn with Matplotlib through LuaLaTeX, and Bangla text is set in Kalpurush and embedded as vector snippets, so the paper compiles with pdfLaTeX alone.

\para{Probes and estimators.} The default linear probes are L2 logistic regressions with $C{=}1$; analyses that tune the penalty or use MLP readouts identify those choices in their protocols. Linear probes are solved by batched GPU L-BFGS with nominal gradient tolerance $10^{-5}$ or objective-stagnation stopping, with objective and batching checks against scikit-learn. A successful stop is not a certificate that every float32 gradient is below the nominal tolerance. Estimators use SPACE rows only, and preprocessing (centering and global scaling) is fitted on SPACE.

\subsection{Resources and reproduction details}
\label{app:resources}
Backbone parameter counts in \cref{tab:models} exclude unused language-model heads and were obtained by instantiating the pinned configurations on the meta device, and model names such as Qwen3-8B refer to the full released family. Encoder extraction used fp16, and Qwen3-8B used bf16. Maximum input length was 128 tokens, with batch size 8. Cached features were read-only during the follow-up. The never-probed all-MiniLM-L6-v2 label constructor uses revision \texttt{1110a243fdf4}, 12-means with 10 initializations for the main fit, and 5 for stability refits.

Bagdhara \citep{sakhawat-etal-2026-words} is distributed under CC BY-SA 4.0. Our export is dated 2026-03-01; its checksum, input-disjoint component statistics, model revisions, software versions and layer-selection scores are in the repository. We do not redistribute original model weights or the raw corpus. XLM-R is MIT-licensed; mBERT, LaBSE and Qwen3 have Apache-2.0 model metadata. BanglaBERT and BanglishBERT's official distribution specifies noncommercial research under CC BY-NC-SA 4.0. The pinned MuRIL-large model metadata does not declare a license; we document the model source instead of assigning it an unverified license. NLLB's distilled-600M model has CC BY-NC 4.0 terms and contributes only appendix literal-gloss diagnostics and, through its encoder, the FBN clusters.

\para{Budget and machine.} Runs used one NVIDIA RTX 3090 with 24\,GiB of GPU memory and a host with 128\,GiB of RAM. The original primary, identity and decoder phases took approximately 5.8, 0.9 and 2.8 hours. The cached-feature \expt{1}--\expt{4} follow-up took 13.8 minutes, including new SPACE labels, layer selection, nulls and mechanism diagnostics; software versions are recorded in its provenance. The analyses of \cref{app:newexp} ran on the same machine (PyTorch 2.6.0 with CUDA 12.4, Transformers 4.57.6) from the frozen caches, with new forward passes only for the pooling-matched design. In the first eight-analysis robustness round, the seven analyses following the reproduction check completed in about two hours. Every analysis records its code hashes, options and environment, and an automatic verification checks it against the original tables. These durations are observed phase times, not a measured energy or carbon budget; development, failed preflight and setup attempts and an earlier interrupted exploratory run add unquantified overhead. The repository also includes all 48 follow-up tests and raw-score aggregates, and the volunteers' answer sheets (\cref{app:human}).

\end{document}